\documentclass[]{clv3}

\usepackage{xcolor}
\definecolor{darkblue}{rgb}{0, 0, 0.5}

\usepackage{tipa}
\usepackage{times}
\usepackage{latexsym}
\usepackage{multirow}   
\usepackage{graphicx}   
\usepackage{enumitem}
\usepackage{amsmath}    
\usepackage{amssymb}    
\usepackage{url}    
\usepackage{colortbl, booktabs} 
\usepackage{makecell} 
\usepackage{color}  
\usepackage{array}
\usepackage{booktabs}
\usepackage{float}
\usepackage{arydshln} 
\usepackage{algorithm}
\usepackage{algpseudocode}
\usepackage{phonrule}
\usepackage{hyperref}
\hypersetup{colorlinks=true,citecolor=darkblue, linkcolor=darkblue, urlcolor=darkblue}
\DeclareMathOperator*{\argmin}{arg\,min}

\begin{document}

\issue{1}{1}{2016}


\runningtitle{Train \& Constrain: Tongue Twister Generation}

\runningauthor{Loakman, Tang and Lin.}

\pageonefooter{Action editor: Tal Linzen. Submission received: 15 March 2024; revised version received: 15 July 2024; accepted for publication: 20 September 2024.}

\title{Train \& Constrain: Phonologically Informed Tongue Twister Generation from Topics and Paraphrases}

\author{Tyler Loakman}
\affil{Department of Computer Science, The~University of Sheffield,\textit{ tcloakman1@sheffield.ac.uk}}

\author{Chen Tang\thanks{The co-first author.}}
\affil{Department of Computer Science, University of Manchester, \textit{chen.tang@manchester.ac.uk}}

\author{Chenghua Lin\thanks{The corresponding author.}}
\affil{Department of Computer Science, University of Manchester,\textit{ chenghua.lin@manchester.ac.uk}}

\maketitle

\begin{abstract}

Previous work in phonologically and phonetically grounded language generation has mainly focused on domains such as puns and poetry. In this article, we present new work on the generation of English tongue twisters - a form of language that is required to be conditioned on a phoneme level to maximize sound overlap, while maintaining semantic consistency with an input topic or phrase and still being grammatically correct. We present \textbf{TwisterLister}, a pipeline for generating phonologically informed tongue twisters from large language models (LLMs) that we use to generate \textbf{TwistList 2.0}, the largest annotated dataset of tongue twisters to date, consisting of 17K+ examples from a combination of human and LLM authors. Our generation pipeline involves the use of a phonologically constrained vocabulary alongside LLM prompting to generate novel, non-derivative tongue twister examples.  We additionally present the results of automatic and human evaluation of smaller models trained on our generated dataset to demonstrate the extent to which phonologically motivated language types can be generated without explicit injection of phonological knowledge. Additionally, we introduce a phoneme-aware constrained decoding module (\textbf{PACD}) that can be integrated into an autoregressive language model and demonstrate that this method generates good quality tongue twisters both with and without fine-tuning the underlying language model. We also design and implement a range of automatic metrics for the task of tongue twister generation that is phonologically motivated and captures the unique essence of tongue twisters, primarily based on phonemic edit distance (\textbf{PED}).\footnote{Code and resources available at \url{https://github.com/tylerL404/Train-and-Constrain-TT}}
\end{abstract}

\section{Introduction}

\begin{figure}[t]
\centering
\includegraphics[width=1\linewidth]{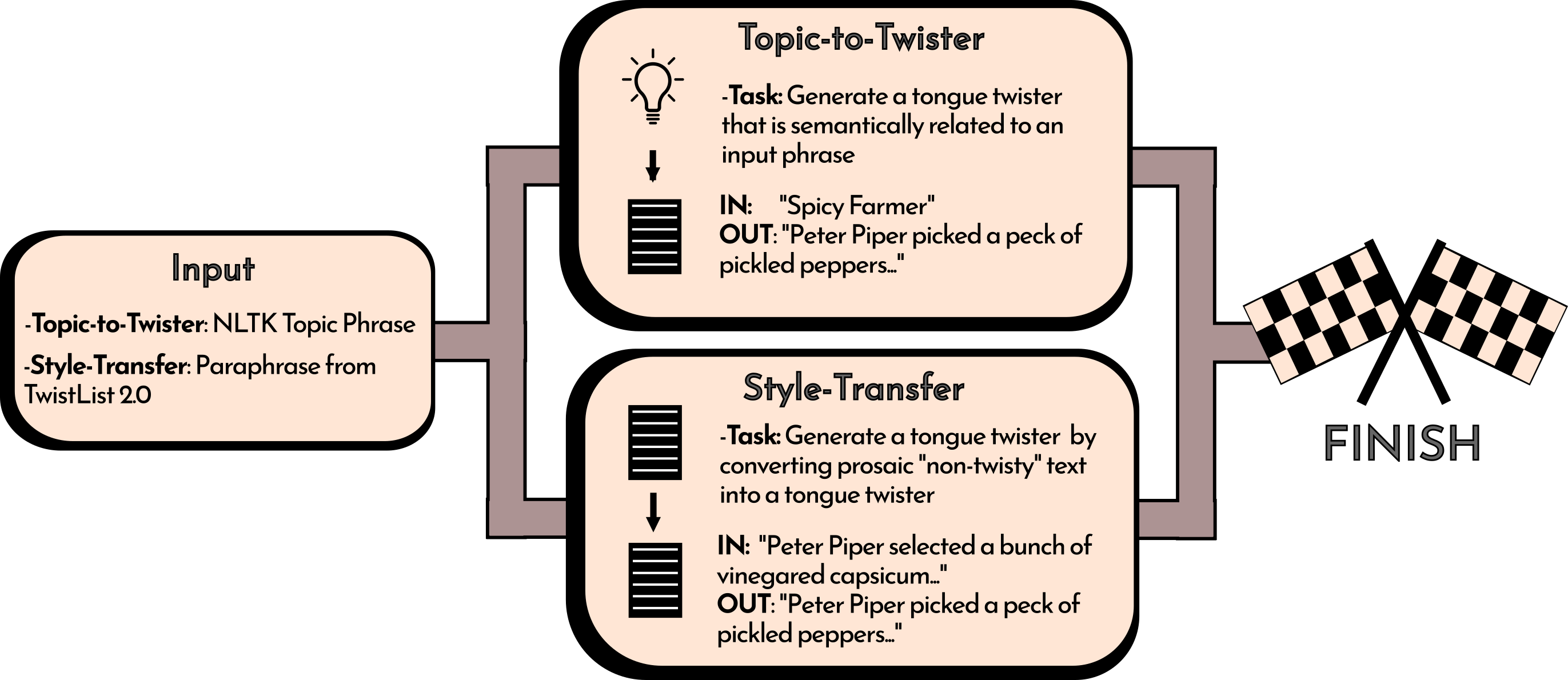}
\caption{Outline of the tongue twister generation pipeline for both the topic-to-twister and style-transfer task settings. Both settings use the same base models and hyperparameters and differ only in the data used for finetuning.}
\label{fig:task-outline}
\end{figure}

Whilst the dawn of large language models (LLMs) such as OpenAI's GPT-4 \cite{openai2023gpt4}, Meta's Llama 2 \cite{llama2} and Google's Gemini \cite{gemini} has brought unprecedented performance improvements in many natural language generation (NLG) tasks, these models are highly resource-hungry in terms of data, compute, and API expenses. Consequently, many works have started to investigate the ways in which LLM capabilities and knowledge can be infused into smaller models, using the larger model for data enhancement via the generation of pseudo-labels~\cite{tang-etal-2023-enhancing} or additional training examples~\cite{yang2023improving}. 

Additionally, LLMs are primarily designed to select the most probable continuation of a span of text based on their training data. 
Creative language, in direct opposition to non-creative language, is predominantly desired to be non-derivative, containing phrases and word sequences that are not ubiquitous in everyday language in order to evoke various emotions and engage a reader rather than purely convey information in a linguistic form. As a result, creative language generation's goals are at odds with the primary training paradigm of LLMs, as the goal is often \textit{not} to select the most probable continuation, and instead surprise and engage readers \cite{roush-etal-2022-language}.

In particular, tongue twisters represent a type of phonologically constrained language that aims to engage a reader with high levels of phoneme overlap to encourage mispronunciations, often containing rhyme and humorous semantics, or simply conveying information in a form that is enjoyable to read due to the articulatory patterns that the lexical choices present. Consequently, tongue twister generation presents myriad unique challenges for NLG, including the need to consider the phonetic realization and underlying phonological representation of the chosen vocabulary, all while still maintaining a grammatically valid output sequence despite the often obscure and highly restricted candidate vocabulary. In addition to being a fruitful area for further investigation by the NLP/NLG communities, tongue twisters also present a wide range of real-world applications, making the case for their automatic generation even stronger. These applications include (1) being used as an educational tool for language teaching \cite{Sugiharto-Japanese, SOMOFF-UNMASTERY, WILSHIRE-1999}; (2) being a source of entertainment and humor stemming primarily from unintentional misarticulation; (3) as a literary device for engaging young children in developing their literacy (such as the approach taken in numerous Dr. Seuss stories, \citealp{seuss}); (4) as a method of designing memorable slogans and tag lines \cite{guerini-etal-2015-echoes}; and (5) as stimuli in neuroscience and physiology research to investigate the localization of functions within the brain and how linguistic perception links to production on a neurological level \cite{oHalloran-apnoea, wong-broca, Kember-dysarthria}. Consequently, the ability to automatically generate tongue twisters constrained on particular topics and phoneme combinations has many real-world applications. Moreover, findings from the generation of tongue twisters also have wider applicability in phoneme-conditioned language generation, such as the more widely studied areas of poetry and lyric generation, where being able to exert phoneme-level control of the output is desirable.

\subsection{Contributions}

Towards the automatic generation of high-quality tongue twisters, we expand upon prior works to present \textbf{TwisterLister}, an LLM-based pipeline for the generation of unique, non-derivative tongue twisters to provide more extensive training data to enable the training of smaller language models. TwisterLister employs semantic and phonological knowledge in the form of sentence embeddings and phonemic edit distance to restrict a candidate vocabulary list to pass to an LLM. In doing so, we create \textbf{TwistList 2.0}, the largest existing dataset of tongue twisters with over 17k examples, of which approximately 15k are derived directly from the proposed TwisterLister pipeline. We motivate the need for this extended wealth of training data by demonstrating the impact on both automatic metrics and human evaluation as a function of training-data volume by fine-tuning various smaller-scale language models (such as BART and Flan-T5, etc.) on various splits of this dataset. We present these results as part of two different tongue twister generation approaches, \textbf{topic-to-twister} and, inspired by \citet{keh-etal-2023-pancetta}, \textbf{style-transfer} (i.e., prose-to-twister). With the aforementioned real-world applications of tongue twisters, we motivate the topic-to-twister setting for applications such as language learning, where multiple outputs can be generated to test an individual's articulatory abilities in a new language whilst simultaneously expanding their vocabulary. On the other hand, the style-transfer setting is motivated by applications such as marketing, where a standard sentence conveying a desired meaning (e.g., a brand's mission statement or a product's features) can be reworded to have increased phonetic complexity to become a tongue twister, consequently engaging the reader and increasing memorability. We additionally introduce \textbf{PACD}, a \textbf{P}honeme \textbf{A}ware \textbf{C}onstrained \textbf{D}ecoding module, that can be used with any causal autoregressive language model to ensure token outputs meet phoneme-level criteria. Overall, our contributions may be summarized as follows:

\begin{itemize}
    \item \textbf{TwisterLister} - A phonologically and semantically informed pipeline for generating English tongue twisters with large language models, both as a stand-alone generation method and as a data synthesis approach for additional training data.
    \item \textbf{TwistList 2.0} - The most extensive English tongue twister dataset to date, containing 17,000+ tongue twisters produced via TwisterLister ($\sim$15k) and human authors ($\sim$2k), including extensive quality control procedures, for use in training tongue twister generation models as well as presenting a resource for the study of this language form on a linguistic level.\footnote{To the best of our knowledge, the previous record belongs to the original TwistList (1.0) from \citet{twistlist}, containing just over 2.1k human-authored examples.}
    \item \textbf{PACD} - A phoneme-aware constrained decoding algorithm that applies hard lexical constraints on the outputs of autoregressive language models to achieve phoneme-level overlap and generate tongue twisters.
    \item \textbf{iPED/oPED} - Novel phonemic edit distance (PED) based metrics for assessing the articulatory characteristics of tongue twisters on a word-initial and overall basis.
    \item A range of experiments training smaller language models (i.e., GPT-2, DialoGPT, BART, Flan-T5, ByT5, and Baichuan) to generate tongue twisters in topic-to-twister and style-transfer settings using TwistList 2.0, including extensive automatic and human evaluation.
    \item Extensive qualitative linguistic analysis of generations from different models trained on varying quantities of training data, with or without the constrained decoding PACD module, in the form of case studies.
\end{itemize}

\section{Phonetics \& Phonology}

Due to the strong reliance on theory and ideas from the fields of phonetics and phonology in this article, we find it apt to begin with a short introduction to these domains. Firstly, phonetics refers to an area of linguistics that studies the production and perception of speech sounds present in spoken languages \cite{gick2013articulatory,Jessen_forensics, ladefoged1996acoustic} whilst the related field of phonology focuses more strongly on the abstract mental representations of speech sounds and the development of feature-based taxonomies for the categorization of related sounds \cite{ClementsG_features, DeLacy_phonology, Klausenburger_phonotactics}.

\subsection{Place \& Manner of Articulation}
\begin{figure}[h!t]
    \centering
    \includegraphics[width=1\linewidth]{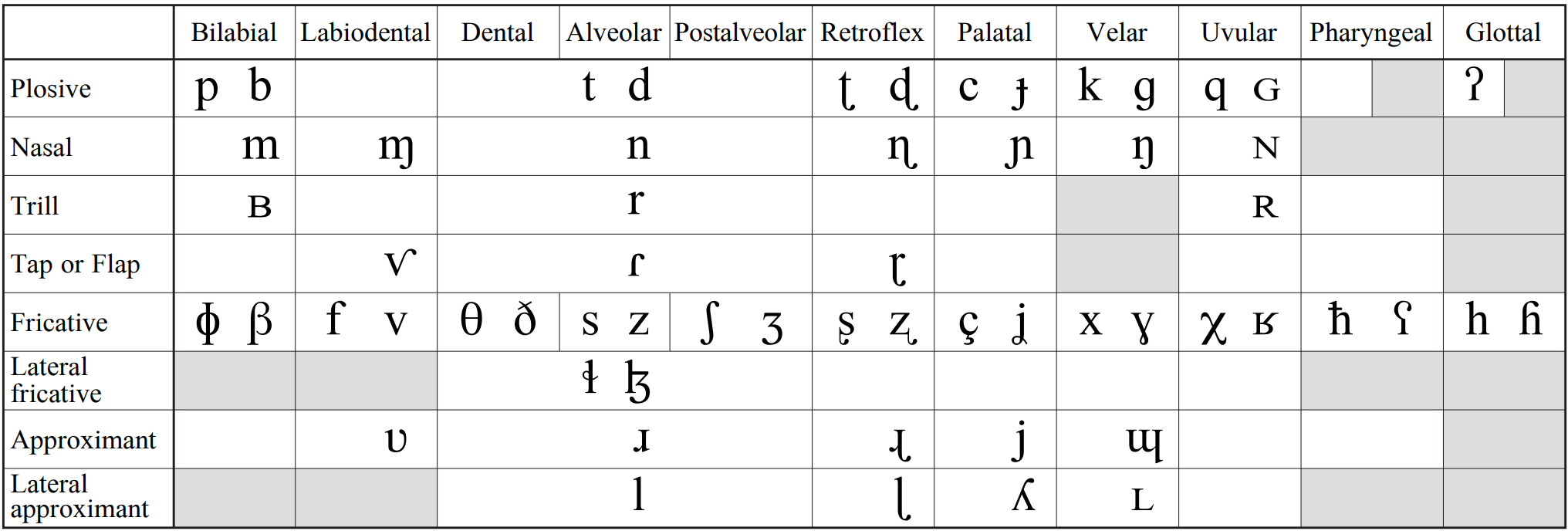}
    \caption{The International Phonetic Alphabet consonant chart. Symbols to the right of a cell are voiced (i.e., involve the vibration of the vocal folds) and sounds to the left are unvoiced. Gray boxes present place/manner combinations that have been deemed impossible for human production. This chart is reproduced from the International Phonetic Association, which distributes it under the CC-BY-SA license. }
    \label{fig:IPA-consonants}
\end{figure}

\autoref{fig:IPA-consonants} presents the primary pulmonic consonants present in human languages. For each consonant, three important pieces of information on their \textit{phonetic} characteristics are interpretable. Firstly, the rows each represent a \textit{manner} of articulation which refers to the physical process that occurs to produce a particular sound (such as a "plosive" involving the build-up and sudden release of air pressure within the mouth, whilst a "nasal" involves the nasal cavity through the lowering of the velum). On the other hand, each column refers to a \textit{place} of articulation, which relates to the main location that articulators (e.g., tongue, teeth, and lips) make contact within the vocal tract (for example, "bilabial" sounds involve both lips and "labiodental" sounds involve the lips and teeth in their production). Finally, the last remaining detail is voicing, which refers to whether or not the glottal folds (also known as vocal folds or vocal cords) are vibrating during the sound's production.

\subsection{Phonological Features}

In addition to physical production-based characteristics, there are also further \textit{phonological} features that can be used to explain the patterns and processes that phonemes undergo in human speech. An example of the different phonological features that phonemes may have is presented in \autoref{tab:feature_chart}. It is phonological features such as these that the phonemic edit distance (PED) measures introduced in \S\ref{sec:TwisterLister} and used throughout this article rely on to determine the similarity between two speech sounds (and therefore the likelihood of mispronunciation when transitioning between them).

\begin{table}[ht]
\centering \small
\resizebox{1\linewidth}{!}{
\begin{tabular}{|c|c|c|c|c|c|c|}
\toprule
  \textbf{Sound} & \textbf{Consonantal} & \textbf{Sonorant} & \textbf{Voiced} & \textbf{Nasal} & \textbf{Continuant} & \textbf{Coronal} \\
  \midrule
  \textipa{/t/} & + & - & - & - & - & + \\
  \textipa{/d/} & + & - & + & - & - & + \\
  \textipa{/s/} & + & - & - & - & + & + \\
  \textipa{/l/} & + & + & + & - & + & + \\
  \textipa{/i/} & - & + & + & - & + & -\\
  \bottomrule
\end{tabular}
}
\caption{A phonological feature chart. \textbf{+} means the feature is present for a given sound, whereas \textbf{-} means the feature is absent. Only a subset of possible sounds and phonological features are shown for demonstration purposes.}
\label{tab:feature_chart}
\end{table}

\subsection{Phonetic Transcription Standards}
Multiple transcription standards exist for phonetic and phonological representations of text. In the majority of this article, the preferred standard is the International Phonetic Alphabet (IPA), where each speech sound is represented by a single symbol (e.g., \textipa{/t, d, s, z/}). The IPA is the standard transcription convention used within linguistic research and is used within \autoref{fig:IPA-consonants} and \autoref{tab:feature_chart}.
We additionally make use of another transcription standard, ARPABET, which uses 1 or 2 characters to represent a given sound (or 3 where vowels have their stress marked). \autoref{tab:standards} presents an example of these standards, alongside another common standard, SAMPA.

\begin{table}[ht]
\centering \small
\resizebox{1\linewidth}{!}{
    \begin{tabular}{|c|c|c|c|}
        \toprule
        \textbf{Standard Orthography} & \textbf{IPA} & \textbf{ARPABET} & \textbf{SAMPA} \\
        \midrule
        Hello World & \textipa{/hEl@U w@:ld/} &  HH EH L OW  W AX L D & hEl@U w@:ld \\
        \bottomrule
    \end{tabular}
    }
    \caption{Phonetic transcriptions of the phrase "Hello World" in the 3 most common standards. IPA refers to the International Phonetic Alphabet. Note that this is just one of the possible productions. Accents differ in their phoneme inventories and phonological representations, which would necessitate different transcriptions.}
    \label{tab:standards}
\end{table}

\section{Related Work}

\subsection{Creative Language Generation}

Numerous efforts have been made toward the generation of creative language forms, with a range of findings regarding whether or not popular LLMs truly exhibit human-level creativity in these tasks. \citet{chakrabarty2023art} present work applying the Torrance Tests of Creative Thinking (TTCT) to objectively analyze the outputs of LLMs and human authors on a narrative writing task, finding that LLM generations perform measurably worse than humans, passing 3-10x fewer criteria outlined by the TTCT. However, it is important to note that this was in comparison to professional authors, who represent a very niche subset of the best human writers available. On the other hand, \citet{gomez-rodriguez-williams-2023-confederacy} compare human and LLM-authored narratives and find that LLMs are able to match or surpass human performance on several of the evaluation criteria they present. However, in this case, the "creativity" of the task was somewhat diminished by having prescribed rules about the topic, characters, and writing style, where the task may be more construed as emulating the writing of an existing work. However, similarly, both \citet{franceschelli2023creativity} and \citet{clark-etal-2021-thats} observe that humans are infrequently able to distinguish creative works written by other humans from those authored by LLMs, with the latter often achieving very high-quality outputs. Overall, there is clear potential and room for improvement in the field of automatically generating creative language forms.

A popular trend is to investigate the extent to which models can be trained to generate language forms where training data is scarce. \citet{wockener-etal-2021-end} investigate this for the generation of poetry using $\sim$16k and $\sim$67k quatrains of English and German poetry, respectively, and notice difficulties in GPT-2 learning sub-lexical phenomena including rhyme from this number of training examples alone. However, poetry presents a highly restrictive form of literary language where many types contain formal constraints regarding length, syllable count, and metrical patterns. Additionally, \citet{van-de-cruys-2020-automatic} presents work on the generation of Shakespearean sonnets, another literary niche that contains an even more limited number of available training samples. They therefore approach the task via adding constraints at decoding time in a pipeline approach that includes the stages of content planning, rhyme generation, and output polishing to imbue literary sensibilities into the outputs of models trained exclusively on non-literary text. Finally, \citet{popescu-belis-etal-2023-gpoet} use data synthesis techniques to increase the amount of rhyming data they can train GPT-2 \cite{GPT-2} on, to realize GPoeT, which shows an increased ability to generate consecutive rhymes.

Tongue twister generation, as a niche subdomain of creative language generation (which in itself is a branch of NLG more generally), has only received attention in recent years. \citet{keh-etal-2023-pancetta} presented PANCETTA, the first major work on the automatic generation of this language form in the modern post-BERT \cite{bert} NLP era, and released TT-Corp, a dataset of over 600 tongue twisters taken from various online sources. They train variations of naive and "phoneme aware" GPT-2-based models \cite{GPT-2} in a topic-to-twister and style-transfer setting, progressing the inclusion of phoneme-level awareness by pre-training models on the International Phonetic Alphabet (IPA) representation of WikiText data, and utilizing these models in conjunction with off-the-shelf orthographic models with the aim of exploiting the link between the phonological and orthographic representations of text.
Shortly following PANCETTA, \citet{twistlist} presented the precursor to the present work and released TwistList (referred to as TwistList \textit{1.0} in this article), a dataset of 2.1k human-authored tongue twisters collected from various online sources in-line with \citet{keh-etal-2023-pancetta}. Additionally, TwistList was used to train a wide range of language models, including GPT-2 \cite{GPT-2}, DialoGPT \cite{dialogpt}, BART \cite{bart} and T5 \cite{T5} solely in a topic-to-twister setting based on orthographic text. Two naive phonemic evaluation metrics were introduced, including Phoneme Overlap (PO) and Initial Phoneme Overlap (IPO), which assessed the homogeneity of the generations in terms of unique sounds. However, these metrics considered all sounds to be equidistant in phonological space (hence being naive).

Other forms of language where phonemic and phonetic information are essential have also been generated, including rap \cite{xue-etal-2021-deeprapper,manjavacas-etal-2019-generation, potash-etal-2018-evaluating} and song lyrics more generally \cite{tian-etal-2023-unsupervised, chang-etal-2023-sudowoodo,zhuo2023lyricwhiz,zhang-etal-2022-qiuniu}. Computational research on creative works is not restricted to such domains, however, with extensive work also existing in the area of narrative generation \cite{hong-etal-2023-visual-writing, tang-etal-2022-etrica, chen-etal-2021-graphplan}, humor generation \cite{loakman-etal-2023-iron,sun-etal-2022-context, tian-etal-2022-unified}, metaphor processing \cite{wang2024mmte,wang-etal-2023-metaphor, li-etal-2023-framebert, li-etal-2022-secret}, and music generation \cite{li2024mert,yuan-etal-2024-chatmusician,yu-etal-2023-musicagent}.

\subsection{Constrained Generation}
An inherent paradox often exists when using language models for creative language generation. Models trained to generate the most probable sequence of tokens are used to output text where the most \textit{probable} continuation is sometimes one of the least \textit{preferable}. In the simplest form, control over the output form can be exercised through simple methods such as restricting the output vocabulary at decoding time \cite{hokamp-liu-2017-lexically,valitutti-etal-2013-everything}, or through the application of penalties for $n$-gram repetition, therefore encouraging diversity \cite{zhang-etal-2021-generic,foster-white-2007-avoiding}.
Several works have been performed in the direction of toolkits to aid in constrained language generation. \citet{roush-etal-2022-language} present the Constrained Text Generation Studio (CTGS), an AI-writing assistant that in its most basic form applies a range of pre-made filters to the output of a probabilistic language model (such as "avoid the letter \textit{e}", \citealp{Wright_2016}), scanning the most probable next-word generations and only selecting those which fall in line with the selected constraints (of which numerous can be applied in parallel). However, due to the design of the constraints, CTGS struggles with the generation of well-formed outputs for models trained using the predominant paradigm of sub-word tokenization.
Similar approaches have also been utilized for the generation of other non-creative language types, such as machine translation and summarization, where stylistic constraints and editorial decisions may result in a preferred output structure. For example, \citet{yao2023collie} present COLLIE, a grammar-based framework for the application of advanced compositional constraints on the outputs of language models, in addition to a tool for generating example task instances from raw text. Additionally, \citet{iso2022autotemplate} presents AutoTemplate, a method of formatting a task structure to realize lexically constrained text generation.
However, whilst many constraint-based systems exist, most work on lexical constraints is focused on the inclusion of specific word choices within the output, and therefore the desired candidate vocabulary has to be known \textit{a-priori}. \citet{lu-etal-2022-neurologic} propose to solve a common downfall of autoregressive decoding where it is necessary to plan ahead and introduce \textit{NeuroLogic A*esque}, a decoding approach that uses lookahead heuristics to more carefully consider future token generations. 
We build upon work in the area of constrained generation in \S\ref{sec:PACD} with PACD, a phoneme aware constrained decoding approach that dynamically applies constraints on allowable tokens at each generation timestep.

\subsection{Knowledge Distillation}
Alongside the advent of LLMs, so too arose the area of knowledge distillation \cite{gupta_KD_survey,hinton2015distilling, bucilua_compression} whereby the aim is to achieve the successful transfer of knowledge from a much larger model (often referred to as a teacher model) to one or more smaller models (often referred to as student models). In doing so, much more robust generalist models, such as GPT-3 \cite{GPT-3} and GPT-4 \cite{openai2023gpt4} can have elements of their abilities passed down to smaller models that do not possess the same level of computing requirements \cite{Yang2023EffectiveDO}. Whilst there are numerous methods of achieving the desired distillation, perhaps the simplest and easiest of these approaches, particularly in domains where data is scarce, is to generate new synthetic training data using the larger models, either via the generation of completely new examples or via data augmentation and perturbation \cite{whitehouse-etal-2023-llm, askari-etal-2023-expand}. However, generating novel instances is not always straightforward, with the quality of generations often being dependent on the task type. For instance, \citet{li-etal-2023-synthetic} find that more subjective tasks may result in lower quality synthetic data, such as not reflecting the same level of diversity as human-written equivalents, something that is key to creative language domains and that we aim to overcome in \S\ref{sec:twisterlister_pipe} for the creation of TwistList 2.0.

\section{TwistList 2.0 Dataset Construction}
\label{sec:TwisterLister}

\begin{figure}[t]
\centering
\includegraphics[width=1\linewidth]{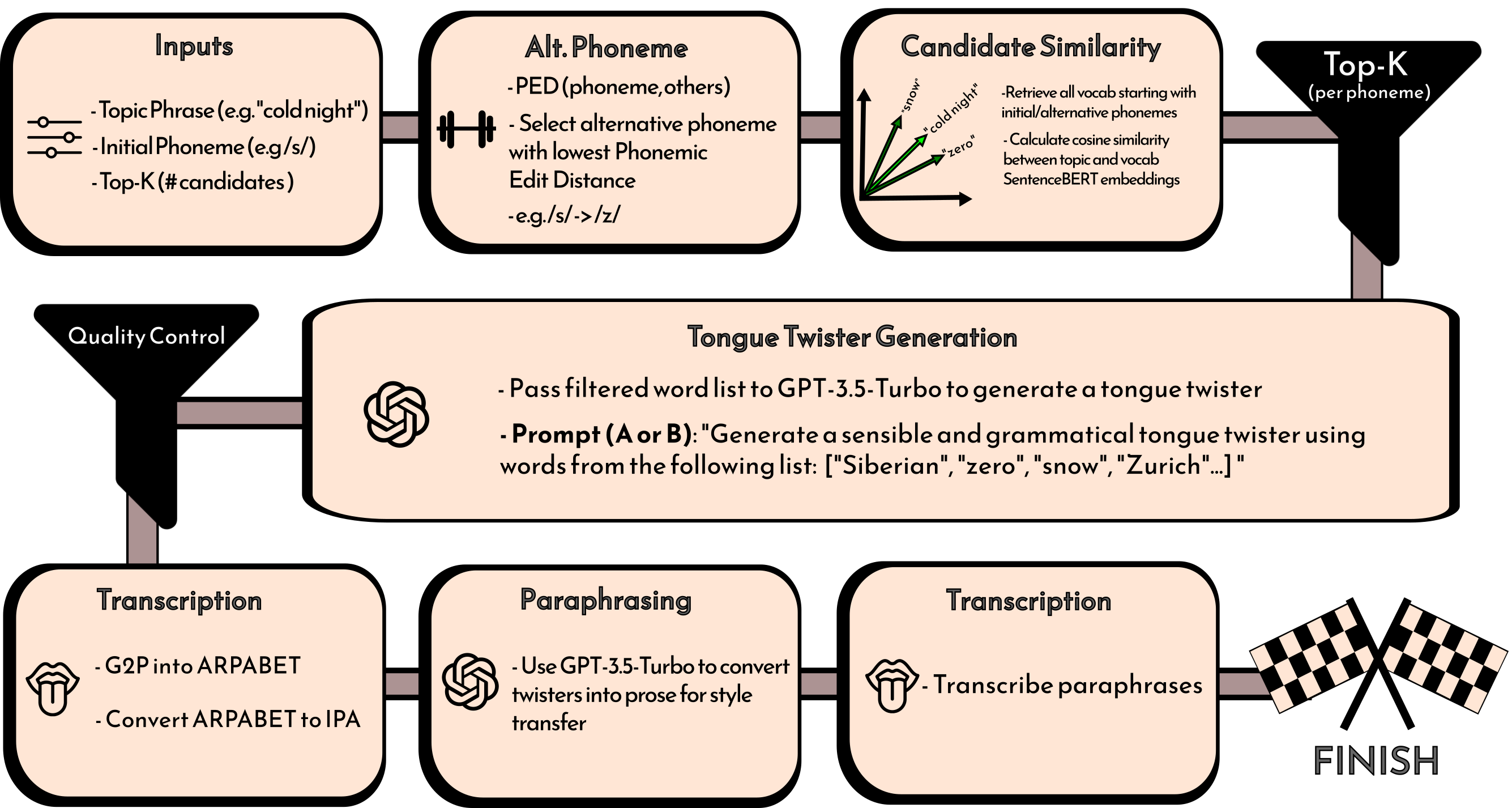}
\caption{Illustrated outline of the TwisterLister dataset generation pipeline used to create TwistList 2.0. PED refers to phonemic edit distance and G2P refers to grapheme-to-phoneme conversion.}
\label{fig:pipeline}
\end{figure}

\citet{twistlist} presented \textit{TwistList 1.0}, a dataset of 2.1k+ human-authored tongue twisters from various sources available on the web including listicles and works of fiction. While this allowed a high level of quality in the examples within the dataset (as all instances were automatically filtered, reviewed by a linguist, and then underwent quality control on a subset), the small size of this dataset (even whilst being the largest dataset of tongue twisters we are aware of to-date) means that smaller models struggle to learn the key features of tongue twisters from such limited examples, specifically regarding the need to balance high levels of phonemic repetition alongside maintaining grammatical coherence. 

To combat this shortfall, we extend TwistList 1.0 8-fold into \textit{TwistList 2.0}, containing 17,000+ unique tongue twister examples. To achieve this feat, we note the near-human performance that was achieved by early versions of ChatGPT in our previous human evaluation studies \cite{twistlist}, and opt to build a generation pipeline we name \textit{TwisterLister} using GPT-3.5-Turbo to generate novel examples to facilitate the training of smaller parameter count models from the resulting synthetic dataset.\footnote{In both \citet{twistlist} and the present work, we access GPT-3.5-Turbo (i.e., "ChatGPT") via the OpenAI API.}

\subsection{TwisterLister Pipeline}\label{sec:twisterlister_pipe}
A key discovery in previous work \cite{twistlist} was the common reliance of ChatGPT on slightly modifying well-known existing tongue twisters that had been memorized from the training data when presented with a new topic (for example, “\textit{silver shiny ships}” generated “\textit{How much wood could a woodchuck chuck if a woodchuck could chuck silver shiny ships}”). To avoid this pitfall, we create a more constrained pipeline for tongue twister generation that promotes the generation of unique, non-derivative examples, illustrated in \autoref{fig:pipeline}. 
Initially, we generate the topics for the tongue twisters by building a set of topic phrases by combining a randomly sampled adverb or adjective with a noun to represent an abstract topic (using part-of-speech tags from NLTK's Brown Corpus, \citealp{francis1979brown}). Following this, we then randomly select a phoneme on which to focus the tongue twister, restricting these choices to consonants (due to these being more commonly exploited in tongue twisters than vowels) and additionally removing any phonemes that are not legal in word-initial position in standard English phonotactics (such as the glottal stop \textipa{/P/}) or phonemes with very few entries in our vocab bank (such as the voiced postalveolar fricative \textipa{/Z/}). 
Following this, we search the CMU Pronouncing Dictionary (CMUDict)\footnote{Available at \url{https://github.com/Alexir/CMUdict/tree/master}.} for words starting with our preferred phoneme and calculate the cosine similarity between the SentenceBERT embedding of our NLTK-generated topic phrase with each candidate word retrieved from the CMUDict. Following this, the top N retrieved candidate words with the highest semantic similarity are kept, and the others are discarded (N = 5 or 10). 

We then calculate the pairwise weighted phonemic edit distance (PED) between our initially selected phoneme and all other allowable phonemes, and select the lowest scoring phoneme (i.e., most similar) to act as the best secondary phoneme.\footnote{Phonemic edit distance is implemented with the \textit{panphon} package \cite{mortensen-etal-2016-panphon}.} Where multiple share the same edit distance, the first reached when iterating over the list is selected. This is due to tongue twisters frequently relying on the reader mispronouncing a sound, often due to confusion with a phonetically/phonemically similar sound (e.g., “she sells sea shells” exploiting \textipa{/S/} and \textipa{/s/}). Consequently, allowing the generation process to select from 2 banks of words that start with similar sounds means that we can more directly promote mispronunciations rather than solely relying on the repetition of a single sound. We then repeat the process of generating a candidate list of words that are semantically related to the topic and start with our desired secondary phoneme. 

We then combine the word lists and shuffle their order to promote alternation between words with different, yet similar, initial phonemes as experimentation showed that requesting the LLM to “alternate” between words from 2 different word banks, and presenting them separately, still resulted in the words being used largely in the same order that they were presented, and therefore not resulting in the desired alternation. The list is then fed into the LLM (GPT-3.5-Turbo, accessed August-September 2023 via the API) with one of the prompts presented in \autoref{tab:prompts}.

\begin{table}[t]
\centering \small
\resizebox{1\linewidth}{!}{
    \begin{tabular}{c}
        \toprule
        \underline{\textbf{Prompt A}} \\
        ``\textit{Generate a sensible and grammatical tongue twister using words from the following list: [word-list].}\\
        \textit{The output should be a single sentence and be grammatical and coherent.}'' \\
        \midrule
        \underline{\textbf{Prompt B}} \\
        ``\textit{Generate a tongue twister by primarily using words from the following list: [word-list].}\\
        \textit{The output should be grammatical and coherent.}'' \\
        \bottomrule
    \end{tabular}
}
\caption{The two prompts used in the TwisterLister pipeline for tongue twister generation from GPT-3.5-Turbo. \textit{[word-list]} refers to the phonetically and semantically conditioned vocabulary selected in the previous steps.}
\label{tab:prompts}
\end{table}

In Prompt A, where we specify single-sentence outputs, we see much more concise outputs that may suffer from coherence issues, whilst in Prompt B we see more coherent outputs but which often resemble standard poetry more than tongue twisters (to which we apply further filtering as discussed in \S\ref{Style-Transfer-dataset}). Consequently, we use a combination of tongue twisters generated with either prompt to have a diverse range of styles.\footnote{We did not perform extensive prompt-engineering to arrive at these but observed the differing behavior in our preliminary testing.} All generations are performed with \textit{max\_tokens} set to 1000 and a temperature of 0.8 to facilitate creative completions. In total, we generated 17,500 example tongue twisters, of which 11,500 were generated with Prompt A, and 6,500 with Prompt B.\footnote{The imbalance here is due to wishing to promote the generation of "twisty" content, which is more prevalent with Prompt A. Consequently, we sample fewer Prompt B examples to contain a moderate amount of literary/poetic text.}

\subsection{Style-Transfer Paraphrase Generation}

In line with \citet{keh-etal-2023-pancetta}, we present an additional task setting alongside the topic-to-twister approach that utilizes style-transfer. Whilst \citet{keh-etal-2023-pancetta} generate non-tongue twister examples of their dataset entries via simple rule-based synonym replacement conditioned on part-of-speech tags, we leverage GPT-3.5-Turbo to paraphrase each entry in our tongue twister dataset. To achieve this, we pass the following system prompt to GPT-3.5-Turbo: \textit{"In this task you will pretend that you're an author who is rewriting existing works into a non-literary form that more resembles prose. You will be presented with a tongue twister and asked to rewrite it using synonym replacement so that there are no longer high levels of phonetic overlap and sound repetition. Example 1: INPUT = "She sells sea shells by the seashore." OUTPUT = "The girl sells conches by the ocean." Example 2: INPUT = "Peter Piper picks pickled peppers" OUTPUT = "Peter Piper selects preserved capsicums"}. We then present the following \textit{user} message to GPT-3.5-Turbo for each dataset entry: \textit{"INPUT = "[twister], OUTPUT = "} where \textit{[twister]} is a standard tongue twister from TwistList 2.0. This approach is superior to simple synonym substitution as the LLM can dynamically select new vocabulary terms in a way that further ensures the new paraphrase uses combinations of vocabulary that are sensible. This is because raw synonym replacement can result in semantic drift (as very few words are true synonyms), potentially risking nonsensical outputs \cite{chiang-lee-2023-synonym}.

\subsection{Refining Outputs}
\label{Style-Transfer-dataset}

To further refine the generated outputs, we process the resulting dataset in several stages. Firstly, all outputs generated using Prompt A (which promotes succinct tongue twisters that are not as coherent) are re-fed into GPT-3.5-Turbo in the prompt “\textit{Improve the following tongue twister by editing it so that is makes more sense and is grammatical: [tongue twister]}”. Consequently, this step fixes errors that arise from the original word lists given to the initial prompt containing morphological variants that are difficult to turn into a coherent output (for example, all nouns being in possessive form, or verb tenses being mixed in a way that is detrimental to coherence).
To further check that the remaining outputs are sensible, we calculate the perplexity (PPL) of the generations using a pre-trained language model (in this case GPT-2) as a basic heuristic for well-formedness. We then compare these scores with the average PPL across the original TwistList dataset of human-authored tongue twisters and remove any outputs from our new dataset that have perplexities that are higher than the original dataset’s mean plus one standard deviation. This stage removed 397 examples.

Additionally, as previously mentioned, Prompt B promoted the generation of longer, more poetic outputs, at the cost of not always resembling tongue twisters. In order to maintain only the best generations, we apply a metric for assessing the phonological characteristics of tongue twisters using weighted phonemic edit distance (PED) \cite{mortensen-etal-2016-panphon}, which is outlined in further detail in \S\ref{phonetic_metrics}. Again, we compare the results to the mean score from the original TwistList dataset, and filter our new dataset by removing any examples that do not score lower (i.e., better) than the original dataset’s mean plus one standard deviation. However, no examples were caught in this filtering stage, suggesting that the majority of the more poetic works still exhibited tongue-twister-esque phonetics.

Next, to encourage diversity, we remove examples based on word-overlap. To achieve this, we apply a pairwise fuzzy-matching algorithm based on the token sort ratio (which is indifferent to word order) across the remaining tongue twisters and remove examples with more than a 60\% overlap with an existing twister in the dataset.\footnote{We arbitrarily decide which to remove and continue until no remaining samples exhibit this level of overlap due to these examples having passed the other two layers of filtering, and therefore being acceptable. We implement this using the RapidFuzz package: \url{https://github.com/rapidfuzz/RapidFuzz}.} This step consequently led to the removal of 1,747 entries, therefore reducing repetition and increasing diversity in the remaining dataset.

We then filter out offensive examples that may have been created by poor topic/phoneme combinations that resulted in undesirable stereotypes or associations being expressed. These include words relating to the topics of racism, sexism, homophobia, transphobia, and additional terms that some may find offensive (including general expletives and references to various anatomy). We perform this by comparing the tongue twisters with a bank of offensive words and removing any twisters containing any of the examples (regardless of context). This stage removed a final 69 entries from our dataset.

Finally, to ensure the diversity of input/output pairs when training in a topic-to-twister fashion, we remove any entries with duplicate NLTK topic phrases, as these cases would result in 2 tongue twister outputs for a single input (but with different phonological characteristics, as they would have been caught and filtered by previous stages if not). This stage removed a final 135 examples.

The final additions to the dataset comprise 15,151 examples, which when combined with the existing TwistList 1.0 results in 17,278 unique tongue twisters in TwistList 2.0. We maintain a distinction between human and machine-authored tongue twisters in the final dataset so that different communities can make use of whichever is most pertinent to their application.

\subsection{Additional Processing \& Quality Control}
\label{sec:QC}
As in \citet{twistlist}, we then enhance the dataset with the addition of phonetic transcription using the \textit{g2p-en} Python package.\footnote{Available at: \url{https://github.com/Kyubyong/g2p/tree/master}.} We experimented with other grapheme-to-phoneme (G2P) solutions to see if an improvement was possible here such as SoundChoice \cite{soundchoice} which better accounts for correctly transcribing homographs, but opted to utilize g2p-en again. This is primarily due to our dataset generation pipeline relying on the CMUDict to retrieve vocabulary, and g2p-en queries the CMUDict for transcriptions before falling back on a trained neural network model to infer pronunciations for out-of-vocabulary (OOV) tokens. Consequently, the majority of our less common vocabulary will already have a gold standard transcription in CMUDict (in the General American accent, due to the limitations of this resource). We then convert these transcriptions into the International Phonetic Alphabet (IPA) to facilitate the use of our phoneme-based metrics, but also to provide a transcription standard that is more common in the linguistics domain.

Finally, whilst in \citet{twistlist} RAKE \cite{rake} was used to extract keywords from the human-authored examples to represent the topic, we skip this stage here and utilize the topic phrases we used in the dataset generation step. As a result, unlike the keywords for the original tongue twister collection, those that are new to TwistList 2.0 have more abstract topics as the topic words are not forced to appear in the output twister (rather, only a semantic link is present). We hypothesize that this may also help to reduce drawbacks seen in the original work, where our trained models often repeated the topic keywords numerous times to achieve a “tongue twister”, rather than learning a deeper representation of semantics.

\paragraph{\textbf{Quality Control}} Quality control on our dataset was performed in multiple ways. Five human evaluators, who are native speakers of English, were provided with 50 sampled instances from the dataset from different conditions to rate the quality of the resulting tongue twisters (25 from Prompt A and 25 from Prompt B). Scores for the criteria were given on a scale from 1 (low quality) to 5 (excellent quality) for 5 criteria: 
(i)~``Twister'' refers to the assessed quality of the tongue twister, and whether it exhibits the expected characteristics of a traditional tongue twister (and is analogous to the "overall" score present in later human evaluation of our model generations in \S\ref{sec:automatic_results_unconstrained}).  (ii) "Topic" refers to how well the input topic phrase is represented in the output via semantics. 
(iii) "Paraphrase Quality" refers to whether or not the paraphrase generation step maintains a meaningful and grammatical text, whilst (iv) "Paraphrase Prosaic" refers to what extent the paraphrase is believed to have successfully removed the sound overlap and tongue twister nature of the original input to more resemble standard text. Finally, (v) "Overall" offers a holistic assessment of the dataset entry as a whole. \autoref{tab:quality_control} presents the breakdown of the human evaluation results. 

Overall, what may be concluded is that the resulting dataset is considered to be of good quality, particularly in regard to the paraphrased versions and overall evaluation, When compared to other aspects, the ratings for the tongue twisters themselves and topic semantics are lower but still indicate reasonably good quality (with 3 being the middle rating, akin to "neither agree nor disagree"). This is particularly true when considering that the tongue twister quality is a highly subjective measure due to entertainment value being a fundamental component that different people perceive to different levels. Additionally, the topic association is lower due to not enforcing that the input terms are present in the output. On the other hand, extracting keywords from twisters and using these as the inputs would lead to an artificially inflated rating for the topic criteria, as the topic would be guaranteed to be represented explicitly in the output. Overall, the dataset samples were given a mean rating of 3.778, akin to a rating very close to "high quality", which is rather good for a creative, and therefore subjective, domain. Further details of human participant recruitment are reported in \S\ref{sec:demographics}.

\begin{table}[t]
\centering \small
\resizebox{1\linewidth}{!}{
\begin{tabular}{ccccc}
\toprule[1pt]
\multicolumn{5}{c}{\textbf{TwistList 2.0 Quality Control}}  \\
\cmidrule(lr){1-5}  
 Twister & Topic & Paraphrase Quality & Paraphrase Prosaic & Overall \\ 
\midrule
3.138** & 3.333* & 4.421* & 3.969** & 3.778** \\  
\bottomrule[1pt]
\end{tabular}
}
\caption{Results of human quality control assessments for TwistList 2.0. Importantly, we only sample from the new machine-generated samples created with TwisterLister for quality control. Assessments regarding the additional 2k+ human-authored examples can be found in the original TwistList work \cite{twistlist}, where 88\% of evaluated tongue twisters were deemed "high quality" (scoring 3 on a 3-point scale). Furthermore, the "Golden" samples evaluated in \S \ref{sec:human_eval_unconstrained} are exclusively from the human-authored portion of TwistList 2.0, due to these constituting the test set. We calculate Fleiss' Kappa for each metric and mark the agreement fair$^*$, moderate$^{**}$, and substantial$^{***}$.}
\label{tab:quality_control}
\end{table}

\subsection{TwistList 2.0 Dataset Summary}

Statistical details of TwistList 2.0 can be seen in \autoref{tab:data_stat}. Additionally, two example entries in the dataset are outlined in \autoref{tab:examples}. As the examples demonstrate, the combination of adjective/adverb and noun used as a topic phrase during the TwistLister generation pipeline can be directly output as part of the final twister, or can instead be represented solely by semantics. For example, in the top example (TT\_ID 68), the input adjective "public" is replicated in the output. The reason for this is due to the chance of the phoneme \textipa{/p/} being selected at random at generation time. Consequently, it makes sense to expect "public" (and morphological variants thereof), to be in the top-k most semantically relevant words to the topic word "public", and therefore constitute part of the constrained candidate vocabulary. The clear selection of the phoneme \textipa{/b/} as the secondary phoneme (selected via minimizing phonemic edit distance) can also be seen, where words such as "broadcaster" and the proper noun "Berman" have been selected (where \textipa{/p/} and \textipa{/b/} differ only in the presence/absence of voicing). Similarly, in the second example (generated from the input "direct language") we see "non-direct" in the output due to the selected initial or secondary phoneme being \textipa{/n/} or \textipa{/m/} (alveolar and bilabial nasal consonants, respectively), whilst "language" has been referenced less directly via terms such as "monolingual", "novelistic" and "Mandarin".
The impact of the different prompt forms can also be seen, with the first example (using Prompt A) being much more succinct than the second example (which used Prompt B). Finally, regarding the paraphrasing used to enable training of style-transfer models, it can be seen that GPT-3.5-Turbo maintains the grammaticality and coherence of the original tongue twister, but replaces much of the vocabulary. However, some terms remain for semantic reasons, such as "BBC" in the first example and "non-verbal" in the second example.

Due to exercising limited control over the paraphrasing stage, some synonym replacements may result in maintaining similar levels of sound overlap (at least in the word-initial alliterative sense). For instance, "pollster" has been replaced by "poller" in the first example, and "methods" has been replaced by "means" in the bottom example. The reason for the former example is similar to why the input phrase \textit{may} be in the output of the final tongue twister as the closest synonym to many words is a derivative of the word itself (i.e., variants of "poll-").

\begin{table}[ht]
\scriptsize
\centering \small
\resizebox{0.99\linewidth}{!}{
\begin{tabular}{l|ccccccc}
\toprule
\textbf{Dataset} & \textbf{Train (A)} & \textbf{Train (B)} & \textbf{Full Train}  & \textbf{Test} & \textbf{Total}\\
\midrule
\textbf{\# Tongue Twisters} & 9653 & 5471 & 15124 & 2124 & 17248 \\
\textbf{\# Vocabulary Size} & 60306 & 61160 & 93517 & 10336 & 98703 \\
\textbf{Avg. \# Topic Words} & 2.00 & 2.00 & 2.00 & 3.16 & 2.14 \\
\textbf{Avg. \# Paraphrase Words} & 26.63 & 57.51 & 37.80 & 17.19 & 35.26 \\
\textbf{Avg. \# Tongue Twister Words} & 23.90 & 53.60 & 34.64 & 14.98 & 32.22 \\

\bottomrule
\end{tabular}
}
\caption{The statistics of TwistList 2.0. The validation set is obtained by randomly selecting the same volume of data as the test set from the shuffled training set. All of the test set input phrases were extracted from TwistList \textit{1.0} entries using RAKE \cite{rake}, rather than being topics we generated using part-of-speech tags. We do this in order to test our models on more robust input forms (where input length is not static), therefore facilitating more possible input combinations than adjective/adverb + noun would allow. This additionally allows comparison to human-authored gold standards in our later testing. \textit{Train (A)} and \textit{Train (B)} refer to samples generated with Prompt A and Prompt B, respectively.}
\label{tab:data_stat}
\end{table}

\begin{table}[htp]
\centering\small
\resizebox{\linewidth}{!}{
\begin{tabular}{p{0.2\linewidth}|p{0.80\linewidth}}
\toprule
\textbf{TT ID:} & 68  \\
\midrule
\textbf{Topic:} &  "public commentator" \\
\hdashline[3pt/5pt]
\textbf{Source:} & GPT-3.5-Turbo \\
\hdashline[3pt/5pt]
\textbf{Prompt:} & A \\
\hdashline[3pt/5pt]
\textbf{Tongue Twister:} & The public-spirited BBC broadcasts presented by persistent presenters perplexed the publicist Berman the broadcaster and the pollster profoundly.
\\
\hdashline[3pt/5pt]
\textbf{Paraphrase:} & The civic-minded BBC airs hosted by determined hosts confused the publicist Berman the broadcaster and the poller deeply.\\
\hdashline[3pt/5pt]

\textbf{Twister ARPABET:} & DH AH0   P AH1 B L IH0 K S P IH1 R IH0 T AH0 D   B IY2 B IY0 S IY1   B R AO1 D K AE2 S T S   P R IY0 Z EH1 N T AH0 D   B AY1   P ER0 S IH1 S T AH0 N T   P R EH1 Z AH0 N T ER0 Z   P ER0 P L EH1 K S T   DH AH0   P AH1 B L IH0 S IH0 S T   B ER1 M AH0 N   DH AH0   B R AO1 D K AE2 S T ER0   AH0 N D   DH AH0   P OW1 L S T ER0   P R OW0 F AW1 N D L IY0.
\\
\hdashline[3pt/5pt]
\textbf{Paraphrase ARPABET:} & DH AH0   S IH1 V Z IH0 K EY2 D M IH0 D   B IY2 B IY0 S IY1   EH1 R Z   HH OW1 S T IH0 D   B AY1   D IH0 T ER1 M AH0 N D   HH OW1 S T S   K AH0 N F Y UW1 Z D   DH AH0   P AH1 B L IH0 S IH0 S T   B ER1 M AH0 N   DH AH0   B R AO1 D K AE2 S T ER0   AH0 N D   DH AH0   P AA1 L ER0   D IY1 P L IY0.
\\
\midrule
\textbf{TT ID:} & 10397  \\
\midrule
\textbf{Topic:} &  "direct language" \\
\hdashline[3pt/5pt]
\textbf{Source:} & GPT-3.5-Turbo \\
\hdashline[3pt/5pt]
\textbf{Prompt:} & B \\
\hdashline[3pt/5pt]
\textbf{Tongue Twister:} & Non-direct, multilingual, and non-verbal, Monolingual speakers were nominal. Some could speak Mandarin natively, While others relied on mediated means. The novelistic twist began when mispronouncing, Novellus words of complex morphology. Non-native speakers struggled with Macromedia, And nonnatives found Mandarin challenging. Nonaccrual in Marathi, they preferred, As net learners of a morphological world. But despite the hurdles, they persevered, Finding beauty in languages unfurled.
\\
\hdashline[3pt/5pt]
\textbf{Paraphrase:} & Indirect, polyglot, and non-verbal, Single-language speakers were nominal. Some could converse in Mandarin from birth, While others depended on mediated methods. The narrative turn commenced when mispronouncing, Novellus terms of intricate structure. Non-indigenous speakers struggled with Macromedia, And non-natives found Mandarin demanding. Non-accumulation in Marathi, they favored, As online learners of a structural world. But despite the obstacles, they persisted, Discovering elegance in languages unfolded.
\\
\bottomrule
\end{tabular}
}
\caption{Example entries from TwistList 2.0. All entries also contain IPA transcriptions used by converting ARPABET directly into IPA. Consequently, some minor pronunciation errors occur as a result of an imperfect mapping between the two standards and the selected accent of the G2P tools used.  We forego presenting the transcriptions for the second example in the interest of space. In the dataset, phonemes that constitute a single word in ARPABET are separated with a space, whilst separate words are delineated via double spaces. IPA transcriptions contain no separators within words (due to being a 1:1 symbol:sound mapping), with a single space between words.}
\label{tab:examples}
\end{table}

\section{Topic-to-Twister and Style-Transfer Tongue Twister Generation}\label{sec:unconstrained}
\label{methodology}

\subsection{Task Definition}
\label{task-definition}

Inspired by \citet{keh-etal-2023-pancetta}, we define two different settings for the task of tongue twister generation we describe as \textit{topic-to-twister} and \textit{style-transfer}. In the former topic-to-twister setting, for a given topic phrase (as generated by randomly sampling adjective and noun combinations, as described in \S \ref{sec:twisterlister_pipe}), we aim to generate a tongue twister $T$, whereby $T$ comprises a sequence of words $ \{w_1, w_2,... w_{n}\}$.
The generated output must satisfy the following constraints: (1) the output should be semantically related to the input topic phrase; (2) the output should show maximal levels of phonological overlap across tokens; and (3) the output should be grammatically valid. On the other hand, for the style-transfer setting we aim to generate a tongue twister again, $T = \{w_1, w_2,... w_{n}\}$, but provide as input a non-tongue twister phrase and aim to convert it via style-transfer into a tongue twister through learning to replace vocabulary with more phonemically similar entries.

\subsection{Trained Models}
\label{models}

In order to realize our goals of tongue twister generation in topic-to-twister and style-transfer settings, we fine-tune a range of popular language models with varying parameter counts on our TwistList 2.0 dataset. 
\begin{itemize}
    \item \textbf{GPT-2} (117M) \cite{GPT-2} - A popular transformer-based text generation model consisting of both an encoder and a decoder.
    \item \textbf{DialoGPT} (117M) \cite{dialogpt} - A version of GPT-2 that has been fine-tuned on an extensive corpus of dialogues to enable better conversational performance (and therefore consequently often better understands natural language task prompts).
    \item \textbf{BART} (139M) \cite{bart} - A popular denoising autoencoder model consisting of a BERT-like encoder \cite{bert} with a GPT-2-like decoder \cite{GPT-2}.
    \item \textbf{Flan-T5} (250M) \cite{Flan-T5} - A further instruction fine-tuned version of the T5 model \cite{T5}.
    \item \textbf{ByT5} (582M) \cite{xue-etal-2022-byt5} - A version of T5 trained with byte/character level tokenization, rather than subwords.
    \item \textbf{Baichuan} (7B) \cite{baichuan} - A large-scale open-source LLM trained on English and Mandarin, achieving SoTA performance on many tasks for a model of its size. Due to the significant size of Baichuan, we perform fine-tuning with the help of Low-Rank Adaptation (LoRA) \cite{LoRA}.
    \item \textbf{ChatGPT} (GPT-3.5-Turbo) \cite{RLHF-openai} - A large language model fine-tuned for chat-based interactions and instruction following, which excels in few and zero-shot tasks.~Importantly, we use ChatGPT in a zero-shot manner and do not perform any fine-tuning.\footnote{GPT-3.5-Turbo was accessed for this purpose during August-September 2023 via the Chat Completions API.}
\end{itemize}

\paragraph{\textbf{Training Splits}}
In order to investigate the benefit of access to different amounts of training data for a data-driven approach to tongue twister generation in topic-to-twister and style-transfer task settings (and to motivate the contribution of our large dataset), we train all of the above models on training sets of various sizes, including 2k, 4k, 8k, and 13k training samples from TwistList 2.0. One exception here is Baichuan, which due to compute requirements we train only on the largest 13k split. We keep the test set (for automatic evaluation) to 2,124 samples, covering the entirety of TwistList 1.0, and have a validation set of equal size. Due to using TwistList 1.0 entries as our test set, all reference-based metrics are compared to human-authored outputs, and human evaluation scores can be directly compared to human performance.

\paragraph{\textbf{Hyperparameters and Training Details}}
To leverage pre-trained parameters, we restore the encoder, decoder, and embedding layers from public checkpoints: (1) \textbf{GPT-2}: gpt2-base \url{https://huggingface.co/gpt2}; (3) \textbf{DialoGPT}: DialoGPT-medium \url{https://huggingface.co/microsoft/DialoGPT-medium}; (4) \textbf{BART}: bart-base \url{https://huggingface.co/facebook/bart-base};  (5) \textbf{Flan-T5}: flan-t5-base \url{https://huggingface.co/google/flan-t5-base}; (6) \textbf{ByT5}: byt5-base \url{https://huggingface.co/google/byt5-base}. The checkpoints of Baichuan are the exception, which require downloading the weights directly from their repositories (\url{https://github.com/baichuan-inc/Baichuan2}). Here we use the 7B checkpoint of Baichuan 2. As the vanilla Baichuan model requires extremely large computing resources, we implement the LoRA~\cite{hu2021lora} technique to reduce computational costs. LoRA adapts large language models by incorporating low-rank modifications. This approach involves adjusting or fine-tuning extensive language models to address specific tasks or domains with a reduced computational cost. By introducing low-rank modifications, LoRA aims to enhance the adaptability and efficiency of these models while maintaining their performance. This technique is particularly beneficial for tailoring pre-trained language models to better suit specialized or narrower domains without requiring excessive computational resources. Our use of ChatGPT consists of GPT-3.5-Turbo via the OpenAI API, where we provide no information other than the same prompt we use with all other models. All other settings remain at default.
 
Our local experiments are carried out on a single Nvidia A40 GPU, which has 48GB of VRAM. When training neural models, we implement the PyTorch Lightning framework to set up training processes. The training parameters are as follows: The \textit{batch size} is set to 16 (excl. Baichuan w/LoRA, where it is 8); the \textit{learning rate} is 1e-4; \textit{max source length} is set to 512, and the \textit{max target length} is set to 100; the optimizer uses Adam \cite{kingma2014adam}, and the $\epsilon$ of Adam is set to 1e-8. The whole training process lasts for 10 \textit{epochs}, and the validation checking runs every half epoch. However, the presented results only consider the checkpoint with the best performance (i.e., lowest loss). 

For all topic-to-twister settings, we use the prompt \textit{`Generate a tongue twister on the topic of "[TOPIC]"'} where \textit{[TOPIC]} refers to the input topic phrase from the test set, and for all style-transfer settings we use the prompt \textit{"Generate a tongue twister by rewriting the following text: [PARAPHRASE]"} where\textit{ [PARAPHRASE]} refers to the non-literary paraphrase of a tongue twister from the test set.

\subsection{Automatic Metric Suite}
\label{automatic-eval}
We present extensive automatic evaluation on the following metrics: \textbf{Perplexity} (\textbf{PPL}), \textbf{BLEU} (\textbf{B-1/B-2/B-3/B-4}) \cite{bleu}, \textbf{ROUGE} (\textbf{Ro-1/Ro-2/Ro-L}) \cite{rouge}, and \textbf{BERTScore} Precision, Recall, and F-Measure \cite{bert-score} (\textbf{BS-P/BS-R/BS-F}). PPL, BLEU, and ROUGE are standard metrics in language generation to assess quality, whilst BERTScore assesses semantic similarity to a gold reference. It should be noted that due to the nature of our task, many potential “gold standard” tongue twisters exist for any given input. Consequently, as with all creative generation works, these reference-based metrics should be interpreted cautiously whilst being aware of their limitations (although we opt to include them for completeness).

\paragraph{\textbf{Readability}}
Tongue twisters are known for their intricate phoneme-level patterns and linguistic complexity, making them challenging to articulate correctly. Consequently, readability metrics can be used to indirectly measure whether or not tongue twisters, due to their complex nature, are using more complex vocabulary in order to meet strict phonemic constraints (such as when a selected phoneme only has a few obscure words that are related to the input topic). We present a range of readability metrics, incorporating the \textbf{Dale-Chall Readability Index} \cite{chall1995readability} (\textbf{Re-D}), \textbf{Flesch–Kincaid Readability Score} \cite{Flesh1948readability} (\textbf{Re-F}), \textbf{Gunning-Fog Index} \cite{gunning-fog} (\textbf{Re-G}) and \textbf{ARI Index} \cite{ARI_index} (\textbf{Re-A}). These comprise a series of complexity and readability metrics that relate to the necessary comprehension level of a text’s audience. They calculate a numerical score based on factors such as sentence length and word complexity, offering a quantitative measure of how difficult a text is to understand. These further complement our other metrics by analyzing linguistic complexity via means other than phonology.\footnote{All readability metrics were implemented from \url{https://pypi.org/project/py-readability-metrics/}.} Additionally, extremely high scores on such metrics can likewise indicate the nonsensical nature of any particularly poor generations. This is because readability metrics take into account factors such as sentence length and syllable counts, with high scores being given to indefinitely long sequences (due to the absence of sentence-final punctuation) as well as convoluted and overly complex syntactic structures and lexical choices.

\paragraph{\textbf{Phonology/Phonetics}}
\label{phonetic_metrics}
We further develop our tongue twister measures and present phonemic edit distance metrics (PEDs) relying on the \textit{weighted phonemic edit distance} function from the \textit{PanPhon} package \cite{mortensen-etal-2016-panphon}. This allows us to more directly analyze the phonetics of our generated outputs by taking into consideration the articulatory similarities and differences between different phonemes. For example, previous phonetic metrics from \citet{twistlist}, PO and Init-PO, treat all phonemes as equidistant in feature space, resulting in a transition from \textipa{/s/} to /\textipa{S}/ being viewed as the same "quality” as a transition from \textipa{/s/} to \textipa{/g/}. In the former case, we have transitioned from a voiceless alveolar fricative to a voiceless \textit{post-}alveolar fricative (therefore moving the tongue slightly further back in the mouth), whilst in the latter case the transition is between a voiceless alveolar fricative and a voiced velar plosive, resulting in a change of voicing, position, and manner of articulation. Consequently, using phonemic edit distance allows us to not punish transitions between phonemically similar sounds (which are more likely to encourage mispronunciations, such as in “she sells sea shells”) as heavily as we punish transitions between unrelated sounds. As with PO and Init-PO, we utilize this weighted edit distance on both a word-initial and overall level, with \textbf{oPED} taking the \textit{overall} average edit distance between every phoneme transition in the tongue twister, whilst \textbf{iPED} calculates the edit distance between word-\textit{initial} phonemes (and is, therefore, a more accurate measure of “soft” alliteration). 

Formally, for iPED and oPED let \textit{X} and \textit{Y} be two phonemes, represented as feature vectors, where each vector contains binary values representing the presence or absence of a phonological feature. Therefore, the weighted feature edit distance between \textit{X} and \textit{Y} is defined as: $D_{\text{wfe}}(X,Y) = \min_{\text{alignments}} \sum_{i=1}^n w_i \cdot d_i$, where \(X\) and \(Y\) are sequences of phonological features represented as feature vectors, \(n\) is the length of the longest common subsequence of \(X\) and \(Y\), \(d_i\) represents the distance between the \(i\)-th feature in the alignment, and \(w_i\) is the weight assigned to the \(i\)-th feature. The minimum is taken over all possible alignments of the sequences \(X\) and \(Y\). Each alignment assigns a distance \(d_i\) between corresponding features in the sequences. The weighted sum of these distances is computed, where each distance is multiplied by its corresponding weight \(w_i\). For iPED, we calculate the mean distance across sequences of word-initial phonemes, whilst for oPED we calculate the mean distance when comparing all adjacent phonemes across the tongue twister. This metric provides a flexible way to measure the similarity between sequences of phonological features, allowing for customization through feature weighting. In the case of oPED and iPED, the lower the score, the better (with a score of 0 relating to 100\% overlap of a single phoneme throughout the tongue twister).

In addition to our novel metrics discussed above (iPED/oPED), we additionally present our two metrics from \citet{twistlist}, Phoneme Overlap (\textbf{PO}) and Initial Phoneme Overlap (\textbf{Init-PO}). \textbf{PO} refers to the average overlap of all phonemes across tokens (\#~unique phonemes / \#~total phonemes), whereas \textbf{Init-PO} is the ratio of unique word-initial phonemes to the number of words (\#~unique word-initial phonemes / \#~words).  Whilst these original phoneme-based metrics reward longer outputs, we argue that all other things equal, a longer tongue twister is better than a shorter one as it provides more entertainment and more opportunities for mispronunciation. Perfect scores on PO and Init-PO can be achieved by the repetition of a single word. Whilst this does not lead to high-quality outputs, these metrics are intended exclusively to be indicators of sound characteristics, rather than an overall guide to quality. In both cases, higher levels of overlap result in lower ("better") scores, and the highest ("worst") achievable score is 1.

\subsection{Human Evaluation Protocol}\label{sec:human_eval_process}
Due to the limitations of automatic evaluation metrics and tongue twisters being a creative domain where articulation abilities are tested, we also perform human evaluation. In line with \citet{loakman-etal-2023-iron}, we aim to be transparent in our human evaluation of a subjective language type that may be considered a type of humorous language. In total, five evaluators were asked to rate 20 outputs from the best performing standard baselines, Flan-T5 and ByT5, in addition to Baichuan, ChatGPT (i.e., GPT-3.5-Turbo), and golden examples from \textbf{TwistList 2.0} on the following criteria: \textbf{Relevance} (how relevant the tongue twister is given the keyword inputs), \textbf{Fluency} (how grammatically valid the output is), \textbf{Difficulty of Articulation} (how difficult a tongue twister is to say), \textbf{Coherence} (how much sense the output makes), and \textbf{Entertainment Value} (how entertaining the output is, considering sounds and semantics) and a holistic \textbf{Overall} criteria. All ratings were given on a 5-point scale where 1 equates to "poor" and 5 equates to "excellent". Importantly, we include both Flan-T5 and ByT5 trained on 2k samples, as well as 13k samples, to investigate whether or not increased training data has a measurable impact on human evaluation in addition to the patterns observed in automatic evaluation. 

\paragraph{\textbf{Evaluator Recruitment and Demographics}}
\label{sec:demographics}
In total, we recruit 5 evaluators via internal notices and word of mouth. All evaluators have university-level education to a minimum of undergraduate level in a range of fields including linguistics, computer science, engineering, and animation, and therefore represent a wide range of academic backgrounds. All evaluators are native speakers of English and report no language processing issues. Evaluators were provided with a 55GBP Amazon gift card for their combined work on dataset quality control and output evaluation, totalling approximately 4hrs of work.

\paragraph{\textbf{Materials Provided to Evaluators}}
Evaluation was performed on an online platform. Participants were provided with a page of detailed instructions on how to navigate the platform and the order in which to perform evaluation tasks. Importantly, all human evaluation responses were given on a 1-5 rating scale (where 1 equates to "poor" and 5 equates to "excellent" for how well a criterion is met). For example "\textit{The tongue twister can be considered logically and semantically coherent.}" for the criteria of Coherence. Additionally, those with non-linguistic backgrounds had certain terms clarified, such as the meaning of "prosaic" in the style-transfer task. Each evaluator rated 450 samples across all evaluations, consisting of 20 samples from 7 models (i.e., Flan-T5$_{\text{2k}}$, Flan-T5$_{\text{13k}}$, ByT5$_{\text{2k}}$, ByT5$_{\text{13k}}$, Baichuan, ChatGPT (i.e., GPT-3.5-Turbo), and the gold-standard) on 2 tasks (280 in total), 50 quality control examples, and 120 examples for our novel constrained decoding algorithm (20 of which used constrained base GPT-2, 20 of which came from our constrained fine-tuned GPT-2, and 20 of which came from unconstrained GPT-2 trained on 13k samples, with the equivalent set up for Baichuan).

\section{Results (Topic-to-Twister and Style-Transfer)}

\subsection{Automatic Results}
\label{sec:automatic_results_unconstrained}
\begin{table*}[ht]
\centering \small
\resizebox{0.95\linewidth}{!}{
\begin{tabular}{l|c|cccc|ccc|c}
\toprule
\textbf{Model} & \textbf{PPL} & \textbf{B-1$\uparrow$} & \textbf{B-2$\uparrow$} & \textbf{B-3$\uparrow$}  & \textbf{B-4$\uparrow$}  & \textbf{Ro-1$\uparrow$} & \textbf{Ro-2$\uparrow$} & \textbf{Ro-L$\uparrow$} & \textbf{Length} \\
\midrule
\textbf{GPT-2$_\text{2k}$} & 25.04 & 0.0391 & 0.0165 & 0.0083 & 0.0043 & 5.1255 & 0.6704 & 4.8414 & 51.81 \\
\textbf{DialoGPT$_\text{2k}$} & 23.74 & 0.0453 & 0.0187 & 0.0091 & 0.0047 & 6.3366 & \underline{1.0349} & 5.9843 & 57.72 \\
\textbf{BART$_\text{2k}$} & 3.98 & 0.0749 & 0.0256 & 0.0105 & 0.0040 & \underline{8.1503} & 0.7608 & \underline{7.4855} & 44.07 \\ 
\textbf{Flan-T5$_\text{2k}$} & 2.86 & \underline{0.1055} & \underline{0.0545} & \underline{0.0325} & \underline{0.0209} & \textbf{11.2417} & \textbf{2.1528} & \textbf{10.3641} & 25.60 \\
\textbf{ByT5$_\text{2k}$} & - & \textbf{0.1440} & \textbf{0.0918} & \textbf{0.0639} & \textbf{0.0477} & 6.2582 & 1.0329 & 5.9611 & 47.56 \\ 
\hdashline[3pt/5pt]
\textbf{GPT-2$_\text{4k}$} & 24.84 & 0.0423 & 0.0168 & 0.0076 & 0.0037 & 5.7569 & 0.6818 & 5.5077 & 53.27 \\
\textbf{DialoGPT$_\text{4k}$} & 24.83 & 0.0422 & 0.0161 & 0.0071 & 0.0032 & 5.9712 & 0.6915 & 5.6415 & 55.91 \\
\textbf{BART$_\text{4k}$} & 4.09 & 0.0685 & 0.0215 & 0.0082 & 0.0031 & \underline{7.6023} & \underline{0.7364} & \underline{6.9026} & 47.30 \\ 
\textbf{Flan-T5$_\text{4k}$} & 2.93 & \underline{0.1090} & \underline{0.0558} & \underline{0.0328} & \underline{0.0207} & \textbf{10.5343} & \textbf{1.7701} & \textbf{9.8546 }& 24.25 \\
\textbf{ByT5$_\text{4k}$} & - & \textbf{0.1397} & \textbf{0.0900} & \textbf{0.0634} & \textbf{0.0479} & 6.8637 & 0.9744 & 6.4977 & 48.18 \\ 
\hdashline[3pt/5pt]
\textbf{GPT-2$_\text{8k}$} & 28.58 & 0.0430 & 0.0143 & 0.0060 & 0.0026 & 5.6638 & 0.5385 & 5.2935 & 54.23 \\
\textbf{DialoGPT$_\text{8k}$} & 24.54 & 0.0458 & 0.0172 & 0.0076 & 0.0035 & 6.3555 & \underline{0.7622} & 5.9901 & 56.59 \\
\textbf{BART$_\text{8k}$}  & 4.33 & 0.0633 & 0.0195 & 0.0072 & 0.0029 & \underline{6.8434} & 0.5874 & \underline{6.1127} & 49.68 \\ 
\textbf{Flan-T5$_\text{8k}$} & 3.11 & \underline{0.1194} & \underline{0.0600} &\underline{ 0.0346} & \underline{0.0217} & \textbf{10.7728} & \textbf{1.3758} & \textbf{10.0456} & 22.97 \\
\textbf{ByT5$_\text{8k}$} & - & \textbf{0.1531} & \textbf{0.0951} & \textbf{0.0647} & \textbf{0.0475} & 6.2602 & 0.7142 & 6.0271 & 40.96 \\
\hdashline[3pt/5pt]
\textbf{GPT-2$_\text{13K}$} & 28.77 & 0.0440 & 0.0152 & 0.0061 & 0.0025 & 5.9412 & 0.6024 & 5.5890 & 56.59 \\
\textbf{DialoGPT$_\text{13K}$} & 25.15 & 0.0504 & 0.0187 & 0.0078 & 0.0034 & \underline{6.7274} & 0.8170 & \underline{6.3148} & 60.05 \\
\textbf{BART$_\text{13K}$} & 4.16 & 0.0595 & 0.0152 & 0.0050 & 0.0017 & 6.6284 & 0.5090 & 5.9523 & 49.79 \\
\textbf{Flan-T5$_\text{15K}$} &  3.10 & \underline{0.1189} & \underline{0.0571} & \underline{0.0319} & \underline{0.0192} & \textbf{10.0983} & \underline{1.1632} & \textbf{9.2976} & 23.87 \\
\textbf{ByT5$_\text{16k}$} & - & \textbf{0.1609*} & \textbf{0.0988} & \textbf{0.0660} & \textbf{0.0476} & 6.4857 &  0.6073 & 6.2272 & 40.05  \\
\textbf{Baichuan+LoRA$_\text{13K}$} & 14.54 & 0.0463 & 0.0227 & 0.0131 & 0.0080 & 6.2215 & \textbf{1.3878} & 5.9212 & 51.76 \\ 
\midrule
\textbf{ChatGPT} & - & 0.1577 & 0.1073* & 0.0788* & 0.0585* & 26.2949* & 13.2789* & 23.7763* & 24.46 \\ 
\bottomrule
\end{tabular}
}
\caption{\label{tab:auto-evaluation-topic}
Results of automatic evaluation on typical metrics for the topic-to-twister task setting. The mean length of the tongue twisters in the dataset is 35.65/34.14/14.98 words for train/evaluation/test, respectively.
Results in \textbf{bold} represent the best performance for a given training data quantity, and \underline{underlined} presents the second-best performance. Results followed by an asterisk * denote the overall best performance. The PPL of ByT5 is not applicable, as the tokens of ByT5 are characters whilst others are sub-words.}
\end{table*}

\begin{table*}[ht]
\centering \small
\resizebox{0.99\linewidth}{!}{
\begin{tabular}{l|ccc|cccc|cccc}
\toprule
\textbf{Model} & \textbf{BS-P$\uparrow$} & \textbf{BS-R$\uparrow$} & \textbf{BS-F1$\uparrow$} & \textbf{IPO$\downarrow$}  & \textbf{PO$\downarrow$}  & \textbf{iPED$\downarrow$} & \textbf{oPED$\downarrow$} & \textbf{Re-D} & \textbf{Re-F} & \textbf{Re-G} & \textbf{Re-A}\\
\midrule
\textbf{GPT-2$_\text{2k}$} & 0.7543 & 0.8265 & 0.7883 & \textbf{0.0849} & \textbf{0.0617*} & 3.8717 & 5.9131 & 17.08 & 45.79 & 48.71 & 59.52 \\
\textbf{DialoGPT$_\text{2k}$} &  0.7680 & 0.8310 & 0.7978 & \underline{0.0954} & \underline{0.0660} &  \underline{3.8308} &  \underline{5.9032} &  12.79 & 16.52 & 18.54 & 21.21  \\
\textbf{BART$_\text{2k}$} & \underline{0.7938} & \underline{0.8384} & \underline{0.8153} & 0.3183 & 0.1848 &  4.5168 & 5.9219 & 14.21 &  11.70 & 14.12 & 11.51 \\ 
\textbf{Flan-T5$_\text{2k}$} & \textbf{0.8075} & \textbf{0.8440} & \textbf{0.8249} &  0.2129 & 0.1610 & 4.0828 & 5.9141 & 13.10 & 14.68 & 16.19 & 18.25 \\
\textbf{ByT5$_\text{2k}$} & 0.7705 & 0.8301 & 0.7982 & 0.1207 & 0.1021 & \textbf{2.5398} & \textbf{5.9048} & 14.61 & 28.54 & 30.20 & 38.35 \\
\hdashline[3pt/5pt]
\textbf{GPT-2$_\text{4k}$} &  0.7648 & 0.8303 & 0.7958 & \textbf{0.0881} &  \textbf{0.0683} & 4.1320 & 5.9440 & 13.51 & 23.08 & 25.14 &   28.89 \\
\textbf{DialoGPT$_\text{4k}$} & 0.7648 & 0.8300 & 0.7956 & \underline{0.0984} & \underline{0.0701} & \underline{3.7237} &  5.9205 & 13.01 & 17.39 & 19.26 & 21.86 \\
\textbf{BART$_\text{4k}$} & \underline{0.7962} & \underline{0.8363} & \underline{0.8156} & 0.2821 & 0.1620 & 4.7537 & 5.9693 & 12.77 & 12.46 & 14.11 & 13.69 \\ 
\textbf{Flan-T5$_\text{4k}$} & \textbf{0.8071} & \textbf{0.8436} & \textbf{0.8244} &  0.2089 & 0.1657 & 3.8280 & \textbf{5.8899} & 13.29 & 16.63 &  17.76 & 20.28 \\
\textbf{ByT5$_\text{4k}$} & 0.7636 & 0.8297 & 0.7941 & 0.1250 & 0.0987 & \textbf{2.2248} & \underline{5.9197} & 16.0915 & 35.6761 & 36.9863 & 47.4017 \\ 
\hdashline[3pt/5pt]
\textbf{GPT-2$_\text{8k}$} & 0.7712 & 0.8315 & 0.7998 & 0.1224 & 0.0841 & 4.3355 & 5.9751 & 12.57 & 14.93 & 16.30 & 18.00  \\
\textbf{DialoGPT$_\text{8k}$} & 0.7745 & 0.8327 &  0.8022 & \textbf{0.1116} & \textbf{0.0775} & \underline{4.0111} &  5.8952 & 11.99 & 13.42 & 15.14 & 16.83   \\
\textbf{BART$_\text{8k}$} & \underline{0.7928} & \underline{0.8343} & \underline{0.8129} & 0.2831 & 0.1675 & 4.9463 & 5.9693   & 11.97 & 12.36 & 13.71 & 13.86  \\ 
\textbf{Flan-T5$_\text{8k}$} & \textbf{0.8160} & \textbf{0.8462} & \textbf{0.8304} & 0.2378 & 0.1866 & 4.1542 & \textbf{5.8669} & 12.74 & 14.40 & 16.24 & 17.31  \\
\textbf{ByT5$_\text{8k}$} & 0.7711 & 0.8306 & 0.7986 & 0.1341 & 0.1087 & \textbf{2.1024*} & \underline{5.8804} & 16.37 & 31.98 & 33.88 & 42.47\\
\hdashline[3pt/5pt]
\textbf{GPT-2$_\text{13K}$} &  0.7748 & 0.8322 & 0.8021 & 0.1261 & 0.0859 & \underline{4.2671} & 5.8985 & 12.06 & 13.46 & 15.21 & 16.64  \\
\textbf{DialoGPT$_\text{13K}$} &  0.7795 & \underline{0.8336} & 0.8053 & \underline{0.1249} & \underline{0.0839} &  4.4665 & 5.9407 & 11.30 & 12.85 &  14.95 & 15.55 \\
\textbf{BART$_\text{13K}$} & \underline{0.7925} & 0.8319 & \underline{0.8115} & 0.2662 & 0.1489 & 4.8871 &  \textbf{5.8184*} & 12.14 & 11.75 & 13.57 & 13.83 \\ 
\textbf{Flan-T5$_\text{13K}$} & \textbf{0.8181} & \textbf{0.8468} & \textbf{0.8319} & 0.2568 & 0.1996 & 4.3122 & 5.8932 & 12.31 & 12.50 & 14.03 & 14.97  \\
\textbf{ByT5$_\text{13k}$} & 0.7742 & 0.8320 & 0.8009 & 0.1554 & 0.1217 & \textbf{2.5520} &  \underline{5.8374} & 15.77 & 31.22 & 34.00 & 39.33 \\
\textbf{Baichuan+LoRA$_\text{13K}$} & 0.7658 & 0.8259 & 0.7940 & \textbf{0.0689*} & \textbf{0.0657} & \textbf{2.6752} &  5.9951 & 16.41 & 34.37 & 31.73 & 44.59 \\ 
\midrule
\textbf{ChatGPT} & 0.8401* & 0.8613* & 0.8503* & 0.3477 & 0.2991 & 4.0547 & 5.9153 & 9.83 & 9.37 & 10.53 & 11.48 \\ 
\midrule
\textbf{Brown Corpus Prose} & - & - & - & 0.4870 & 0.2275 & 5.1431 & 5.9561 & 11.09 & 13.15 & 15.56 & 16.60 \\
\bottomrule
\end{tabular}
}
\caption{\label{tab:additional-evaluation-topic}
 Results of automatic evaluation on tongue twister related metrics for the topic-to-twister setting. Results in \textbf{bold} represent the best performance for a given training data quantity, and \underline{underlined} presents the second-best performance. Results followed by an asterisk * denote the overall best performance. Brown Corpus Prose presents the scores of standard prose from the Brown Corpus from NLTK (across 602 sentences with a minimum length of 25 words).
}
\end{table*}

\paragraph{\textbf{Topic-to-Twister}}
We present the results for automatic evaluation in the topic-to-twister task setting in \autoref{tab:auto-evaluation-topic} and \autoref{tab:additional-evaluation-topic}. Firstly, regarding the reference-based metrics (BLEU, ROUGE, and BERTScore) we see clear performance differences across our chosen models. On average across almost all metrics, we see the performance from worst to best ordered as GPT-2, DialoGPT, BART, ByT5, and Flan-T5 for our fine-tuned models. However, we see that Baichuan's performance is variable, mostly outperforming BART and underperforming Flan-T5 and ByT5. However, Baichuan performs worse than BART considering a range of metrics (B-1, Ro-1, Ro-L, BS-P, BS-R, BS-F1) when considering the same training data amount (13k). Flan-T5 and ByT5 alternate in performance, with the latter tokenizer-free model performing better on BLEU-based metrics, but worse on ROUGE.
When specifically considering changes alongside an increase in training data, in \autoref{tab:auto-evaluation-topic} we see little to no improvement in reference-based metrics within the topic-to-twister setting, with performance decreasing as training data increases in some cases.
However, it is pertinent to mention that reference-based metrics are imperfect for the task of creative language generation due to the one-to-many dilemma prevalent in many NLG tasks \cite{gupta-etal-2019-investigating}, whereby there are numerous potential tongue twisters to generate from any given input topic. This is also particularly true when considering the TwistList 2.0 dataset, where the topics are often only related to the input phrase in a high-level conceptual fashion (as we do not enforce the generation of the topic phrase within the tongue twister itself).
Finally, ChatGPT, when prompted in the same manner as our other models demonstrates significantly higher performance than any of our fine-tuned models.

Regarding the readability and phonemic metrics presented in \autoref{tab:additional-evaluation-topic} we see a range of patterns. Firstly, for IPO (formerly referred to as Init-PO) and PO, we see the GPT-2 based models "outperform" BART and the T5 models almost across the board. However, naive phoneme-overlap-based metrics do not take into account more sophisticated phonemic characteristics. When considering our new metrics based on phonemic edit distance (iPED and oPED) we see less distinction across any of the presented models. However, one notable finding is that Baichuan performs significantly better than any of the other models trained on 13k samples in the word-initial IPO and iPED metrics, suggesting high levels of word-initial phonemic overlap across tokens when compared to other models. Likewise, we see that the tokenizer-free ByT5 model outperforms Flan-T5 by a significant margin regarding the phoneme-based metrics, suggesting that the finer-grained tokenization of ByT5 is preferable for identifying the sound patterns implicitly included in tongue twisters via grapheme combinations. ChatGPT, on the other hand, scores the highest on the IPO and iPED metrics (i.e., worst). This, however, is not in itself indicative of poor tongue twisters, but rather the model's focus on producing high-quality comprehensible and grammatical text, therefore being less likely to fall victim to degenerate patterns of repeating the same word over and over to achieve overlap. As with the reference-based metrics discussed earlier, raw reliance on phoneme-based metrics can also be misleading. For example, intuitively, prior works have demonstrated that high scores (i.e., lower values) can be achieved in word-initial-based metrics simply by repeating the same word, rather than producing a complex and entertaining tongue twister. Regarding readability scores, we see GPT-2 and Baichuan present high scores (i.e., high difficulty of readability) when compared to our other models. However, high readability metrics can be indicative of numerous things, including desirable behavior (using complex structures and sophisticated multi-syllabic vocabulary) as well as behavior that is not necessarily desirable (e.g., producing gratuitously long sentences). It is for this reason that we exclude an indicator of the preferred metric direction in the case of our readability metrics, yet include the scores for completeness.

\begin{table*}[ht]
\centering \small
\resizebox{1\linewidth}{!}{
\begin{tabular}{l|c|cccc|ccc|c}
\toprule
\textbf{Model} & \textbf{PPL} & \textbf{B-1$\uparrow$} & \textbf{B-2$\uparrow$} & \textbf{B-3$\uparrow$}  & \textbf{B-4$\uparrow$}  & \textbf{Ro-1$\uparrow$} & \textbf{Ro-2$\uparrow$} & \textbf{Ro-L$\uparrow$} & \textbf{Length} \\
\midrule
\textbf{GPT-2$_\text{2k}$} & 15.59 &  0.1055 & 0.0738 & 0.0534 & 0.0397 & 15.2637 & 6.2852 & 14.8342 & 62.53    \\
\textbf{DialoGPT$_\text{2k}$} & 14.54 & 0.1001 & 0.0688 & 0.0491 & 0.0359 &  14.4003 & 5.7688 & 13.9574 & 67.65 \\
\textbf{BART$_\text{2k}$} & 1.96 & 0.2498 & 0.1797 & 0.1324 & 0.0997 & 25.4414 & 11.3534 & 24.8061 & 36.87 \\ 
\textbf{Flan-T5$_\text{2k}$} & 1.64 & \underline{0.5597} & \underline{0.4426} & \underline{0.3591} & \underline{0.2964} & \textbf{48.7184} & \textbf{24.3008} & \textbf{47.9246} & 15.18 \\
\textbf{ByT5$_\text{2k}$} & - & \textbf{0.6770} & \textbf{0.5847} & \textbf{0.5210} & \textbf{0.4731} & \underline{46.8991} & \underline{22.3860} & \underline{46.0446} & 16.53  \\
\hdashline[3pt/5pt]
\textbf{GPT-2$_\text{4k}$} & 15.15 & 0.1161 & 0.0837 & 0.0621 & 0.0472 & 16.7069 & 7.4608 &  16.3085 & 62.70 \\
\textbf{DialoGPT$_\text{4k}$} & 13.96 & 0.1099 & 0.0773 & 0.0565 & 0.0425 & 15.8087 & 6.7687 & 15.3450 & 65.72 \\
\textbf{BART$_\text{4k}$} & 2.00 & 0.2561 & 0.1859 & 0.1378 & 0.1041 & 26.3945 & 12.0011 & 25.8563 & 37.72 \\ 
\textbf{Flan-T5$_\text{4k}$} & 1.62 & \underline{0.5500} & \underline{0.4351} & \underline{0.3532} & \underline{0.2917} & \underline{48.8662} & \underline{24.6492} & \underline{48.1269} & 14.99  \\
\textbf{ByT5$_\text{4k}$} & - & \textbf{0.7177} & \textbf{0.6242} & \textbf{0.5596} & \textbf{0.5111} & \textbf{49.1321} &  \textbf{24.7326} & \textbf{48.2396} & 15.90 \\
\hdashline[3pt/5pt]
\textbf{GPT-2$_\text{8k}$} & 14.26 & 0.1152 & 0.0833 & 0.0622 & 0.0476 & 16.7724 & 7.5582 & 16.3746 & 63.30  \\
\textbf{DialoGPT$_\text{8k}$} & 13.19 & 0.1096 & 0.0772 & 0.0572 & 0.0438 & 15.7687 & 7.0081 & 15.248 &  67.10  \\
\textbf{BART$_\text{8k}$} & 1.95 & 0.2448 & 0.1803 & 0.1358 & 0.1043 & 30.1371 & 14.4970 & 29.4553 & 37.50  \\ 
\textbf{Flan-T5$_\text{8k}$} & 1.63 & \underline{0.5699} & \underline{0.4571} & \underline{0.3756} &\underline{0.3135} & \textbf{51.1949} & \textbf{26.9544} & \textbf{50.4617} & 14.91  \\
\textbf{ByT5$_\text{8k}$} & - & \textbf{0.7124} & \textbf{0.6224} & \textbf{0.5605} & \textbf{0.5136} & \underline{50.4737} & \underline{25.9246} & \underline{49.5851} & 15.85 \\
\hdashline[3pt/5pt]
\textbf{GPT-2$_\text{13K}$} &  13.55 & 0.1229 & 0.0901 & 0.0683 &    0.0529 & 17.9146 & 8.4624 & 17.4801 & 60.71 \\
\textbf{DialoGPT$_\text{13K}$} & 12.84 & 0.1114 & 0.0796 & 0.0597 & 0.0462 & 16.2061 &  7.3163 & 15.7076 & 66.90 \\
\textbf{BART$_\text{13K}$} & 1.82 & 0.2298 & 0.1717 & 0.1309 & 0.1014 & 34.0662 & 16.9713 & 33.3075 & 37.78  \\
\textbf{Flan-T5$_\text{13K}$} & 1.61 & 0.5832  &  0.4704 & 0.3885 & 0.3258 & \underline{52.7077} & \underline{28.6604} & \underline{51.8808} & 14.82     \\
\textbf{ByT5$_\text{13k}$} & - & \textbf{0.7356*} & \textbf{0.6465*} & \textbf{0.5846*} & \textbf{0.5376*} & 51.9886 & 27.7015 & 51.1895 & 15.55 \\
\textbf{Baichuan+LoRA$_\text{13K}$} & 5.15 & \underline{0.6046} & \underline{0.5006} & \underline{0.4229} & \underline{0.3618} & \textbf{60.1989*} & \textbf{37.1423*} & \textbf{59.1040*} &  15.27 \\ 
\midrule
\textbf{ChatGPT} & - & 0.3288 & 0.2167 & 0.1491 & 0.1057 & 39.3536 & 13.9654 & 34.9884 & 17.43 \\ 
\bottomrule
\end{tabular}
}
\caption{\label{tab:auto-evaluation-paraphrase}
Results of automatic evaluation on typical reference-based metrics in NLG for the style-transfer setting. The mean length of the tongue twisters in the dataset is 35.65/34.14/14.98 words for train/evaluation/test, respectively. Results in \textbf{bold} represent the best performance for a given training data quantity, and \underline{underlined} presents the second-best performance. Results followed by an asterisk * denote the overall best performance. The PPL of ByT5 is not applicable, as the tokens of ByT5 are characters whilst others are sub-words.
}
\end{table*}

\begin{table*}[ht]
\centering \small
\resizebox{1\linewidth}{!}{
\begin{tabular}{l|ccc|cccc|cccc}
\toprule
\textbf{Model} & \textbf{BS-P$\uparrow$} & \textbf{BS-R$\uparrow$} & \textbf{BS-F1$\uparrow$} & \textbf{IPO$\downarrow$}  & \textbf{PO$\downarrow$}  & \textbf{iPED$\downarrow$} & \textbf{oPED$\downarrow$} & \textbf{Re-D} & \textbf{Re-F} & \textbf{Re-G} & \textbf{Re-A}\\
\midrule
\textbf{GPT-2$_\text{2k}$} & 0.8170 & 0.8662 & 0.8405 & \underline{0.1264} & \underline{0.0830} & 4.7213 & 5.9688 & 10.66 & 7.45 & 9.52 & 6.94 \\
\textbf{DialoGPT$_\text{2k}$} &  0.8172 & 0.8634 & 0.8394 & \textbf{0.1177*} & \textbf{0.0786*} & 4.8873 & \underline{5.9570} & 9.92 & 7.66 &  9.76 & 7.74 \\
\textbf{BART$_\text{2k}$} & 0.8162 & 0.9031 & 0.8570 & 0.5112 & 0.2474 & \underline{4.5594} & \textbf{5.8120} & 14.76 & 12.21 & 14.13 & 11.47  \\ 
\textbf{Flan-T5$_\text{2k}$} &  \textbf{0.9281} & \textbf{0.9275} & \textbf{0.9277} & 0.5465 & 0.4290 & 4.7362 & 5.9696 & 10.51 & 6.08 & 7.77 & 5.07 \\
\textbf{ByT5$_\text{2k}$} & \underline{0.9166} & \underline{0.9171} & \underline{0.9166} & 0.4392 & 0.3949 & \textbf{4.2451} & 5.9703 & 10.78 & 6.81 & 8.72 & 5.90 \\
\hdashline[3pt/5pt]
\textbf{GPT-2$_\text{4k}$} & 0.8247 & 0.8726 & 0.8477 & \textbf{0.1221} & \textbf{0.0834} & 4.7674 & 5.9775 & 10.88 & 7.02 &  8.93 & 6.15 \\
\textbf{DialoGPT$_\text{4k}$} & 0.8213 & 0.8668 & 0.8432 & \underline{0.1280} & \underline{0.0847} & 4.8948 & 5.9825 & 10.34 &  7.83 & 10.16 & 7.35  \\
\textbf{BART$_\text{4k}$} & 0.8157 &  0.9026 & 0.8565 &  0.4028 & 0.2267 & \textbf{3.8435} & \textbf{5.7562*} & 16.74 & 14.10 & 15.72 & 14.19 \\ 
\textbf{Flan-T5$_\text{4k}$} & \textbf{0.9262} & \textbf{0.9266} & \textbf{0.9263} &  0.5362 & 0.4290 & 4.6793 & 5.9743 & 10.72 & 6.10 &  7.83 & 5.05 \\
\textbf{ByT5$_\text{4k}$} & \underline{0.9198} & \underline{0.9199} & \underline{0.9197} & 0.4436 & 0.4011 & \underline{4.2007} & \underline{5.9553} & 10.47 & 6.40 & 8.03 & 5.59 \\
\hdashline[3pt/5pt]
\textbf{GPT-2$_\text{8k}$} & 0.8231 & 0.8714 &  0.8463 & \textbf{0.1188} & \textbf{0.0821} & 4.7338 & 5.9779 & 10.83 & 7.30 & 9.23 & 6.58 \\
\textbf{DialoGPT$_\text{8k}$} & 0.8183 & 0.8638 & 0.8402 & \underline{0.1329} & \underline{0.0876} & 4.9483 & 5.9787 & 10.10 & 8.23 & 10.47 & 8.03 \\
\textbf{BART$_\text{8k}$} & 0.8073 & \underline{0.9073} & \underline{0.8538} & 0.4451 & 0.2620 & 5.1885 & \textbf{5.8138} & 14.46 & 13.17 & 13.46 & 11.00 \\ 
\textbf{Flan-T5$_\text{8k}$} & \textbf{0.9289} &\textbf{ 0.9290} & \textbf{0.9288} & 0.5262 & 0.4289 & \underline{4.6310} & 5.9707 & 10.67 & 5.95 & 7.67 &  4.99 \\
\textbf{ByT5$_\text{8k}$} & \underline{0.9208} & \underline{0.9219} & \underline{0.9212} & 0.4549 & 0.4063 & \textbf{4.2830} & \underline{5.9598} & 10.68 & 6.52 & 8.04 & 5.85 \\
\hdashline[3pt/5pt]
\textbf{GPT-2$_\text{13K}$} &  0.8258 & 0.8738 & 0.8489 & \textbf{0.1256} & \textbf{0.0870} & 4.7119 & \textbf{5.9536} & 11.65 &  8.59 &  10.84 & 7.73  \\
\textbf{DialoGPT$_\text{13K}$} & 0.8216 & 0.8676 & 0.8437 & \underline{0.1348} & \underline{0.0891} & 5.0811 & 5.9733 & 10.51 & 9.01 & 11.03 & 8.79 \\
\textbf{BART$_\text{13K}$} &  0.7903 &  0.9141 &  0.8470 &  0.4201 & 0.3216 & \textbf{4.0013*} & 6.1303 & 17.97 & 13.25 & 16.91 & 9.82 \\ 
\textbf{Flan-T5$_\text{13K}$} & \underline{0.9311} & \underline{0.9309} & \underline{0.9309} &  0.5226 & 0.4288 & 4.5770 & 5.9681 & 10.64 & 5.98 & 7.63 & 5.06 \\
\textbf{ByT5$_\text{13k}$} & 0.9236 & 0.9246 & 0.9240 & 0.4619 & 0.4128 & \underline{4.2982} & \underline{5.9634} & 10.73 & 6.35 & 7.93 &  5.67 \\
\textbf{Baichuan+LoRA$_\text{13K}$} & \textbf{0.9442*} & \textbf{0.9411*} & \textbf{0.9425*} &  0.4908 & 0.4236 & 4.4789 & 5.9771 & 9.98 & 5.60 & 7.15 & 4.62 \\ 
\midrule
\textbf{ChatGPT} & 0.8851 & 0.8898 & 0.8873 & 0.5495 & 0.3948 & 4.7725 & 5.9445 & 10.37 & 7.67 & 9.06 & 8.25 \\ 
\midrule
\textbf{Brown Corpus Prose} & - & - & - & 0.4870 & 0.2275 & 5.1431 & 5.9561 & 11.09 & 13.15 & 15.56 & 16.60 \\
\bottomrule
\end{tabular}
}
\caption{\label{tab:additional-evaluation-paraphrase}
 Results of automatic evaluation on tongue-twister-related metrics for the style-transfer setting. Results in \textbf{bold} represent the best performance for a given training data quantity, and \underline{underlined} presents the second-best performance. Results followed by an asterisk * denote the overall best performance. Brown Corpus Prose presents the scores of standard prose from the Brown Corpus from NLTK (across 602 sentences with a minimum length of 25 words).}
\end{table*}

\paragraph{\textbf{Style-Transfer}}
We present the results for automatic evaluation in the style-transfer task setting in \autoref{tab:auto-evaluation-paraphrase} and \autoref{tab:additional-evaluation-paraphrase}. Overall, we see much the same pattern as with the topic-to-twister setting, with performance ordering of DialoGPT, GPT-2, BART, ByT5, and Flan-T5 across our referenced metrics (again, with our T5-based models alternating in ranking). However, unlike in the previous setting, the style-transfer setting appears to favor Baichuan, which presents the highest scores on ROUGE-based referenced metrics (when considering models also trained on 13k samples), whilst ByT5 performs the best on BLEU. Additionally, scores across the board are higher than seen in the topic-to-twister setting, but this is to be expected as the style-transfer task setting provides a structure for the generated tongue twister based on the length and word choices present in the original. Consequently, increased amounts of overlap between the desired output and the gold reference are expected due to not all words requiring modification. In contrast to the topic-to-twister setting, however, we do not see ChatGPT outperform all models, rather it is beaten by Baichuan and Flan-T5 in most cases. In terms of the performance difference on these metrics as the amount of available training data is increased, unlike the topic-to-twister setting we see a more clear growth in performance on reference-based metrics alongside training data in the majority of cases. 
Regarding the readability and phonemic metrics presented in \autoref{tab:additional-evaluation-paraphrase}, we again see similar patterns with GPT-2 based models scoring well on the naive phoneme-based metrics (IPO/PO), but all models performing similarly when regarding the more informed iPED/oPED measures. Regarding readability, scores overall are seen to be lower than in the topic-to-twister setting, suggesting more legible text. On one hand, this may indicate that the style has failed to transfer, with the paraphrase representing a well-written standard non-literary text (and therefore the model has effectively resorted to auto-encoding). On the other hand, this may also be an artifact of following the original structure of the non-literary paraphrase, therefore avoiding unnaturally long sentences and nonsensical outputs.

\subsection{Human Evaluation}
\label{sec:human_eval_unconstrained}

\begin{table}[ht]
\centering \small
\resizebox{1\linewidth}{!}{
\begin{tabular}{l|llllllll}
\toprule[1pt]
\multirow{2}{*}{\textbf{Score (1 to 5)}} & \multicolumn{7}{c}{\textbf{Trained Topic-to-Twister}}  \\
\cmidrule(lr){2-8}  
        & \textbf{Flan-T5$_{\text{2k}}$} & \textbf{Flan-T5$_{\text{13k}}$} & \textbf{ByT5$_{\text{2k}}$} & \textbf{ByT5$_{\text{13k}}$} & \textbf{Baichuan$_\text{13K}$} & \textbf{ChatGPT} & \textbf{Golden} \\  
\midrule
\textbf{Relevance} & 2.077$^{***}$ & 1.625$^{***}$ &2.180$^{**}$ & 2.070$^{**}$ &  1.688$^{**}$  & \textbf{4.824$^{**}$}  & \underline{4.647}$^{**}$  \\  
\textbf{Articulation} & 1.800$^{**}$  & 1.882$^{**}$ & 3.050$^{**}$ & 3.290$^{***}$ &  1.176$^{**}$  & \underline{2.667}$^{**}$  & \textbf{3.375}$^{*}$  \\
\textbf{Fluency} & 2.000$^{**}$  & 3.462$^{**}$ & 2.290$^{**}$ & 3.380$^{**}$ &  1.450$^{**}$  & \underline{4.632}$^{**}$  & \textbf{4.944$^{**}$}  \\
\textbf{Coherence} & 1.200$^{**}$  & 2.133$^{**}$ & 1.610$^{**}$ & 1.930$^{**}$ &  1.118$^{**}$  & \underline{4.333}$^{**}$  & \textbf{4.444$^{***}$} \\
\textbf{Entertainment} & 1.200$^{*}$  & 1.833$^{*}$ & 1.620$^{**}$ & 2.070$^{**}$ &  1.000$^{*}$  & \textbf{3.267$^{*}$ } & \underline{3.077}$^{**}$  \\
\textbf{Overall} & 1.063$^{*}$  & 1.888$^{*}$ & 1.800$^{**}$ & 2.300$^{**}$ &  1.316$^{**}$  & \underline{3.538}$^{**}$  & \textbf{3.909$^{**}$ } \\
\bottomrule[1pt]
\end{tabular}
}
\caption{Results of human evaluation in the topic-to-twister task setting. The best scores are in \textbf{bold}, and the second-best are \underline{underlined}. We calculate Fleiss' Kappa for each metric, and we mark the extent of agreement with the following markings: $^{*}$ fair agreement; $^{**}$ moderate agreement; $^{***}$ substantial or almost perfect agreement.}
\label{tab:human_eval_topic_unconstrained}
\end{table}

\begin{table}[ht]
\centering \small
\resizebox{1\linewidth}{!}{
\begin{tabular}{l|llllllll}
\toprule[1pt]
\multirow{2}{*}{\textbf{Score (1 to 5)}} & \multicolumn{7}{c}{\textbf{Style-Transfer}}  \\
\cmidrule(lr){2-8}  
& \textbf{Flan-T5$_{\text{2k}}$} & \textbf{Flan-T5$_{\text{13k}}$} & \textbf{ByT5$_{\text{2k}}$} & \textbf{ByT5$_{\text{13k}}$} & \textbf{Baichuan$_\text{13K}$} & \textbf{ChatGPT} & \textbf{Golden} \\  
\midrule
\textbf{Relevance} & 4.467$^{*}$  & 4.714$^{**}$  &4.090$^{*}$ &4.160$^{*}$& \underline{4.882}$^{**}$  & 4.692$^{**}$  & \textbf{5.000$^{*}$}  \\  
\textbf{Articulation} & 1.462$^{**}$  & 2.231$^{**}$  &3.710$^{**}$ &3.520$^{*}$& 2.250$^{**}$  & \underline{2.471}$^{**}$  & \textbf{3.313$^{**}$}  \\
\textbf{Fluency} & 4.611$^{**}$  & 4.895$^{***}$  &4.150$^{**}$ &4.270$^{**}$& 4.800$^{*}$  & \textbf{5.000$^{**}$}  & \underline{4.950}$^{**}$  \\
\textbf{Coherence} & 4.188$^{**}$  & \underline{4.733}$^{**}$  &3.500$^{**}$ &3.550$^{**}$& 4.375$^{**}$  & 3.929$^{**}$  & \textbf{4.786$^{*}$}  \\
\textbf{Entertainment}  & 1.733$^{*}$ & 2.000$^{*}$  &3.400$^{**}$ &3.240$^{*}$& \underline{2.846}$^{*}$  & 2.583$^{*}$  & \textbf{3.455$^{**}$}  \\
\textbf{Overall} & 2.308$^{*}$ & 3.000$^{*}$ &3.500$^{**}$ &3.510$^{*}$& 3.333$^{*}$ & \underline{3.500}$^{*}$ & \textbf{3.941$^{**}$} \\
\bottomrule[1pt]
\end{tabular}
}
\caption{Results of human evaluation in the style-transfer task setting. The best scores are in \textbf{bold}, and the second-best are \underline{underlined}. We calculate Fleiss' Kappa for each metric, and we mark the extent of agreement with the following markings: $^{*}$ fair agreement; $^{**}$ moderate agreement; $^{***}$ substantial or almost perfect agreement. }
\label{tab:human_eval_style}
\end{table}

The results of human evaluation for the topic-to-twister setting are presented in \autoref{tab:human_eval_topic_unconstrained}, and the results for the style-transfer setting are in \autoref{tab:human_eval_style}.

\paragraph{\textbf{Topic-to-Twister}}Firstly, for the topic-to-twister setting in \autoref{tab:human_eval_topic_unconstrained}, we can see that the highest scores for all criteria go to the human-authored "golden" samples, or those generated with ChatGPT. With a rating of "3" being considered the midpoint for "neither agree nor disagree" with the given criteria statements, it is evident that our finetuned unconstrained generation models struggle with the open-ended topic-to-twister task setting. However, when investigating the finetuned model performance we do see some patterns start to emerge that indicate the benefit of having such an extensive dataset as TwistList 2.0. For instance, Flan-T5 benefits from additional training samples, particularly regarding the metrics of Fluency and Coherence, and moderately in Entertainment and the holistic Overall rating. These findings for Fluency and Coherence are intuitive, as additional training samples increase the likelihood of generating grammatical and semantically coherent outputs due to the increased training data through which to learn these patterns. On the other hand, Articulation and Entertainment refer to more creative-language-specific metrics that are more abstract, and consequently difficult to learn from the training data. Relevance is the only metric shown to decrease when moving from 2k training samples to 13k, and we hypothesize that this is due to the increased wealth of training data that contains a more abstract link between the input topic and the generation, therefore decreasing the likelihood of the input words being directly present in the output (which is a straightforward way of performing well on Relevance metrics). Overall, however, we see ByT5 is preferable to Flan-T5, outperforming it in human evaluation on the criteria of Relevance, Articulation, Entertainment, and Overall, but underperforming in regard to Fluency and Coherence in the 13k instance. This additionally sheds some light on how humans perceive quality tongue twisters, with articulation difficulty being a more integral feature for a tongue twister than grammatical validity and semantic coherence. Finally, we see Baichuan struggle immensely with the topic-to-twister task setting (which is explored further in \S\ref{sec:case_studies}), frequently opting to repeat the input topic phrase continuously, therefore artificially increasing Relevance scores, but performing poorly on all other metrics.

\paragraph{\textbf{Style-Transfer}} On the other hand, we see better performance across the board for the style-transfer setting as facilitated by the additional high-quality paraphrases we include in TwistList 2.0. We hypothesize that the reason for this is that style-transfer requires already having access to a well-formed input, and additionally acts as an extended, very prescriptive form of an input topic, where the entire tongue twister is predefined in structure and semantics. As a result, we see Baichuan outperform ChatGPT on the criteria of Entertainment and Relevance, whilst Flan-T5 trained on the 13k split outperforms ChatGPT regarding semantic coherence of the output. Interestingly, we observe that ByT5 trained on only 2k examples performs more closely to the 13k version than is seen in Flan-T5, suggesting the alternative tokenization approach allows ByT5 to learn the relevant patterns from fewer examples. Importantly, we see Baichuan perform much better in the style-transfer task than in the topic-to-twister setting, suggesting that Baichuan requires much more explicit instruction to generate high-quality outputs and understand a given prompt.

\subsection{Human vs. Automatic Metrics}

\begin{table}[ht]
\centering \small
\resizebox{1\linewidth}{!}{
\begin{tabular}{l|llllllll}
\toprule[1pt]
\multirow{2}{*}{} & \multicolumn{8}{c}{\textbf{Automatic Metrics}}  \\
\cmidrule(lr){2-9} 
& \textbf{Re-D} & \textbf{Re-F} & \textbf{Re-G} & \textbf{Re-A} & \textbf{IPO} & \textbf{PO} & \textbf{iPED} & \textbf{oPED} \\  
\midrule
\textbf{Relevance} &-.395 & -.442& -.423& -.494&.394 & .496& -.004&.104 \\  
\textbf{Articulation} & -.218 & -.271&-.224 &-.218 &-.063 & -.087 & -.226 & .110 \\
\textbf{Fluency} & -.400 & -.590 & -.564 & -.630 & .516 & .674& -.001& -.013 \\
\textbf{Coherence} & -.320&-.482 &-.551 & .482& - .486 & .622 & -.003 & .064\\
\textbf{Entertainment} & .301& -.287& -.267& -.278& .151& .262& -.077& -.106\\
\textbf{Overall}  & .342& .386&.372 &.386 & .211& .361& -.129&.055 \\
\bottomrule[1pt]
\end{tabular}
}
\caption{Spearman correlation coefficients for each human evaluation criterion with reference-free automatic metrics. We take the mean score across evaluators for the combined unconstrained topic-to-twister and style-transfer task settings.}
\label{tab:human_automatic}
\end{table}

\paragraph{\textbf{\textit{Human-Machinne Correlation}}} In order to see the effectiveness of our selected automatic metrics, we calculated the Spearman correlations between our 8 reference-free metrics, including readability (i.e., Re-D, Re-F, Re-G, and Re-A) and our phonetic metrics (i.e., IPO, PO, iPED, and oPED), against the human evaluation ratings for all criteria (i.e., relevance, articulation, fluency, coherence, entertainment, and overall). Correlations are presented in \autoref{tab:human_automatic}. To evaluate the predictive power of the automatic metrics for human ratings, we developed a standard multiple linear regression model for each criterion. The model is defined as follows:
\begin{equation}
y_i = \beta_0 + \sum_{j=1}^{p} \beta_j X_{ij} + \epsilon_i
\end{equation}
where \( y_i \) represents the human score, \( \beta_0 \) is the intercept, \( \beta_j \) are the coefficients for the \( p \) automatic metrics, and \( \epsilon_i \) is the error term. No regularization was applied to the model.
The \( R^2 \) values, which indicate the proportion of variance explained by the automatic metrics, were validated through 5-fold cross-validation to ensure the stability and significance of the predictions. The average \( R^2 \) values across the folds were as follows: Fluency (\( R^2 = .553 \)), Coherence (\( R^2 = .496 \)), Relevance (\( R^2 = .393 \)), Overall (\( R^2 = .260 \)), Articulation (\( R^2 = .241 \)), and Entertainment (\( R^2 = .152 \)). 
Furthermore, all coefficients in the models were found to be statistically significant, with p-values below 0.01 (\( \alpha = 0.01 \)), indicating that each of the automatic metrics significantly contributes to the prediction of human ratings.

Overall, we intuitively see the Entertainment criterion being the hardest to predict due to the inherent subjectivity of this criterion. On the other hand, we see our naive phonemic metrics (IPO and PO) demonstrate moderate correlations with Relevance, Fluency, and Coherence. This is due to the high relevance of tongue twisters often being seen in examples where the input topic is simply repeated, which results in high levels of phoneme overlap. Similarly, high fluency scores are given to more sentences that better reflect standard non-literary text, which consequently score lower on IPO and PO, and less coherent outputs are often produced by repeating the same word. Moreover, our "informed" phonemic metrics (iPED/oPED) show little correlation with human results on these criteria, but iPED demonstrates evidence of a correlation with articulatory difficulty. However, the articulatory difficulty still remains challenging to predict, even from these phonemic metrics. We hypothesize that this may be related to the "visual tongue twister" effect \cite{MCCUTCHEN-VISUAL}. This is due to human evaluation being performed online and asynchronously, where we cannot force participants to speak aloud each tongue twister. Consequently, we hypothesize that the naive metrics may correlate better with human judgments as human ratings were confounded by the visual impact of the tongue twister (for example, seeing a particular grapheme repeated numerous times representing the same sound). Furthermore, there are additional effects from the influence of other factors such as fluency and coherence affecting articulation due to violating expectations and reducing legibility. Consequently, iPED/oPED should be used as indicators of text resembling a tongue twister, but not as a holistic metric for overall quality. It is clear from comparison with standard non-literary text that the phonetic metrics are able to differentiate the specific characteristics of tongue twisters from that of standard text, but phonemic complexity is not the sole contributor to the perception of articulatory difficulty.

\paragraph{\textbf{\textit{GPT-4o "Human" Evaluation}}} We additionally perform evaluation on the same samples presented to human evaluators in the unconstrained topic-to-twister and style-transfer settings using GPT-4o via prompting the model with the same rubric presented to human evaluators (see Appendix \ref{apx:evaluation-rubric}).\footnote{Specifically, \textit{gpt-4o-2024-05-13} via the API.} Overall, we see moderate-to-high correlation between model scores and human-assigned scores for all criteria: Relevance ($\rho$ = .671), Articulation ($\rho$ = .653), Fluency ($\rho$ = .768), Coherence ($\rho$ = .658), Entertainment ($\rho$ = .716), and Overall ($\rho$ = .776).\footnote{All significant at $\alpha$ = .01.} This demonstrates that our human evaluation rubric is clear and well defined, and can be effectively followed by state-of-the-art LLMs.

\section{Tongue Twister Generation with Phoneme Aware Constrained Decoding}
\label{sec:PACD}
In the following section, we present work on a constrained decoding-based approach to tongue twister generation. In contrast to \S\ref{sec:unconstrained}, here we focus exclusively on the topic-to-twister task setting due to the text-continuation nature of our decoding approach. The benefit of this algorithm, in contrast to the finetuned models presented in  \S\ref{sec:unconstrained} is that constrained decoding \textit{guarantees} that only desirable tokens appear in the output due to the layering of hard phoneme-based constraints as previously discussed. Additionally, due to how this system interacts with language model token predictions, this process can be applied to any autoregressive language model, including both pre-trained base models and further fine-tuned models. 

\subsection{Task Definition}
For a given input prompt we aim to generate a tongue twister $T$, whereby $T$ comprises a sequence of words $ \{w_1, w_2,... w_{n}\}$. In contrast with the previous section, in this task setting $T$ is a continuation of the input prompt that we generate token by token, evaluating the language model's next token predictions at each step.
As per \S\ref{sec:unconstrained}, the generated output must satisfy the following constraints: (1) the output should be semantically related to the input topic phrase; (2) the output should show maximal levels of phonemic overlap across tokens; and (3) the output should be grammatically valid.

\subsection{Phoneme-Aware Constrained Decoding Module (PACD)}

\begin{algorithm}[ht]
    \caption{Phoneme-Aware Constrained Decoding (PACD)}\label{alg:algorithm}
    \begin{algorithmic}[1]
        \For{each $s$ in $S$}
            \State $ph_1$ = $\text{G2P}(\text{topic in }$s$)[0]$
            \State $ph_2$ = \textsc{$\argmin$}(\textsc{$PED$}($ph_1, WIP \setminus \{ph_1\}$))
            \While{$\text{len}($s$^*) < \text{max\_length}$}
                \State Retrieve next word probabilities $P = \{p_1, \ldots, p_n\}$ from $LM(s^*)$
                \For{each $rank$, $p$ in \textsc{enumerate}($P$)}
                    \If{$p$ in $F$ and $rank$ $\leq$ function\_window}
                        \State append $p$ to $s^*$
                        \State \textbf{break}
                    \EndIf
                    \State $candidates$ = []
                    \If{$\text{len}(p) > \text{$min\_stem\_length$}$ and $\text{G2P}(p)[0] == ph_1$ or $ph_2$ }
                        \State append $p$ to $candidates$
                        \State $temp\_prompt = s$*$ + p$
                        \For{i in range(4)}
                            \State $next\_token$ = $LM(temp\_prompt)$
                            \If{$next\_token$.isalpha() and $next\_token$[0] != " "}
                                \State $longest$ = "".join($candidates$)
                                \State$longest$ += $next\_token$
                                \State append $longest$ to $candidates$
                            \EndIf
                            \If{$next\_token$[-1] == " "}
                                \State{\textbf{break}}
                            \EndIf
                        \EndFor    
                        \State{$temp\_prompt$ = ""}
                        \For{$candidate$ in $candidates$.sort(longest-to-shortest)}
                            \If{$candidate \in D$ and \textsc{count}($candidate$ in $s^*$) $<$ $max\_repetition$}
                                \State{append $candidate$ to $s^*$}
                            \EndIf
                        \EndFor
                    \EndIf 
                \EndFor
            \EndWhile    
        \EndFor         
    \end{algorithmic}
\end{algorithm}

We present an outline of our Phoneme-Aware Constrained Decoding algorithm (PACD) in Algorithm \ref{alg:algorithm}.\footnote{PACD is intended to be pronounced as "packed".} To summarize, for every starting prompt $s$ in our test set \textit{S}, we firstly perform grapheme-to-phoneme conversion \textit{G2P} with the g2p-en package and extract the initial phoneme of the first word in the topic phrase part of \textit{s} in order to increase the likelihood of retrieving a semantically related output (with the phoneme denoted as $ph_1$). However, where the selected phoneme is not a valid consonant, we randomly select a phoneme from a list of phonotactically legal word-initial consonant phonemes for English, \textit{WIP}. Following this, we calculate the weighted phonemic edit distance (PED) between $ph_1$ and all other legal word-initial phonemes and select the lowest scoring (i.e., most similar) as our secondary phoneme $ph_2$ (analogous to the system in \S\ref{sec:TwisterLister} for TwisterLister). Following this, we autoregressively generate new tokens up to the limit defined by \textit{max\_length} based on numerous criteria. To do this, we feed the starting prompt \textit{s} into our language model of choice, $LM$, and retrieve the next token probabilities \textit{P}. Then, in descending order (i.e., most-probable to least-probable next token) we iterate through predictions \textit{p} $\in$ \textit{P} until specific criteria are met. Firstly, to increase the likelihood of generating grammatical output, if a function word (such as an article, pronoun, conjunction, preposition, or auxiliary verb) from our function word list \textit{F} is within the range defined by \textit{function\_window}, we allow it to generate. For example, when generating the first word, if \textit{function\_window} is set to 3, and "The" is the token with the 2\textsuperscript{nd} highest probability, we allow it to generate as it is within our allowed range of top-3. 

To account for subword tokenization in non-function words, we next check that the predicted token is longer than the limit defined by \textit{min\_stem\_length}. We do this as we find the result of not limiting this to be a reliance on outputting the grapheme that most closely corresponds to a desired phoneme (rather than a sequence that better resembles a morpheme), significantly increasing inference time and decreasing output quality. We then employ our phoneme constraints by feeding the predicted word stem $p$ into a grapheme-to-phoneme model \textit{G2P} and comparing the first phoneme to $ph_1$ and $ph_2$, continuing if it matches either. Consequently, in this stage, we have ensured that generated words are either closed-class grammatical function words or start with one of the two phonologically similar sounds selected in lines 2-3 of Algorithm~\ref{alg:algorithm}. Following this, we optionally engage the subword loop (lines 15-25 in Algorithm \ref{alg:algorithm}). Within this loop, we temporarily append our candidate word stem to the current prompt \textit{s*} to create \textit{temp\_prompt}. Following this, we feed \textit{temp\_prompt} to the language model $LM$ and take only the token with the highest probability, \textit{next\_token}. We then check that \textit{next\_token} is alphabetical and does not start with whitespace (as this would indicate the model was predicting a new word, rather than a continuation). If this is the case, we append it to \textit{temp\_prompt} and perform the loop again (up to 4 times, allowing words that consist of 1-5 subwords). Within this loop, we build a list of potential words, \textit{candidates}, by appending the concatenated subwords (e.g., ["anti", "antidis", "antidisestablish", "antidisestablishment", "antidisestablishmentarian"]) We terminate this loop early if a predicted \textit{next\_token} ends with whitespace, as this indicates that the model has predicted the end of the current word. Once we have our list of \textit{candidates}, we iterate through them from the longest (i.e., consisting of the most subwords) to the shortest, assessing the following criteria.

Firstly, we check that each \textit{candidate} $\in$ \textit{candidates} is longer (in characters) or equal to the length defined by \textit{min\_word\_length} and that \textit{candidate} $\in$ \textit{D}, where \textit{D} is the English dictionary as defined by the Enchant Python package.\footnote{Available at \url{https://pypi.org/project/pyenchant/}.} Once this check is complete, we ensure that we have not already generated this specific $candidate$ more times than permitted by \textit{max\_repetition}, in order to avoid falling into the perpetual loop of repeating the same words that language models are prone to (see \S\ref{sec:case_studies}). Here we use $s^*$ to denote the starting prompt \textit{s} with \textit{newly} generated tokens appended, which is to say that $s^* - s$ equals only the LM-generated words. Finally, if no \textit{candidate} meets the criteria, we increase \textit{rank} by looking at the next-best prediction in \textit{P} until the vocabulary is exhausted. In the case where no suitable candidates exist in the vocabulary, we simply move on to the next \textit{s}.

To illustrate the algorithm with an example, consider \textit{S} to consist of two input topics \textit{s}: [fun, sadness]. For the first example ("fun"), we select $ph_1$ by performing G2P on the topic, returning \textipa{/fUn/}, and select the word-initial \textipa{/f/} as $ph_1$. We then decide $ph_2$ by selecting the next phoneme that has the lowest phonemic edit distance to \textipa{/f/}, returning \textipa{/v/} as $ph_2$. Following this, we feed the full prompt \textit{"Generate a tongue twister on the topic of 'fun'. "} to the language model, and retrieve the next-token probabilities \textit{P}. For example, \textit{P} could be $\{1: The, 2: It, 3: A...\}$, where the set is the length of the decoded vocabulary. We then iterate through the predictions until a word meets our criteria. For instance, the most likely continuation, "The", is in the function word list $F$ and within the $function\_window$ due to being at \textit{rank} 1, so we append it to the prompt and now have \textit{"Generate a tongue twister on the topic of 'Fun'. \underline{The}"}, which we now denote $s*$. We then feed this extended prompt into the language model to retrieve the second word, where $P$ may look like \textit{\{1: grey, 2: big, 3: fun\}}. Here, options in \textit{rank} 1 and 2 ("grey" and "big") do not start with $ph_1$ or $ph_2$ and are also not function words in $F$. However, the word in rank 3, "fun" is transcribed phonemically as \textipa{/fUn/}, where the initial phoneme \textipa{/f/} matches $ph_1$, so we enter the subword loop and find the candidate "funniest", allow it to generate, and append it to the prompt, resulting in \textit{"Generate a tongue twister on the topic of 'Fun'. \underline{The funniest}"}. We repeat this until we generate new tokens up to \textit{max\_length}, and then start the process again for the remaining topic in \textit{S}, which is "sadness". We additionally make sure that we do not generate the same word more than once, as determined by \textit{max\_repetition}, and we do not generate words shorter in length than \textit{min\_word\_length}.

\subsection{Constrained Models}

To demonstrate the effectiveness of our decoding module, we utilize 2 decoder-only autoregressive language models as our $LM$: \textbf{GPT-2} \cite{GPT-2} and \textbf{Baichuan} \cite{baichuan}, to which our module will be applied on top. In addition, we investigate to what extent fine-tuning a model towards tongue twister generation is beneficial, by additionally using our finetuned GPT-2 and Baichuan from \S\ref{sec:unconstrained}, referred to herein as \textbf{GPT-2$_\text{13K}$} and \textbf{Baichuan$_\text{13K}$}. Importantly, we assess only these models fine-tuned on the largest amount of data, 13k.

Regarding the other settings for PACD, we set $max\_length$ to 30 (as a sensible midpoint generation length ascertained from \autoref{tab:auto-evaluation-topic}), \textit{function\_window} to 1, and implement $F$ as the NLTK stopwords list with all punctuation removed. Additionally, we set \textit{min\_stem\_length} to 2 and \textit{min\_word\_length} to 3 (as all standard 1 or 2-letter words $\{I, a, I'm, am, at, in, up, on\} \in F$). Additionally, \textit{max\_repetition} is set to 1, in effect banning wholesale repetition (though allowing plural/singular forms, and case variants) to avoid the patterns seen in standard autoregressive models. Finally, we use the \textit{g2p-en} package for our \textit{G2P} model, as in the creation of TwisterLister (\S\ref{sec:QC}). Finally, we only decode the top 2500 predictions in each timestep rather than the entire vocabulary in order to speed up inference significantly, as it is rare to select tokens ranked below this point. Importantly, due to the computational cost of our algorithm, we load Baichuan using 8-bit quantization to make inference possible. Overall, for GPT-2, PACD takes approximately 5-10 seconds to generate a 30-word tongue twister (with or without subword generation), whilst Baichuan takes 10-15 seconds when generating full words only, and 30-100+ seconds when allowing subwords on a consumer CPU (i5 9600k). We posit future work on the parallelization of elements of PACD to be more compute-efficient and take advantage of the GPU.

\section{Results (PACD)}

We perform automatic evaluation in the same manner as \S\ref{sec:automatic_results_unconstrained}, and report both reference-based (BLEU/ROUGE/BERTScore) and unreferenced metrics (Init-PO/PO/iPED/oPED and the readability metric suite). We additionally perform human evaluation in the same manner as \S\ref{sec:human_eval_unconstrained} and with the same evaluators, following the protocol for the topic-to-twister task setting. Each evaluator is presented with 20 examples (using the same inputs as in \S\ref{sec:human_eval_unconstrained}) from base GPT-2 and Baichuan with the addition of PACD, or finetuned GPT-2$_\text{13k}$ and Baichuan$_\text{13k}$ with and without PACD.

\subsection{Automatic Evaluation}

\begin{table*}[ht]
\label{tab:auto_constrained}
\centering \small
\resizebox{1\linewidth}{!}{
\begin{tabular}{l|cccc|ccc}
\toprule
\textbf{Model} & \textbf{B-1$\uparrow$} & \textbf{B-2$\uparrow$} & \textbf{B-3$\uparrow$}  & \textbf{B-4$\uparrow$}  & \textbf{Ro-1$\uparrow$} & \textbf{Ro-2$\uparrow$} & \textbf{Ro-L$\uparrow$} \\
\midrule
\textbf{GPT-2$_\text{13k}$ -w/o} & 0.0440 & 0.0152 & \underline{0.0061} & \underline{0.0025} & 5.9412 & 0.6024 & 5.5890 \\
\textbf{GPT-2$_\text{13k}$ -w} & \textbf{0.0832} & \underline{0.0170} & 0.0046 & 0.0013 &  \textbf{8.6673} & \underline{0.9067} & \textbf{7.5011} \\
\textbf{GPT-2$_\text{13k}$ -ws} & 0.0584 & 0.0045 & 0.0003 & 0.0000 & 0.0089 & 0.0000 & 0.0792 \\
\textbf{GPT-2 -w} &  \underline{0.0759} & 0.0142 &  0.0032 & 0.0010 & \underline{8.0608} & 0.6403 & \underline{6.8765} \\ 
\textbf{GPT-2 -ws} & 0.0594 & 0.0059 & 0.0009 & 0.0001 & 0.0940 & 0.0121 & 0.0790  \\
\hdashline[3pt/5pt]
\textbf{Baichuan$_\text{13K}$ -w/o}& 0.0463 & \textbf{0.0227} & \textbf{0.0131} & \textbf{0.0080} & 6.2215 & \textbf{1.3878} & 5.9212\\
\textbf{Baichuan$_\text{13K}$-w} & 0.0493 & 0.0041 & 0.0004 & 0.0000 & 0.0846 & 0.0081 & 0.0677\\
\textbf{Baichuan$_\text{13K}$ -ws} & 0.0525 & 0.0051 & 0.0006 & 0.0001 & 0.0940 & 0.0106 & 0.0742 \\
\textbf{Baichuan -w} & 0.0547 & 0.0054 & 0.0009 & 0.0002 & 0.0885 & 0.0109 & 0.0750 \\ 
\textbf{Baichuan -ws} & 0.0600 & 0.0080 & 0.0019 & 0.0005 & 0.1010 & 0.0175 & 0.0846 \\ 
\bottomrule
\end{tabular}
}
\caption{\label{tab:auto-evaluation-constrained}
Results of automatic evaluation on typical reference-based metrics in NLG for the constrained decoding approach to tongue twister generation in the topic-to-twister task setting. Results in \textbf{bold} represent the best performance, and \underline{underlined} results show the second-best performance. -w/o denotes without constraints, -w denotes the addition of our PACD module, and -ws denotes the addition of PACD with subword generation enabled.}
\end{table*}

\begin{table*}[ht]
\label{tab:additional_constrained}
\centering
\resizebox{1\linewidth}{!}{
\begin{tabular}{l|ccc|cccc|cccc}
\toprule
\textbf{Model} & \textbf{BS-P$\uparrow$} & \textbf{BS-R$\uparrow$} & \textbf{BS-F1$\uparrow$} & \textbf{IPO$\downarrow$}  & \textbf{PO$\downarrow$}  & \textbf{iPED$\downarrow$} & \textbf{oPED$\downarrow$} & \textbf{Re-D} & \textbf{Re-F} & \textbf{Re-G} & \textbf{Re-A}\\
\midrule
\textbf{GPT-2$_\text{13k}$ -w/o} &  0.7748 & \underline{0.8322} & 0.8021 & \textbf{0.1261} & \textbf{0.0859} & 4.2671 & 5.8985 & 12.06 & 13.46 & 15.21 & 16.64 \\
\textbf{GPT-2$_\text{13k}$ -w} & 0.8056 & 0.8207 & 0.8129 & 0.2305 &  0.2317 & 3.0376 & 5.8930 & 13.88 & 14.58 & 17.58 & 18.77 \\
\textbf{GPT-2$_\text{13k}$ -ws} & 0.8018 & 0.8267 & 0.8139 & 0.1724 & 0.2287 & 1.7252 & 5.7686 & 11.60 & 12.24 & 15.59 & 17.20 \\
\textbf{GPT-2 -w} & 0.7986 &  0.8208 & 0.8094 & 0.1689 & \underline{0.2110} & 1.6529 & 5.7699 & 11.36 & 12.66 & 15.76 & 15.91 \\ 
\textbf{GPT-2 -ws} & \underline{0.8119} & 0.8258 & \underline{0.8186} & 0.2469 & 0.2614 & 3.1642 & 5.8991 & 9.02 & 10.45 & 14.16 & 13.65 \\
\hdashline[3pt/5pt]
\textbf{Baichuan$_\text{13K}$ -w/o}  & \textbf{0.9442} & \textbf{0.9411} & \textbf{0.9425} &  0.4908 & 0.4236 & 4.4789 & 5.9771 & 9.98 & 5.60 & 7.15 & 4.62 \\ 
\textbf{Baichuan$_\text{13K}$ -w} & 0.7915 & 0.8202 & 0.8053 & 0.1629 & 0.2416 & \underline{1.4324} & \underline{5.6876} & 10.72 & 11.18 & 13.52 & 15.18 \\
\textbf{Baichuan$_\text{13K}$ -ws} & 0.7887 & 0.8215 & 0.0742 & \underline{0.1512} & 0.2194 & \textbf{1.2898} & \textbf{5.6708} & 11.95 & 12.59 & 14.82 & 17.21 \\
\textbf{Baichuan -w} & 0.8014 & 0.8204 & 0.8106 & 0.2102 & 0.2461 & 2.4449 & 5.8094 & 9.08 & 10.55 & 14.26 &  14.25\\ 
\textbf{Baichuan -ws} & 0.7984 & 0.8217 & 0.8096 & 0.1992 & 0.2339 & 2.3284 & 5.8016 & 9.89 & 11.37 & 14.86 & 15.38 \\ 
\midrule
\textbf{Brown Corpus Prose} & - & - & - & 0.4870 & 0.2275 & 5.1431 & 5.9561 & 11.09 & 13.15 & 15.56 & 16.60 \\
\bottomrule
\end{tabular}
}
\caption{\label{tab:additional_constrained_auto}
Results of automatic evaluation on tongue-twister-related metrics for the constrained decoding approach to tongue twister generation in the topic-to-twister task setting. Results in \textbf{bold} represent the best performance and \underline{underlined} presents the second-best performance. -w/o denotes without constraints, -w denotes with the addition of our PACD module, and -ws denotes the addition of PACD with subword generation enabled. Brown Corpus Prose presents the scores of standard prose from the Brown Corpus from NLTK (across 602 sentences with a minimum length of 25 words).}

\end{table*}

The results of the automatic evaluation for the constrained decoding approach (PACD) are presented in \autoref{tab:auto-evaluation-constrained} for referenced metrics and \autoref{tab:additional_constrained_auto} for unreferenced metrics. Firstly for GPT-2, regarding the reference-based metrics, surprisingly we see finetuned GPT-2 with the addition of our constrained decoding module (\textit{GPT-2$_\text{13k}$ -w}) outperform the standard finetuned model (\textit{GPT-2$_\text{13k}$ -w/o}) on B-1 and B-2, in addition to all ROUGE-based metrics (Ro-1, Ro-2, and Ro-L). The exception here is for the higher-order BLEU measures, B-3 and B-4, where the unconstrained model achieves higher overlap. Base GPT-2 also benefits from the addition of our PACD module (\textit{GPT-2 -w}), outperforming the unconstrained finetuned model on the recall-based ROUGE metrics. Observing the unreferenced results in \autoref{tab:additional_constrained_auto}, performance ordering varies for the BERTScore semantic metrics (BS-P, BS-R, and BS-F1), with the unconstrained finetuned model (\textit{GPT-2$_\text{13k}$ -w/o}) trading places with the constrained equivalents (\textit{GPT-2$_\text{13k}$ -w and }-ws) for the most performant. However, base GPT-2 with the addition of PACD (\textit{GPT-2 -w})  consistently comes in second place for these metrics. Furthermore, we see the original finetuned unconstrained model perform the best when considering the original naive phoneme-based metrics IPO and PO, whilst the un-finetuned, yet constrained model, \textit{GPT-2 -w}, places second. It is here that the less naive newly presented phoneme-based metrics, iPED and oPED demonstrate their usefulness. For instance, whilst finetuned GPT-2 without any constraints (GPT-2$_\text{13k}$ -w/o) performed best on the naive metrics (IPO/PO), it performed the worst on the more linguistically informed metrics (iPED/oPED). This is because good tongue twisters exploit the relationships between similar sounds, whilst IPO/PO penalize these transitions as being low-quality, whilst our informed metrics reflect favorably on such transitions, penalizing them less than transitions between weakly related phonemes.

When looking at Baichuan, we see that the unconstrained finetuned model is the highest overall scorer on B-2, B-3, B-4, and Ro-2, as well as all BERTScore measures (BS-P, BS-R, and BS-F1). Interestingly, when analyzing the impact of the addition of PACD, we see the non-finetuned Baichuan with subword generation enabled (\textit{Baichuan -ws}) to outperform all other versions of Baichuan that contain PACD, even without being finetuned on the specific style of text we are aiming to generate (however, scores are low overall). Regarding the phonemic metrics, we again see the benefit of PACD in reducing the scores on all phoneme-based metrics by successfully increasing sound overlap. Additionally, we see finetuned Baichuan with the addition of PACD (\textit{Baichuan$_\text{13k}$ -w} and \textit{-ws}) to outperform the non-finetuned models in the phonetic metrics. Finally, the addition of PACD to Baichuan can be seen to lead to a significant increase in readability scores (i.e., an increase in reading difficulty). However, these scores still remain below the formal non-literary text of the Brown Corpus.

These results demonstrate the effectiveness of our constrained decoding approach that takes into consideration which word-initial phonemes would be the best to center generation around to encourage mispronunciation. Additionally, overall we see similar performance between our models using full-word and subword versions of PACD.

\subsection{Human Evaluation}
\label{sec:human_eval_constrained_}
\begin{table}[h]
\centering \small
\resizebox{1\linewidth}{!}{
\begin{tabular}{l|ccc|ccc}
\toprule[1pt]
\multirow{2}{*}{\textbf{Score (1 to 5)}} & \multicolumn{6}{c}{\textbf{Constrained Topic-to-Twister}}  \\
\cmidrule(lr){2-7}  
& \textbf{GPT-2$_\text{13k}$ -w/o} & \textbf{GPT-2$_\text{13k}$ -w} & \textbf{GPT-2 -w} & \textbf{Baichuan$_\text{13k}$ -w/o} &\textbf{Baichuan$_\text{13k}$ -w} &\textbf{Baichuan -w}\\ 
\midrule
\textbf{Relevance} & 2.16$^{***}$ & \underline{2.44}$^{***}$ & 2.42$^{***}$ & \textbf{3.04}$^{**}$& 2.39$^{**}$& 2.41$^{**}$ \\
\textbf{Articulation } & 2.80$^{***}$ & \textbf{3.85}$^{***}$ & \underline{3.69}$^{***}$ &3.21$^{**}$ & 3.58$^{***}$& 3.28$^{**}$  \\
\textbf{Fluency } & \underline{2.66}$^{***}$ & 2.39$^{**}$ &\textbf{3.13}$^{***}$ &2.24$^{**}$ & 1.84$^{**}$&1.88$^{**}$ \\
\textbf{Coherence } & \underline{2.17}$^{**}$ & 2.00$^{**}$  & \textbf{3.41}$^{***}$ & 2.09$^{**}$& 1.68$^{**}$& 1.64$^{**}$\\
\textbf{Entertainment}  & 2.00$^{***}$ & \underline{2.14}$^{**}$  & \textbf{2.97}$^{**}$ &1.81$^{**}$ & 1.67$^{**}$&1.64$^{**}$ \\
\textbf{Overall} & 1.72$^{***}$ & 2.00$^{**}$ &\textbf{2.91}$^{***}$ & \underline{2.09}$^{**}$ &1.98$^{**}$ & 1.89$^{**}$\\
\bottomrule[1pt]
\end{tabular}
}
\caption{Results of human evaluation in the topic-to-twister task setting using constrained decoding. -w denotes to models using our decoding algorithm, and -w/o denotes fine-tuned models without this additional module. The best scores are in \textbf{bold}, and the second-best are \underline{underlined}. We calculate Fleiss' Kappa for each metric, and we mark the extent of agreement with the following markings: $^{*}$ fair agreement; $^{**}$ moderate agreement; $^{***}$ substantial or almost perfect agreement.}
\label{tab:human_eval_constrained}
\end{table}

Human evaluation is performed identically to the evaluation reported in \S\ref{sec:human_eval_unconstrained} (and with the same evaluators). Importantly, however, due to enforcing a 30-word output length, we ask evaluators to not penalize generations on the criteria of "Fluency" for being cut off prematurely. Additionally, due to the similarity in the outputs with and without sub-word generation for PACD as seen in \S\ref{sec:case_studies}, we only perform human evaluation on the outputs of PACD with subword generation disabled, to minimize evaluator fatigue and potential acquiescence bias due to seeing similar results for either method. The results of human evaluation on outputs using our constrained generation PACD module are presented in \autoref{tab:human_eval_constrained}. We additionally perform human evaluation on GPT-2 and Baichuan trained on 13k samples (the former of which was excluded from \S\ref{sec:human_eval_unconstrained}) in order to have a point of comparison without the addition of PACD (referred to as \textit{GPT-2$_\text{13k}$ -w/o} and \textit{Baichuan$_\text{13k}$ -w/o}, respectively).

Regarding GPT-2, we see a clear benefit from the addition of our constrained decoding module. Firstly, across all evaluation criteria, either base (i.e., vanilla) or finetuned GPT-2 receives the highest scores (indicated in bold) when the PACD module is applied. However, we do see finetuned GPT-2 without the addition of PACD (\textit{GPT-2$_\text{13k}$ -w/o}) outperforming the equivalent model with PACD enabled on the criteria of Fluency and Coherence, which evaluate grammar and semantics, respectively. One explanation for this is that GPT-2 (overall) has been shown to perform poorly in the topic-to-twister setting when considering coherent and grammatical outputs (see also \S\ref{sec:case_studies}). Consequently, the addition of even more restrictive decoding rules brought about by PACD is slightly detrimental to the general quality of the output. On the other hand, non-finetuned language models are primarily designed to output standard prose that is grammatical and sensible initially. Consequently, the addition of the PACD decoding module does not damage the overall readability of the outputs too severely. Importantly, however, we do not analyze the performance of base GPT-2 without the PACD decoding module, as the performance of GPT-2 in zero-shot scenarios is poor, therefore making this not a meaningful point of comparison.
Interestingly, however, base GPT-2 with PACD (\textit{GPT-2 -w}) outperforms the finetuned models either with or without the presence of the additional decoding module, suggesting that the most desirable approach may be "train \textit{or} constrain", rather than "train \textit{and} constrain", as this allows phoneme-level control without sacrificing grammatically.

In contrast to GPT-2, when considering Baichuan, we see that human ratings decrease when moving from standalone fine-tuned Baichuan (\textit{Baichuan$_\text{13k}$ -w/o}) to either of the versions containing PACD. This, however, is not the case for the Articulation criteria, which increases in both PACD versions of Baichuan (i.e., \textit{Baichuan$_\text{13k}$ -w} and \textit{-ws}). This suggests a possible trade-off regarding the addition of constrained decoding to existing LMs, with larger models suffering from the additional control being exerted over their outputs, whilst smaller models are able to be guided towards desirable characteristics of the output language form by complimenting the reduced knowledge capacity of the smaller models during fine-tuning and zero-shot inference.

In conclusion, a significant finding from the human evaluation results is that the PACD module results in a substantial increase in the Difficulty of Articulation criteria (named "Articulation" in \autoref{tab:human_eval_constrained}). This indicates the success of our constraint cascade in selecting phoneme pairs that contribute to maximal levels of phoneme-level confusion, and therefore mispronunciation (i.e., the key to a successful tongue twister).

\section{Case Studies \& Qualitative Analysis}
\label{sec:case_studies}
\subsection{Topic-to-Twister}
\label{sec:unconstrained_case_topic}

\begin{table}[htp]
\centering \small
\resizebox{\linewidth}{!}{
\begin{tabular}{p{0.11\linewidth}|p{0.89\linewidth}}
\toprule
\textbf{Input:} &  "rural brewery" \\
\textbf{Golden:} & "Rory the warrior and Roger the worrier were reared wrongly in a rural brewery."\\
\midrule
\textbf{GPT-2$_{\text{2k}}$} & "In the bustling city of Chattanooga, where Chattanooga's vibrant craft beer scene thrives, The Chattanooga's craft beer scene thrives, [With Chattanooga's craft beer scene thriving, The Chattanooga's craft beer scene thrives,]\textbf{*3} With Chattanooga's craft bee
" \\
\hdashline[3pt/5pt]
\textbf{GPT-2$_{\text{13k}}$} & "In the bustling town of Chattanooga, where the townspeople reside, The Chattanooga's taproom is filled with tapestries of flavors. Chattanooga's taproom is filled with the finest tapestry, From the tapestry of Chattanooga's charm, it's a sight to see. Chattanooga's taproom is filled with the finest tapestry, From the tapestry of Chattanooga's charm, it's a sight to see. Chattanooga's taproom is filled with the finest tapes" \\
\midrule
\textbf{DialoGPT$_{\text{2k}}$} & "In the bustling city of Bethlehem, a bustling brewery was bustling with craftsmanship and craftsmanship was its main attraction. It was also the place to find the best beer in the city. It was also the place to find the best breweries in the city. It was also the place to find the best breweries in the country. In the bustling city, a bustling brewery was bustling with craftsmanship and craftsmanship. It was also the place to find the best beer in the city. It was als" \\
\hdashline[3pt/5pt]
\textbf{DialoGPT$_{\text{13k}}$} & "In the bustling city of Tharrington, where the city's charm is known, There's a thriving tavern called Tharrington's, Where the taverns thrive with a vibrant atmosphere. The tavern's owner, a skilled craftsman, built a taproom that was truly grand, With a taproom full of flavors, it was truly a sight. The tavern's atmosphere was so grand, it was like a vibrant symphony, With taverns and taverns, it was truly a sigh" \\
\midrule
\textbf{BART$_{\text{2k}}$} & "In the bustling city of New Zealand, a New Zealander named New Zealanders gathered at the New Zealand's New Zealand brewery, where New Zealandians gathered to discuss the New Zealander's New Zealanders' New Zealand heritage." \textbackslash" \textbackslash" \textbackslash" \textbackslash" " " " suscept " suscept ' ' char non"\\
\hdashline[3pt/5pt]
\textbf{BART$_{\text{13k}}$} & "In the village of Vineland, where Vineland's vineyard thrived, Vineland and Vineland thrived. Vineland flourished, as Vineland joined the scene, As Vineland added to the scene. Vinland's vineyards, a vibrant scene, Where Vineland was the scene." " \" " \" " thwart " thwart thwarted" \\
\midrule
\textbf{Flan-T5$_{\text{2k}}$} & "In the township of Westfield, a brewer named Westman brewed a brew in the townhouse, while Westfield's breweries brewed brews in the townhouses of Westfield." \\
\hdashline[3pt/5pt]
\textbf{Flan-T5$_{\text{13k}}$} & "In the quaint town of Greenland, where gin and lager are the mainstays, Craftsmen from Greenland and Greenland gather to create a gin and lager masterpiece. The gin and gin are renowned for their exceptional craftsmanship, While the gin and gin are renowned for their exceptional taste. Greenland's gin and gin are renowned for their exceptional" \\
\midrule
\textbf{ByT5$_{\text{2k}}$} & "In the village's villages, the village's villagers visit the village's villages, while Victorian villagers visit the village's villages. The village's villagers visit the village's villages, while Victorian villagers visit the village's villages. The village's villagers visit the village's villages, and the farmhouses of Victorian villages and villages in Victorian villages. The village's villagers visit the village's villages, while Victorian villagers visit the village's villages. The village's villagers visit the village's villages, all from Victorian villages, and Victorian villages are" \\
\hdashline[3pt/5pt]
\textbf{ByT5$_{\text{13k}}$} & "In the vicinity of the village, the farmers from the farmhouse filled the villages with their farmhouses and farmsteads, while the farmers from the farmhouse farms filled the villages with their farmhouses." \\
\midrule
\textbf{Baichuan$_{\text{13k}}$} & "[Breweries brew beer,]"\textbf{*18}\\
\midrule
\textbf{ChatGPT} & "In a rural brewery, they brewed a brew so unique, Where hops happily hopped and barley barreled sleek. With malty melodies and yeasty jigs, they thrived, But rural brewery’s ruckus rhythm rarely arrived.  The rural brewery’s brewmaster was a friendly fellow, Fermenting fizzy flavors, flowing with yellow. With each sip, the patrons picked passionately, Pouring pints, pretty pleased, with plenty of glee.  But with the twist of tongue and tricky tales, The rural brewery’s name became like snails. Try as they might, lips tangled and twisted, Rural brewery remained quite disjointed and resisted.  The rural brewery’s brand built on rustic charm, But their tongue twister seemed to cause alarm. With each attempt, patrons tangled like vines, Rural brewery’s name spinning in their minds.  So, remember to take your time and don’t hurry, When uttering words about a rural brewery. Though the tongue may twist and the speech may waver, The taste of their brew will make you a believer!"\\
\bottomrule
\end{tabular}
}
\caption{Example outputs from all fine-tuned models where different training data amounts were used (2k v. 13k training examples) in addition to Baichuan with 13k training examples, and zero-shot ChatGPT in the topic-to-twister task setting. Text within square brackets indicates repeated spans, with the number of repetitions indicated next to it.}
\label{tab:topic-case}
\end{table}


\autoref{tab:topic-case} presents example generations using 2k and 13k training samples to investigate the different benefits and drawbacks of each model, as well as the effects of variable levels of training data in a qualitative manner. We additionally include Baichuan trained on 13k samples, and ChatGPT (GPT-3.5-Turbo) in a zero-shot manner. For these examples, we randomly select an input topic from the first 100 samples in our test set.

\paragraph{\textbf{GPT-2}} Firstly, when considering the generations from GPT-2, the generation from the 2k training sample model does not demonstrate any clear phonetic patterns, with no prominent sound repetition present. However, some orthographic repetition can be observed, with "Chattanooga", "craft" and "scene" presenting 3 ways in which the graphemes <c> and <s> appear together (as <c> is often realized as \textipa{/s/}, like <s>, in words such as "celery"). On the other hand, the 13k training sample generation presents a much better tongue twister, demonstrating repetition of \textipa{/t/} in "\textit{\textbf{t}own of Cha\textbf{tt}anooga[...] Cha\textbf{tt}anooga's \textbf{t}aproom is filled with \textbf{t}apestries}". This is additionally complemented by the repetition of the affricate \textipa{/tS/} in the phrase "\textit{\textbf{Ch}attanooga's \textbf{ch}arm}". Whilst these tongue twisters appear quite successful, it is hard to ignore the fact that in both instances the models have resorted to the repetition of very similar clauses/sentences. However, all in all, it would appear that GPT-2$_{\text{13k}}$ has successfully generated a tongue twister that is largely grammatically coherent, demonstrates phonetic overlap, and is semantically related to the input (even if the rural nature of Chattanooga, a city in Tennessee, may be up for debate). Consequently, this first instance lends support to the proposed benefit of extended quantities of training data for the task of tongue twister generation as provided by our extension of TwistList 1.0 into TwistList 2.0, due to GPT-2$_{\text{13k}}$ demonstrating better performance than GPT-2$_{\text{2k}}$. This finding also supports our claims that the proposed TwisterLister pipeline creates high-quality tongue twisters.

\paragraph{\textbf{DialoGPT}} Regarding DialoGPT in the topic-to-twister setting, we see high levels of redundant repetition in the 2k training sample output, such as "bustling with craftmanship and craftmanship". However, some elements are clearly tongue twister-esque, such as the initial "\textit{In the bustling city of Bethlehem, a bustling brewery was bustling}", exploiting the voiced bilabial plosive, \textipa{/b/}. In the 13k training sample output, we curiously observe "Bethlehem" swapped for "Tharrington", and the exploitation of the voiceless dental fricative \textipa{/T/} also found in "thriving" and (of course) "thrive". Overall, DialoGPT demonstrates a more meaningful narrative-like generation alongside more training data, in addition to a change in primary phonemes potentially arising from the distributional properties of different phonemes across the larger training split.

\paragraph{\textbf{BART}} BART, on the other hand, demonstrates some less than desirable traits in the 2k setting, using extreme levels of repetition for "New Zealand" (and morphological variants) resulting in a poor quality output that only clearly represents the input topic via the inclusion of the word "breweries". Additionally, BART resorts to generating nonsense output towards the end, producing myriad punctuation and random words. In the case of 13k training samples, the output is less overtly repetitive and is more coherent (though still far from perfect), exploiting \textipa{/v/} and \textipa{/T/} (a voiced labiodental fricative and a voiceless dental fricative). For example, "\textit{In the \textbf{v}illage o\textbf{f} \textbf{V}ineland, where \textbf{V}ineland's \textbf{v}ineyard \textbf{th}rived, \textbf{V}ineland and \textbf{V}ineland \textbf{th}rived.}" However, the generation does exhibit significant breakdown towards the end, once again producing nonsensical output (though phonetically consistent) with "thwart" and variants thereof.

\paragraph{\textbf{Flan-T5}} Flan-T5 generates the shortest output seen across the models for this input. The generated tongue twister in the example only appears to engage with the repetition of \textipa{/b/} (rather than alongside a similar phoneme), but does so successfully, such as "\textit{[...] a \textbf{b}rewer named Westman \textbf{b}rewed a \textbf{b}rew in the townhouse, whilst Westfield's \textbf{b}reweries \textbf{b}rewed \textbf{b}rews[...]}". However, the semantic coherence of this output is lacking, with the discussion of 2 parallel events (``Westman, who is in Westfield, brewing, whilst breweries in Westfield also brew''). Additionally, the \textipa{/b/} repetition is arrived at exclusively through the exploitation of morphological variants of "brew", suggesting the signal that may have been picked up during training is on a morphological level, rather than orthographic or phonemic/phonetic (as <b> alone does not constitute a morpheme). With additional training data, the generation length changes significantly for Flan-T5. In the generation from Flan-T5$_\textit{13k}$, there is clear repetition of the voiced velar plosive \textipa{/g/} and the voiceless counterpart \textipa{/k/}, such as "\textit{\textbf{C}raftsmen from \textbf{G}reenland [...] \textbf{g}ather to \textbf{c}reate a \textbf{g}in and la\textbf{g}er}". However, unlike in the 2k example, the output here exhibits significant semantic redundancy as seen in phrases such as "\textit{The gin and gin [...] while the gin and gin}".

\paragraph{\textbf{ByT5}} ByT5, the largest of the models we train on numerous splits (at 582M parameters), performs poorly when considering the 2k training data split, with the output degrading into repetition. However, the repetition consists of full phrases and sentences rather than a single noun and maintains grammaticality throughout (albeit lacking in coherence). As for the version trained on 13k samples, this remedies the repetition issue. Interestingly, this version also demonstrates more clearly a phonemic pattern in the outputs, swapping between \textipa{/v/} in "\textit{\textbf{v}illager}" (and words with the same root) and \textipa{/f/} in "\textit{\textbf{f}armers}" (and words with the same root), which are voiceless/voiced counterparts of each other, and therefore an ideal pattern for a tongue twister to exploit. This is also demonstrated in the 2k version, but only weakly, with one instance of an \textipa{/f/}-initial word, "\textit{\textbf{f}armhouses}".

\paragraph{\textbf{Baichuan}} Baichuan, as the largest model we train (exclusively on the 13K training data split) with 7B parameters, performs rather poorly in the topic-to-twister setting. The output shown here consists exclusively of a single phrase "Breweries brew beer" repeated 18 times. Whilst this is a valid tongue twister in terms of relevance to the input, Baichuan is shown to very quickly get stuck in a loop, degrading the quality of the overall output. This pattern is seen frequently across other outputs from Baichuan.

\paragraph{\textbf{ChatGPT}} Finally, ChatGPT (GPT-3.5-Turbo) presents the longest tongue twister of the examples listed. However, this is an exception for this randomly selected generation, rather than the rule, as seen in the automatic evaluation results, where the average generation length for ChatGPT was 17.43 words. In terms of quality, ChatGPT excels at generating well-formed grammatical text, which is shown to be the case in the example generation. Additionally, the generation exploits numerous different phonemic patterns, including repetition of \textipa{/h/} "... \textit{\textbf{h}ops \textbf{h}appily \textbf{h}opped}", \textipa{/b/} "\textit{In a rural \textbf{b}rewery they \textbf{b}rewed a \textbf{b}rew[...]}", and \textipa{/\*r/} "\textit{But \textbf{r}u\textbf{r}al b\textbf{r}ewe\textbf{r}y's \textbf{r}uckus \textbf{r}hythm \textbf{r}a\textbf{r}ely a\textbf{rr}ived}". Additionally, one demonstrated ability of ChatGPT that is not exploited by any of the other fine-tuned models (due to the fine-tuning focus) is the incorporation of additional literary techniques based on speech sounds, such as rhyme (e.g., "thr\textbf{ived}"/"arr\textbf{ived}" and "t\textbf{ails}"/"sn\textbf{ails}". In addition, the coherence level of the ChatGPT output is also very significant, with the output demonstrating a clear narrative that would engage readers. Consequently, we can see why ChatGPT may score poorly on phoneme-based metrics (cf., \autoref{tab:additional-evaluation-topic}) due to incorporating a lot of words to enhance the grammar of the tongue twister, as well as exhibiting low-level local repetition of a particular sound in a phrase, rather than maintaining the focus on the same sound for the entirety of the output.\\

Overall, most models appear to benefit from the presence of increased levels of training data in this example, particularly regarding producing more grammatical and coherent output. The only model for which this observation does not hold is Flan-T5, which does not have a clear change in generation quality from 2k to 13k training samples. This is in some respects to be expected, as larger models are often able to abstract away language patterns more easily than smaller models from the same amount of training data (therefore resulting in a law of diminishing returns between training data quantity and model size). However, the smaller models demonstrate the utility of our TwistList 2.0 dataset in presenting a sufficiently large increase in training data to result in tongue twister generations of demonstrably improved quality. This also shows that there is room to improve generated tongue twisters with an increase in training data exclusively before the requirement for more complex training paradigms is necessary.

\subsection{Topic-to-Twister (PACD)}

In contrast with the previous approach, our constrained decoding algorithm involves the random selection of an initial phoneme and nearest neighbor, through which to condition generation. Consequently, numerous tongue twisters can be generated for a single input, as long as the selected phoneme pairs differ each time. For this reason, we present an additional case study of model outputs using our Phoneme Aware Constrained Decoding (PACD) module.

\begin{table}[h!]
\centering \small
\resizebox{\linewidth}{!}{
\begin{tabular}{p{0.2\linewidth}|p{0.80\linewidth}}
\toprule
\textbf{Input:} &  "rural brewery" \\
\textbf{Golden:} & "Rory the warrior and Roger the worrier were reared wrongly in a rural brewery."\\
\midrule
\textbf{GPT-2$_\text{13k}$} & "In the bustling town of Chattanooga, where the townspeople reside, The Chattanooga's taproom is filled with tapestries of flavors. Chattanooga's taproom is filled with the finest tapestry, From the tapestry of Chattanooga's charm, it's a sight to see. Chattanooga's taproom is filled with the finest tapestry, From the tapestry of Chattanooga's charm, it's a sight to see. Chattanooga's taproom is filled with the finest tapes"\\
\midrule
\textbf{GPT-2$_\text{13k}$ -w} & "Brewers brew beer in the barn brewing process before brewers bring their beers to the bar boys breweries prior to being brewed beforehand but before being bottled previously because of"\\
\hdashline[3pt/5pt]
& "With their rust wood working wonders with their water works while watering wheat wine wines within their win ware warehouses where they are waiting for the weather to warm up"\\
\hdashline[3pt/5pt]
\textbf{GPT-2$_\text{13k}$ -ws} & "Brewers brew beer in the breweries belonging to both baker boys and bar patrons but the bartender prefers the beers brewed by bud brothers and bro pal pit bulls before"\\
\hdashline[3pt/5pt]
& "With a rust wood wind ward rip weed from the woods while watering the water works well within the walls of the warehouse where the work room is wall rolling"\\
\midrule
\textbf{GPT-2 -w } & "But before you do begin building your own brewery please bear in place that you are not a brewer but a person who has been brewing for a bit because
"\\

\hdashline[3pt/5pt]
& "What is the reason for this one with the word rural rather than regional where it is written when it is referring to the region of the world where the"\\
\hdashline[3pt/5pt]
\textbf{GPT-2 -ws} & "But I believe that the best place to begin is with the basics before you proceed beyond that point because the more you practice brewing beer properly both in the"\\
\hdashline[3pt/5pt]
& "What is the reason for this request with the word rural when you are referring to a region that is not a regionally relevant one where there is a real
"\\
\midrule
\textbf{Baichuan$_\text{13k}$} & "[Breweries brew beer]"\textbf{*18}"\\
\midrule
\textbf{Baichuan$_\text{13k}$ -w} & "Try to drink tea in the town during dinner time drinking delicious drinks from the tap tasting different types of tasty treats trying to taste the differ den between two"\\
\hdashline[3pt/5pt]
& "What is the relations with rural roads running through resident rial ways within regional regions where rivers run through rocky rid rig road way runs through region rich river routes"\\
\hdashline[3pt/5pt]
\textbf{Baichuan$_\text{13k}$ -ws} & "Try to drink tea in the township of taverna towns during the daytime drinking tour dates downtown drinker drinks dry den dutch town tavern a day time tasting tours tourist"\\
\hdashline[3pt/5pt]
& "What is the relationship between the rural region and the regional railway route running through rut rum runners runs through rocky regions with rich resources while retaining its rustic roots"\\
\midrule
\textbf{Baichuan -w} & "Try to twist the tongue as tight together as time does during the day today due to the temper ter differ ten times too different degrees depending on the direction"\\
\hdashline[3pt/5pt]
& "Write a one word response to each of the words below with a relevant rural reference within the reply which is not a repeated ref rain water recycling wind renewable"\\
\hdashline[3pt/5pt]
\textbf{Baichuan -ws} & "Try to twist the tongue as tightly as desired during the time of delivery trying to deliver the twistiest tong toe track ter day till the target is delivered tight"\\
\hdashline[3pt/5pt]
& "Write a one word response to each of the words below with a relevant rural reference within the reply which is not a repeated ref rain water recycling rainwater runoff"\\
\bottomrule
\end{tabular}
}
\caption{Example outputs from various models using our novel constrained decoding algorithm for tongue twister generation. -w denotes models using our PACD decoding algorithm on full words only, whilst -ws indicates the subword version of PACD. We present 2 generations for each model using our PACD decoding approach. Each example exploits the phonemes \textipa{/b, p/} and \textipa{/\*r, w/}, respectively. To achieve this, we overwrite the initial phoneme selection process of $ph_1$ in the PACD algorithm in order to present diverse outputs from the same input (as \textipa{/b, p/} would be selected exclusively otherwise).}
\label{tab:constrained_case}
\end{table}

\paragraph{\textbf{GPT-2$_\text{13k}$ -w/o}}
As discussed in \S\ref{sec:unconstrained_case_topic}, GPT-2 finetuned on 13k samples without the addition of our constrained decoding module demonstrates repetition of \textipa{/t/} in "\textit{\textbf{t}own of Cha\textbf{tt}anooga[...] Cha\textbf{tt}anooga's \textbf{t}aproom is filled with \textbf{t}apestries}". This is additionally complemented by the repetition of the affricate /\textipa{tS}/ in "\textbf{Ch}attanooga's \textbf{ch}arm".

\paragraph{\textbf{GPT-2$_\text{13k}$ -w/-ws}} On the other hand, finetuned GPT-2 with the additional constraints imposed by our PACD module demonstrates several different characteristics. Firstly, as is the case with the non-finetuned model, our constrained generations are limited in length to 30 tokens, making all generations of equal length, and consequently shorter than the output of GPT-2 without these constraints. Regarding the tongue twisters, each example can be seen to be related to the input keywords, but vary between direct and abstract relations. For example, the first generation which enforced the selection of tokens starting with \textipa{/b/} or \textipa{/p/} demonstrates a clear direct reference to the input word "brewery", something which has been afforded to the model as the selected initial phonemes match one of the word-initial phonemes of the input (which is the approach we take in \S\ref{sec:human_eval_constrained_}).

Lastly, the generations exploiting \textipa{/\*r/} and \textipa{/w/} are similarly abstract (but perhaps to a lesser extent), referring to relevant words such as "water", "wine", and "wood working" (the latter activity being more expected in a rural locale). Regarding grammatically, the \textipa{/b, p/} example is largely grammatical (if "boys" had a possessive apostrophe, to create "the bar boys' breweries" as a noun phrase") but is cut off at 30 tokens, resulting in an incomplete sentence. However, semantically it is hard to follow due to the high level of temporal adverbs used.
Likewise, the \textipa{/\*r, w/} generation is mostly grammatical, with the presence of numerous relative clauses leading to difficulty in understanding (however, this does require some liberty in accepting that "wheat wine wines" and "win ware warehouses" are allowable compound nouns). With the subword loop turned on for GPT-2 using PACD, we see similar outputs in style to the full-word version. However, one difference we observe is a slight reduction in grammaticality towards the end of the \textipa{/b,p/} version, with the phrase "and bro pal put bulls before", which is difficult to parse. Similar effects can be seen with the \textipa{/\*r, w/} version, where the outputs are similar in word choice (as expected, given some whole words will form the stem of longer words in the subword version) and the overall result is perhaps slightly more difficult to parse grammatically.

\paragraph{\textbf{GPT-2 -w/-ws}} Regarding non-finetuned base GPT-2 with the addition of our module, all generations demonstrate similar levels of grammaticality to those seen in our finetuned model. On the other hand, the generations can be considered less literary, with the /b, p/ example suggesting the model is generating instructional text in "\textit{But before you do begin building your own brewery[...]}". Similarly to previously discussed generations with the PACD module, all of these generations are hindered by the 30-token limit, with examples ending in "because", "his desire to try" (which suggests the verb "try" will take an additional argument) and "the". Overall, similar results are seen with the subword loop enabled (-ws), resulting in variations of the same output.

\paragraph{\textbf{Baichuan$_\text{13k}$}} As discussed previously, finetuned Baichuan without the addition of PACD produces a valid 3-word tongue twister phrase that achieves alliteration of \textipa{/b/}, but quickly falls into the degenerate pattern of repeating this phrase 18 times, rather than continuing to extent the tongue twister in a unique and entertaining fashion.

\paragraph{\textbf{Baichuan$_\text{13k}$ -w/-ws}} With the addition of full-word PACD to the finetuned version of Baichuan, we successfully avoid the repetition trap (due to repetition restrictions at decoding time). For the first example, the enforcement of \textipa{/t, d/} is evident in grammatical phrases such as \textit{"\textbf{T}ry \textbf{t}o \textbf{d}rink the \textbf{t}ea in the \textbf{t}own \textbf{d}uring \textbf{d}inner \textbf{t}ime..."}, and is continued throughout. Overall, the first generation is grammatical, only demonstrating a clear degradation towards the end with the generation of "differ den" which hinders fluency and coherence. On the other hand, in the \textipa{/\*r, w/} example, the sound overlap is demonstrated, but the coherence and fluency of the output are shown to suffer much earlier, with sequences such as \textit{"rivers run through rocky rid rig road ways runs through region rich river routes".} An overall pattern can be seen that generation quality begins to drop after a single suboptimal token is selected due to the detrimental impact it has on the following token's prediction. When comparing to the subword enabled PACD (-ws) we see the initial difference in the 7th word of the \textipa{/t, d/} example, where "town" has been extended to "township", resulting in the remainder of the generation diverging from that of the full-word version of PACD, which is similarly reflected in the \textipa{/\*r, w/} example as "relation" becomes "relationship". Overall, due to the constraints in place, both present difficult-to-articulate sequences, with the overall quality between subword and full-world versions being minimal and subjective.
\paragraph{\textbf{Baichuan -w/-ws}} Regarding base Baichuan with the addition of PACD, we see a lack of topical relevance in the \textipa{/t, d/} examples for both full-word and subword versions, with the input of "rural brewery" not being reflected in the semantics of the output (although the generation has noted the request for a tongue twister in the starting stem, and references it directly). This is also seen in the \textipa{/\*r,w/} versions but to a lesser extent, as the topic word "rural" starts with an allowable phoneme, \textipa{/\*r/}, causing it to generate in the phrase "\textbf{r}ural \textbf{r}eference". In this example, therefore, we see that finetuned Baichuan is better able to understand the desired topic (due to being trained on the topic-to-twister setting), whilst the base model produces overtly generic text, but with repetitive phonology.
\\

Overall, the examples demonstrate the effectiveness of the simple constrained decoding module that constitutes PACD. In the examples generated through this module, the ability to control token repetition means we avoid the pitfalls of standard finetuned models that frequently default to phrasal level repetition, ensuring that tongue twisters generated with PACD use a wide range of vocabulary. Additionally, due to the constraints being phoneme-based, we can exercise control over the present sounds even when permitting the generation of a small set of function words to facilitate grammatical correctness. To this end, similar to the pipeline of TwisterLister (\S\ref{sec:twisterlister_pipe}), PACD allows us to ensure phoneme overlap within the output, rather than relying on this signal being picked up during fine-tuning. The case studies additionally demonstrate that rule-based decoding of a pre-trained language model can result in tongue twisters that are comparable to those of finetuned models, and that future work in the area should make extensive use of constrained decoding approaches such as PACD, rather than exclusively finetuning on input-output pairs.

\subsection{Style-Transfer}

We now qualitatively investigate the generations from our additional task setting using style-transfer. \autoref{tab:style-case} presents example generations across 2k training samples and 13k training samples in order to investigate the different benefits and drawbacks of each model, as well as the effects of variable levels of training data in a qualitative manner. We further include Baichuan trained on 13k samples, and zero-shot ChatGPT (GPT-3.5-Turbo).\\

As demonstrated in the \textit{Input} and \textit{Gold Standard} fields, the generated paraphrase is not easily identifiable as a tongue twister and is, therefore, a prime example of a successful paraphrase to enable style-transfer (as the twisted style has been removed in order to have parallel data to aid in identifying the desired style to transfer from the source to the target), whilst the gold standard reference from which the paraphrase was derived exploits the common articulation difficulty and confusion between the rhotic \textipa{/\*r/} and lateral \textipa{/l/} \cite{Georgievska-Jancheska2019-qh}, encouraging metathesis (the accidental transposition of speech sounds across words).

\begin{table}[h!]
\centering \small
\resizebox{\linewidth}{!}{
\begin{tabular}{p{0.2\linewidth}|p{0.80\linewidth}}
\toprule
\textbf{Input:} &  "The small crimson truck traveled along Limuru street." \\
\textbf{Golden:} & "The little red lorry went down Limuru road."\\
\midrule
\textbf{GPT-2$_{\text{2k}}$} & "The small carriages of the small carriages, with their small wheels, were moving along Limuru's road."The small yellow truck traveled along Limuru's road. The yellow cars of the yellow cars, with their yellow wheels, were traveling along Limuru's road.""The yellow yellow truck traveled along Limuru's road." The yellow cars of the yellow cars, with their yellow wheels, were traveling along Limuru's road.""The yellow yellow truck traveled along Limuru's road
" \\
\hdashline[3pt/5pt]
\textbf{GPT-2$_{\text{13k}}$} & "The small red truck, with its small red tires, was a sight to see. The small red truck, with its small red tires, was a sight to see. The small red truck, with its small red tires, was a sight to see. The small red truck, with its small red tires, was a sight to see. The small red truck, with its small red tires, was a sight to see. The small red truck, with its small red tires, was a sight" \\
\midrule
\textbf{DialoGPT$_{\text{2k}}$} & "It passed through the town of Limburg, where the small town of Limburg's inhabitants were located. It passed through the town of Limburg, where the town of Limburg's inhabitants were located. It passed through the town of Limburg, where the town of Limburg's inhabitants were located. It passed through the town of Limburg, where the town of Limburg's inhabitants were located. It passed through the town of Limburg, where the town of Limburg's inhabitants were located
" \\
\hdashline[3pt/5pt]
\textbf{DialoGPT$_{\text{13k}}$} & "It passed by the small houses and the small houses' small houses, while the small houses' small houses and the small houses' small houses were observed."The minuscule minivan passed by the mansions and mansions' mansions, while the mansions' mansions and the mansions' mansions were seen." Improved version: The minuscule minivan passed by the mansions and mansions' mansions, while the mansions' mansions and the mansions" \\
\midrule
\textbf{BART$_{\text{2k}}$} & "The limo journeyed along Limuru thoroughfare." " churn " churned B hollow f un \" umbled inc ops op edy - standard r unilaterally worsened thumb pl thrust res ed churn" \\
\hdashline[3pt/5pt]
\textbf{BART$_{\text{13k}}$} & ""The little red limousine traveled down Limuru thoroughfare." \textbackslash" \textbackslash" \textbackslash" \textbackslash"\textbackslash" \textbackslash" \textbackslash"\textbackslash" \textbackslash" \textbackslash uff \textbackslash uff \textbackslash" \textbackslash" \textbackslash" L " t \textbackslash", ' churn M   L \textbackslash" " uff \textbackslash churn" \\
\midrule
\textbf{Flan-T5$_{\text{2k}}$} & "The yellow limo limped along Limuru limoway." \\
\hdashline[3pt/5pt]
\textbf{Flan-T5$_{\text{13k}}$} & "The little lilac lorry limped down Limuru lanes." \\
\midrule
\textbf{ByT5$_{\text{2k}}$} & "The little limuru truck looked along Limuru street." \\
\hdashline[3pt/5pt]
\textbf{Byt5$_{\text{13k}}$} & "The little red truck wandered along Limuru thoroughfare." \\
\midrule
\textbf{Baichuan$_{\text{13k}}$} & "Limuru's little red truck drove along Limuru road."\\
\midrule
\textbf{ChatGPT} & "Truck crimson, small, along Limuru street did travel."\\
\bottomrule
\end{tabular}
}
\caption{Example outputs from all fine-tuned models where different training data amounts were used (2k v. 13k) in addition to Baichuan with 13k training examples, and zero-shot ChatGPT in the style-transfer task setting. Please note that the mass majority of ChatGPT outputs resemble tongue twisters, whilst this randomly selected input topic generated twisted syntax in this particular case.}
\label{tab:style-case}
\end{table}

\paragraph{\textbf{GPT-2}} Firstly, the main element to notice with either generation from GPT-2 is the excessive length. This is particularly startling when you consider the nature of style-transfer, where the ideal tongue twister version of a text would be of similar length to the original. However, upon further inspection, it is clear that GPT-2 resorts to blatant repetition of very similar content to constitute the delta in length. Consequently, if we only take the first sentence of the 2k version as the output "\textit{The small carriages of small carriages, with their small wheels, were moving along Limuru's road."}, the output can more clearly be seen as a style-transferred version of the input. However, this generation is still poor, with little clear phoneme-level repetition (excluding repetition of the same word, such as "small" in the quoted passage) and much semantic redundancy. Regarding the 13k training example output, GPT-2 yet again exhibits large levels of repetition, but in this case, the repetition is purer, being a verbatim repetition of the initial sentence numerous times. In addition, the output is free of any grammatical errors. In this example, there is clear use of the sibilant \textipa{/s/} in "\textit{The \textbf{s}mall red truck with it\textbf{s} \textbf{s}mall red tires, was a \textbf{s}ight to \textbf{s}ee"}. Additionally, the input semantics of a small truck have been maintained, but the location (Limuru, a town in Kenya) has been lost.

\paragraph{\textbf{DialoGPT}} On the other hand, the 2k version of the dialogue fine-tuned GPT-2, DialoGPT lacks semantic coherence due to using deixis in the form of "it", which is not resolved as a cataphoric reference later in the text, making "it" ambiguous. A potential attempt at phoneme repetition is present with "Limburg" and "located", but this is tenuous. Overall, the generation with 2k training examples does not clearly present a tongue twister. On the other hand, the extended 13k example demonstrably resembles a tongue twister, achieving repetition of \textipa{/s/}  (e.g., "\textit{\textbf{S}mall houses and the \textbf{s}mall houses' \textbf{s}mall houses}') and \textipa{/m/} (e.g., \textit{`The \textbf{m}inuscule \textbf{m}inivan passed by the \textbf{m}ansion and the \textbf{m}ansions' \textbf{m}ansions}"). However, this generation also exhibits a strange structure, including "improved version" as part of the output. If only a subset of this generation is considered, \textit{"The minuscule minivan passed by the mansion and the mansions' mansions"}, the output constitutes a high-quality tongue twister. Whilst it may be semantically bizarre, to discuss the ownership of mansions by other mansions, it is not grammatically invalid, and human-authored tongue twisters also frequently convey unusual semantics to increase their strange nature.

\paragraph{\textbf{BART}} Unlike the GPT-2 based models, BART produces style-transferred versions of the input that are much closer in length to the original input, even without needing to exclude sentence-level repetition. In the 2k example, the initial sentence "\textit{The limo journeyed along Limuru thoroughfare}" is a grammatically and semantically valid output, but does not resemble a tongue twister outside of the lateral \textipa{/l/} in both "limo" and "Limuru". However, the sentence following this consists primarily of noise. On the other hand, alongside the increase in training data comes an improvement in tongue twister quality, with "\textit{The \textbf{l}ittle \textbf{r}ed \textbf{l}imousine trave\textbf{l}ed down \textbf{L}imuru thoroughfare}" appearing to exploit the articulatory similarity between \textipa{/l/} and \textipa{/\*r/} without relying on single-word repetition. Again, however, the output devolves into noise towards the end, consisting of various punctuation and subwords.

\paragraph{\textbf{Flan-T5}} Flan-T5 (alongside ByT5) presents the most clear paraphrases of the desired output, containing no unnecessary repetition or noise. In the 2k example, Flan-T5 repeats similar phonemes, exploiting the similar phonetic categories, laterals and glides (or semivowels), of which \textipa{/l/} and \textipa{/j/} belong to, respectively, "\textit{The \textbf{y}e\textbf{ll}ow \textbf{l}imo \textbf{l}imped a\textbf{l}ong \textbf{L}imuru \textbf{l}imoway.}". On the other hand, from more training examples, Flan-T5 presents a paraphrase that is more faithful to the original input, relying on the repetition of \textipa{/l/} exclusively: "\textit{The \textbf{l}itt\textbf{le} \textbf{l}i\textbf{l}ac \textbf{l}orry \textbf{l}imped down \textbf{L}imuru \textbf{l}anes.}" (assuming we take "lorry" to be more semantically related to "truck" than "limousine" is).

\paragraph{\textbf{ByT5}} ByT5 shows equivalent performance to Flan-T5 in this example, opting to alliterate different parts of speech to Flan-T5, but overall having a similar effect in the 2k example. On the other hand, the 13k version doesn't present alliteration as clearly as Flan-T5, but does alternate between related phonemes, \textipa{/l, \*r, w/}, in "\textbf{l}ittle \textbf{r}ed truck \textbf{w}andered".

\paragraph{\textbf{Baichuan}} In the style-transfer task setting, Baichuan can be seen to perform much better than witnessed in the previous topic-to-twister task setting. Similar to other generations, Baichuan exploits the \textipa{/\*r/} and \textipa{/l/} similarity with "\textit{\textbf{L}imu\textbf{r}u's \textbf{l}ittle \textbf{r}ed t\textbf{r}uck d\textbf{r}ove a\textbf{l}ong \textbf{L}imu\textbf{r}u \textbf{r}oad'}', where every word of the output contains one of these sounds, and alternation is frequent. 

\paragraph{\textbf{ChatGPT}} Finally, ChatGPT in a zero-shot setting appears to strongly misinterpret the given prompt, outputting non-standard syntax such as swapping the order of an adjective and noun in "truck crimson" (reminiscent of the speaking style of Star Wars' Yoda). Additionally, the ChatGPT generation does not appear to reflect a tongue twister in any clear sense. It is important to note, however, that this is not a common pattern with ChatGPT outputs, but is the case for the randomly selected case study example.
\\

Overall, as with the topic-to-twister setting, we can see a clear benefit in the style-transfer task formulation of using more training data, with all models producing better quality outputs both in terms of phonetic patterns, grammatical validity, and semantic coherence. An overarching theme, however, is that several models misinterpret the desire to paraphrase a single sentence as a tongue twister, instead producing outputs that are very long. However, if edited in post (often simply by taking the first sentence), these generations are often still valid (if not perfect).

\section{Conclusion}

In this article, we have presented multiple novel contributions towards the development of more phonetically- and phonologically-aware models for the task of tongue twister generation. We presented a pipeline for the generation of tongue twisters at scale using large language models that encourage uniqueness and non-derivative examples through the careful selection of a candidate vocabulary (TwisterLister) to develop a large dataset of machine-generated tongue twisters (TwistList 2.0). We then finetuned a series of smaller language models on the resulting dataset and observed that the topic-to-twister and style-transfer task settings for tongue twister generation exhibit different characteristics when considering the benefit of additional training data, as well as observing that different models perform differently in the two task settings. These findings demonstrate a fundamental difference in the requirements of the two approaches and further motivated the need for additional training data as supplied by TwistList 2.0 through the use of the TwisterLister pipeline. We additionally presented a novel algorithm (PACD) that implements hard lexical constraints based on the phonemic characteristics of words that can be used to realize tongue twisters from a causal autoregressive language model by accessing the next token predictions and applying a cascade of filters. We then extensively evaluated the generations from our proposed approaches, presenting both automatic and human evaluation. In the former, we additionally presented 2 novel metrics for measuring the sound complexity of a generated tongue twister based on the concept of phoneme edit distance (iPED and oPED). With these fine-tuned models and constrained decoding module, we then provided an in-depth exploration of the generation characteristics through case studies investigating the propensity of each model to generate high-quality tongue twisters, as well as the differences seen via the addition of more training data (to move away from reliance on automatic metrics that do not reveal fundamental qualitative differences). Overall, we find that straight fine-tuning of language models for a tongue twister generation task still has substantial room for improvement to meet human-authored standards. However, we additionally demonstrate that simple constrained decoding approaches are able to generate better tongue twisters than only finetuned models, particularly due to the phoneme-level awareness which allows for more difficult-to-articulate sound combinations to be present in an output. We additionally envision the techniques and approaches presented herein to be beneficial to the creative NLG community, particularly in regard to the generation of phonemically conditioned language forms (e.g., poetry, lyrics, and puns) to explore methods of constraining token outputs whilst simultaneously taking advantage of the power of modern LLMs.

We hope to witness increased interest in the area of tongue twister generation, as well as other niche areas of creative language generation that pique the interest of newcomers to the NLG domain, as well as people from wider domains such as literature and (non-computational) linguistics, and the general public (where creative language generation may offer a more accessible and intriguing entry into the NLP and machine learning communities). In furthering work in this area, we believe reinforcement learning approaches may prove fruitful, in addition to incorporating a differentiable version of something akin to phonemic edit distance as an additional loss function to optimize. Furthermore, whilst we present new metrics (oPED/iPED), we encourage the development of more robust general metrics for tongue twisters and other forms of creative language that can adequately balance the requirements of being grammatical as well as exhibiting significant levels of sound repetition.

\appendix
\appendixsection{Evaluation Rubric}
\label{apx:evaluation-rubric}
Below is the evaluation rubric presented to human evaluators. The prompts have the following format: \texttt{"\textbf{[criterion description]}\textbackslash n Instruction: Generate a tongue twister relating to \textbf{[input]}\textbackslash n Response: \textbf{[output]}\textbackslash n Rate the response from 1 to 5:\textbackslash n \textbf{[rubric]}"} Where \textit{criterion description} refers to the first line of the examples below (e.g., "Is the model proficient in..."), \textit{[input]} is the input to the model (either a topic in the topic-to-twister setting, or standard non-literary text in the style-transfer setting), \textit{[output]} is an LM's response to a given input topic/text, and \textit{[rubric]} is the 5-point rating system as outlined below.\\

\noindent\textbf{Relevance}\\
Is the model proficient in generating text that is relevant to the input topic?\\
1 - The model completely ignores the input topic and generates irrelevant text.\\
2. The model generates text that is mostly irrelevant to the input topic, with minimal and unclear association.\\
3. The model generates text that is partially relevant to the input topic, but the association is weak and inconsistent.\\
4. The model generates text that is mostly relevant to the input topic, with clear association but occasional lapses.\\
5 - The model excels in generating relevant text, where the responses are consistently on topic and the association is clear.\\

\noindent\textbf{Difficulty of Articulation}\\
Is the model proficient in generating text that is difficult to pronounce due to alliteration and phonetic complexity?\\
1.  The model generates text that is no harder to pronounce and articulate than standard writing.\\
2.  The model generates text that is slightly more challenging to say than standard writing, demonstrating some simple techniques such as alliteration.\\
3. The model generates text that is somewhat difficult to say but clearly exhibits techniques such as alliteration.\\
4. The model generates text that is generally difficult to say, with techniques such as alliteration, but also alternating between similar sounds.\\
5. The model generates text that is highly phonetically complex and difficult to say, consistently exploiting repetition of closely related sounds and alliteration.\\

\noindent\textbf{Fluency}\\
Is the model proficient in generating grammatical and well-formed text?\\
1. The model produces text with no grammatical phrases or spans.\\
2. The model generates text that is largely ungrammatical but with some grammatically valid sequences.\\
3. The model generates text that contains grammatically valid sequences but overall is difficult to parse.\\
4. The model generates text that is generally grammatical and well-formed, with minor errors or awkward phrasing.\\
5. The model produces text that is grammatically correct and well-formed, demonstrating a strong command of the English language.\\

\noindent\textbf{Coherence}\\
Is the model proficient in generating semantically coherent and logical text?\\
1. The model neglects to generate semantically coherent text, producing text that is nonsensical in meaning.\\
2. The model generates text that is mostly incoherent, with only occasional hints of logical meaning.\\
3. The model generates text that is partially coherent, but the text lacks logical structure and consistency.\\
4. The model generates text that is generally coherent, with a logical structure and clear meaning, though minor inconsistencies may be present.\\
5. The model excels in generating semantically coherent text, where the text is logically structured and maintains a clear and consistent meaning.\\

\noindent\textbf{Entertainment}\\
Is the model proficient in generating text that a human reader would find entertaining or amusing?\\
1. The model demonstrates no creativity, either in the content or the structure of the text, resulting in uninteresting or unamusing outputs.\\
2. The model generates text with minimal creativity, resulting in outputs that are only slightly interesting or amusing.\\
3. The model generates text that is somewhat entertaining or amusing, but the creativity in content and structure is limited.\\
4. The model generates text that is generally entertaining and amusing, showing noticeable creativity in content and structure, though some outputs may be less engaging.\\
5. The model excels in creating entertaining and amusing text, demonstrating creativity in both content and structure and consistently producing engaging and enjoyable outputs.\\

\noindent\textbf{Overall Quality} \\
Is the model proficient in generating high quality English tongue twisters?\\
1. The model fails to generate high quality text that is recognisable as a tongue twister.\\
2. The model generates text that slightly resembles a tongue twister.\\
3. The model generates text that is recognisable as a tongue twister, but is lacking in fluency, coherence, or entertainment value.\\
4. The model generates text that is easily recognisable as a tongue twister, but may fall short of high quality.\\
5. The model excels in generating texts that resemble high-quality tongue twisters that are difficult to pronounce, make sense, and are entertaining.\\

\begin{acknowledgments}
Tyler Loakman is supported by the Centre for Doctoral Training in Speech and Language Technologies (SLT) and their Applications funded by UK Research and Innovation [grant number EP/S023062/1]. Chen Tang is supported by the China Scholarship Council (CSC) for his doctoral study (File No.202006120039).
\end{acknowledgments}

\starttwocolumn
\bibliography{main}

\begin{thebibliography}{92}
\expandafter\ifx\csname natexlab\endcsname\relax\def\natexlab#1{#1}\fi

\bibitem[{Anil et~al.(2023)Anil, Borgeaud, Wu, Alayrac, Yu et~al.}]{gemini}
Anil, Rohan, Sebastian Borgeaud, Yonghui Wu, Jean-Baptiste Alayrac, Jiahui Yu, et~al. 2023.
\newblock Gemini: A family of highly capable multimodal models.

\bibitem[{Askari et~al.(2023)Askari, Aliannejadi, Meng, Kanoulas, and Verberne}]{askari-etal-2023-expand}
Askari, Arian, Mohammad Aliannejadi, Chuan Meng, Evangelos Kanoulas, and Suzan Verberne. 2023.
\newblock Expand, highlight, generate: {RL}-driven document generation for passage reranking.
\newblock In \emph{Proceedings of the 2023 Conference on Empirical Methods in Natural Language Processing}, pages 10087--10099.

\bibitem[{Brown et~al.(2020)Brown, Mann, Ryder, Subbiah, Kaplan, Dhariwal, Neelakantan, Shyam, Sastry, Askell, Agarwal, Herbert{-}Voss, Krueger, Henighan, Child, Ramesh, Ziegler, Wu, Winter, Hesse, Chen, Sigler, Litwin, Gray, Chess, Clark, Berner, McCandlish, Radford, Sutskever, and Amodei}]{GPT-3}
Brown, Tom~B., Benjamin Mann, Nick Ryder, Melanie Subbiah, Jared Kaplan, Prafulla Dhariwal, Arvind Neelakantan, Pranav Shyam, Girish Sastry, Amanda Askell, Sandhini Agarwal, Ariel Herbert{-}Voss, Gretchen Krueger, Tom Henighan, Rewon Child, Aditya Ramesh, Daniel~M. Ziegler, Jeffrey Wu, Clemens Winter, Christopher Hesse, Mark Chen, Eric Sigler, Mateusz Litwin, Scott Gray, Benjamin Chess, Jack Clark, Christopher Berner, Sam McCandlish, Alec Radford, Ilya Sutskever, and Dario Amodei. 2020.
\newblock Language models are few-shot learners.
\newblock \emph{CoRR}, abs/2005.14165.

\bibitem[{Buciluǎ, Caruana, and Niculescu-Mizil(2006)}]{bucilua_compression}
Buciluǎ, Cristian, Rich Caruana, and Alexandru Niculescu-Mizil. 2006.
\newblock Model compression.
\newblock In \emph{Proceedings of the 12th ACM SIGKDD International Conference on Knowledge Discovery and Data Mining}, KDD '06, page 535–541, Association for Computing Machinery, New York, NY, USA.

\bibitem[{Chakrabarty et~al.(2023)Chakrabarty, Laban, Agarwal, Muresan, and Wu}]{chakrabarty2023art}
Chakrabarty, Tuhin, Philippe Laban, Divyansh Agarwal, Smaranda Muresan, and Chien-Sheng Wu. 2023.
\newblock Art or artifice? large language models and the false promise of creativity.

\bibitem[{Chall and Dale(1995)}]{chall1995readability}
Chall, Jeanne~Sternlicht and Edgar Dale. 1995.
\newblock \emph{Readability revisited: The new Dale-Chall readability formula}.
\newblock Brookline Books.

\bibitem[{Chang et~al.(2023)Chang, Zhang, Jiang, Chen, Zhang, and Pu}]{chang-etal-2023-sudowoodo}
Chang, Yongzhu, Rongsheng Zhang, Lin Jiang, Qihang Chen, Le~Zhang, and Jiashu Pu. 2023.
\newblock Sudowoodo: A {C}hinese lyric imitation system with source lyrics.
\newblock In \emph{Proceedings of the 2023 Conference on Empirical Methods in Natural Language Processing: System Demonstrations}, pages 99--105.

\bibitem[{Chen et~al.(2021)Chen, Shu, Takamura, and Nakayama}]{chen-etal-2021-graphplan}
Chen, Hong, Raphael Shu, Hiroya Takamura, and Hideki Nakayama. 2021.
\newblock {G}raph{P}lan: Story generation by planning with event graph.
\newblock In \emph{Proceedings of the 14th International Conference on Natural Language Generation}, pages 377--386.

\bibitem[{Chiang and Lee(2023)}]{chiang-lee-2023-synonym}
Chiang, Cheng-Han and Hung-yi Lee. 2023.
\newblock Are synonym substitution attacks really synonym substitution attacks?
\newblock In \emph{Findings of the Association for Computational Linguistics: ACL 2023}, pages 1853--1878.

\bibitem[{Chung et~al.(2022)Chung, Hou, Longpre, Zoph, Tay, Fedus, Li, Wang, Dehghani, Brahma, Webson, Gu, Dai, Suzgun, Chen, Chowdhery, Castro-Ros, Pellat, Robinson, Valter, Narang, Mishra, Yu, Zhao, Huang, Dai, Yu, Petrov, Chi, Dean, Devlin, Roberts, Zhou, Le, and Wei}]{Flan-T5}
Chung, Hyung~Won, Le~Hou, Shayne Longpre, Barret Zoph, Yi~Tay, William Fedus, Yunxuan Li, Xuezhi Wang, Mostafa Dehghani, Siddhartha Brahma, Albert Webson, Shixiang~Shane Gu, Zhuyun Dai, Mirac Suzgun, Xinyun Chen, Aakanksha Chowdhery, Alex Castro-Ros, Marie Pellat, Kevin Robinson, Dasha Valter, Sharan Narang, Gaurav Mishra, Adams Yu, Vincent Zhao, Yanping Huang, Andrew Dai, Hongkun Yu, Slav Petrov, Ed~H. Chi, Jeff Dean, Jacob Devlin, Adam Roberts, Denny Zhou, Quoc~V. Le, and Jason Wei. 2022.
\newblock Scaling instruction-finetuned language models.

\bibitem[{Clark et~al.(2021)Clark, August, Serrano, Haduong, Gururangan, and Smith}]{clark-etal-2021-thats}
Clark, Elizabeth, Tal August, Sofia Serrano, Nikita Haduong, Suchin Gururangan, and Noah~A. Smith. 2021.
\newblock All that{'}s {`}human{'} is not gold: Evaluating human evaluation of generated text.
\newblock In \emph{Proceedings of the 59th Annual Meeting of the Association for Computational Linguistics and the 11th International Joint Conference on Natural Language Processing (Volume 1: Long Papers)}, pages 7282--7296.

\bibitem[{Clements and Ridouane(2011)}]{ClementsG_features}
Clements, G.~Nick and Rachid Ridouane. 2011.
\newblock \emph{Where Do Phonological Features Come From? : Cognitive, Physical and Developmental Bases of Distinctive Speech Categories.}, 1st ed. edition.
\newblock Language Faculty and Beyond Series. John Benjamins Publishing Company, Amsterdam/Philadelphia.

\bibitem[{Van~de Cruys(2020)}]{van-de-cruys-2020-automatic}
Van~de Cruys, Tim. 2020.
\newblock Automatic poetry generation from prosaic text.
\newblock In \emph{Proceedings of the 58th Annual Meeting of the Association for Computational Linguistics}, pages 2471--2480.

\bibitem[{De~Lacy(2007)}]{DeLacy_phonology}
De~Lacy, Paul~V. 2007.
\newblock \emph{The Cambridge handbook of phonology / [electronic resource]}.
\newblock Cambridge University Press, Cambridge.

\bibitem[{Devlin et~al.(2019)Devlin, Chang, Lee, and Toutanova}]{bert}
Devlin, Jacob, Ming-Wei Chang, Kenton Lee, and Kristina Toutanova. 2019.
\newblock {BERT}: Pre-training of deep bidirectional transformers for language understanding.
\newblock In \emph{Proceedings of the 2019 Conference of the North {A}merican Chapter of the Association for Computational Linguistics: Human Language Technologies, Volume 1 (Long and Short Papers)}, pages 4171--4186.

\bibitem[{Flesch(1948)}]{Flesh1948readability}
Flesch, R.R. 1948.
\newblock A new readability yardstick.
\newblock \emph{Journal of Applied Psychology}, 32:2211--2223.

\bibitem[{Foster and White(2007)}]{foster-white-2007-avoiding}
Foster, Mary~Ellen and Michael White. 2007.
\newblock Avoiding repetition in generated text.
\newblock In \emph{Proceedings of the Eleventh {E}uropean Workshop on Natural Language Generation ({ENLG} 07)}, pages 33--40.

\bibitem[{Franceschelli and Musolesi(2023)}]{franceschelli2023creativity}
Franceschelli, Giorgio and Mirco Musolesi. 2023.
\newblock On the creativity of large language models.

\bibitem[{Francis and Kucera(1979)}]{francis1979brown}
Francis, W~Nelson and Henry Kucera. 1979.
\newblock The brown corpus.
\newblock \emph{Department of Linguistics, Brown University}.

\bibitem[{Geisel(1965)}]{seuss}
Geisel, Theodore~Seuss. 1965.
\newblock \emph{Fox in socks: Dr. Seuss’s book of tongue tanglers}.
\newblock Random House.

\bibitem[{Georgievska-Jancheska(2019)}]{Georgievska-Jancheska2019-qh}
Georgievska-Jancheska, Tatjana. 2019.
\newblock Lambdacism, rhotacism and sigmatism in preschool children: Frequency and distribution.
\newblock \emph{Open Access Macedonian Journal of Medical Science}, 7(3):336--340.

\bibitem[{Gick, Wilson, and Derrick(2013)}]{gick2013articulatory}
Gick, Bryan, Ian Wilson, and Donald Derrick. 2013.
\newblock \emph{Articulatory phonetics}.
\newblock John Wiley \& Sons.

\bibitem[{G{\'o}mez-Rodr{\'\i}guez and Williams(2023)}]{gomez-rodriguez-williams-2023-confederacy}
G{\'o}mez-Rodr{\'\i}guez, Carlos and Paul Williams. 2023.
\newblock A confederacy of models: a comprehensive evaluation of {LLM}s on creative writing.
\newblock In \emph{Findings of the Association for Computational Linguistics: EMNLP 2023}, pages 14504--14528, Association for Computational Linguistics, Singapore.

\bibitem[{Guerini, {\"O}zbal, and Strapparava(2015)}]{guerini-etal-2015-echoes}
Guerini, Marco, G{\"o}zde {\"O}zbal, and Carlo Strapparava. 2015.
\newblock Echoes of persuasion: The effect of euphony in persuasive communication.
\newblock In \emph{Proceedings of the 2015 Conference of the North {A}merican Chapter of the Association for Computational Linguistics: Human Language Technologies}, pages 1483--1493, Association for Computational Linguistics, Denver, Colorado.

\bibitem[{Gunning(1971)}]{gunning-fog}
Gunning, R. 1971.
\newblock \emph{The Technique of Clear Writing}.
\newblock McGraw-Hill.

\bibitem[{Gupta and Agrawal(2022)}]{gupta_KD_survey}
Gupta, Manish and Puneet Agrawal. 2022.
\newblock Compression of deep learning models for text: A survey.
\newblock \emph{ACM Trans. Knowl. Discov. Data}, 16(4).

\bibitem[{Gupta et~al.(2019)Gupta, Mehri, Zhao, Pavel, Eskenazi, and Bigham}]{gupta-etal-2019-investigating}
Gupta, Prakhar, Shikib Mehri, Tiancheng Zhao, Amy Pavel, Maxine Eskenazi, and Jeffrey Bigham. 2019.
\newblock Investigating evaluation of open-domain dialogue systems with human generated multiple references.
\newblock In \emph{Proceedings of the 20th Annual SIGdial Meeting on Discourse and Dialogue}, pages 379--391.

\bibitem[{Hinton, Vinyals, and Dean(2015)}]{hinton2015distilling}
Hinton, Geoffrey, Oriol Vinyals, and Jeffrey Dean. 2015.
\newblock Distilling the knowledge in a neural network.
\newblock In \emph{NIPS Deep Learning and Representation Learning Workshop}.

\bibitem[{Hokamp and Liu(2017)}]{hokamp-liu-2017-lexically}
Hokamp, Chris and Qun Liu. 2017.
\newblock Lexically constrained decoding for sequence generation using grid beam search.
\newblock In \emph{Proceedings of the 55th Annual Meeting of the Association for Computational Linguistics (Volume 1: Long Papers)}, pages 1535--1546.

\bibitem[{Hong et~al.(2023)Hong, Sayeed, Mehra, Demberg, and Schiele}]{hong-etal-2023-visual-writing}
Hong, Xudong, Asad Sayeed, Khushboo Mehra, Vera Demberg, and Bernt Schiele. 2023.
\newblock Visual writing prompts: Character-grounded story generation with curated image sequences.
\newblock \emph{Transactions of the Association for Computational Linguistics}, 11:565--581.

\bibitem[{Hu et~al.(2021)Hu, Shen, Wallis, Allen-Zhu, Li, Wang, Wang, and Chen}]{hu2021lora}
Hu, Edward~J, Yelong Shen, Phillip Wallis, Zeyuan Allen-Zhu, Yuanzhi Li, Shean Wang, Lu~Wang, and Weizhu Chen. 2021.
\newblock Lora: Low-rank adaptation of large language models.
\newblock \emph{arXiv preprint arXiv:2106.09685}.

\bibitem[{Hu et~al.(2022)Hu, Shen, Wallis, Allen-Zhu, Li, Wang, Wang, and Chen}]{LoRA}
Hu, Edward~J, Yelong Shen, Phillip Wallis, Zeyuan Allen-Zhu, Yuanzhi Li, Shean Wang, Lu~Wang, and Weizhu Chen. 2022.
\newblock Lo{RA}: Low-rank adaptation of large language models.
\newblock In \emph{International Conference on Learning Representations}.

\bibitem[{Iso(2022)}]{iso2022autotemplate}
Iso, Hayate. 2022.
\newblock Autotemplate: A simple recipe for lexically constrained text generation.
\newblock In \emph{Proceedings of the 17th International Natural Language Generation Conference}.

\bibitem[{Jessen(2008)}]{Jessen_forensics}
Jessen, Michael. 2008.
\newblock Forensic phonetics.
\newblock \emph{Language and linguistics compass}, 2(4):671--711.

\bibitem[{Keh et~al.(2023)Keh, Feng, Gangal, Alikhani, and Hovy}]{keh-etal-2023-pancetta}
Keh, Sedrick~Scott, Steven~Y. Feng, Varun Gangal, Malihe Alikhani, and Eduard Hovy. 2023.
\newblock {PANCETTA}: Phoneme aware neural completion to elicit tongue twisters automatically.
\newblock In \emph{Proceedings of the 17th Conference of the European Chapter of the Association for Computational Linguistics}, pages 491--504.

\bibitem[{Kember, Connaghan, and Patel(2017)}]{Kember-dysarthria}
Kember, Heather, Kathryn Connaghan, and Rupal Patel. 2017.
\newblock Inducing speech errors in dysarthria using tongue twisters.
\newblock \emph{International journal of language \& communication disorders}, 52(4):469--478.

\bibitem[{Kingma and Ba(2014)}]{kingma2014adam}
Kingma, Diederik~P and Jimmy Ba. 2014.
\newblock Adam: A method for stochastic optimization.
\newblock \emph{arXiv preprint arXiv:1412.6980}.

\bibitem[{Klausenburger(1970)}]{Klausenburger_phonotactics}
Klausenburger, Jürgen. 1970.
\newblock \emph{French Prosodics and Phonotactics : An Historical Typology.}, 1st ed. edition.
\newblock Beihefte Zur Zeitschrift Für Romanische Philologie Series. Walter de Gruyter GmbH, Tübingen.

\bibitem[{Ladefoged(1996)}]{ladefoged1996acoustic}
Ladefoged, Peter. 1996.
\newblock \emph{Elements of acoustic phonetics}.
\newblock University of Chicago Press.

\bibitem[{Lewis et~al.(2020)Lewis, Liu, Goyal, Ghazvininejad, Mohamed, Levy, Stoyanov, and Zettlemoyer}]{bart}
Lewis, Mike, Yinhan Liu, Naman Goyal, Marjan Ghazvininejad, Abdelrahman Mohamed, Omer Levy, Veselin Stoyanov, and Luke Zettlemoyer. 2020.
\newblock {BART}: Denoising sequence-to-sequence pre-training for natural language generation, translation, and comprehension.
\newblock In \emph{Proceedings of the 58th Annual Meeting of the Association for Computational Linguistics}, pages 7871--7880.

\bibitem[{Li et~al.(2024)Li, Yuan, Zhang, Ma, Chen, Yin, Xiao, Lin, Ragni, Benetos, Gyenge, Dannenberg, Liu, Chen, Xia, Shi, Huang, Wang, Guo, and Fu}]{li2024mert}
Li, Yizhi, Ruibin Yuan, Ge~Zhang, Yinghao Ma, Xingran Chen, Hanzhi Yin, Chenghao Xiao, Chenghua Lin, Anton Ragni, Emmanouil Benetos, Norbert Gyenge, Roger Dannenberg, Ruibo Liu, Wenhu Chen, Gus Xia, Yemin Shi, Wenhao Huang, Zili Wang, Yike Guo, and Jie Fu. 2024.
\newblock {MERT}: Acoustic music understanding model with large-scale self-supervised training.
\newblock In \emph{Proceedings of the 12th International Conference on Learning Representations (ICLR)}.

\bibitem[{Li, Guerin, and Lin(2022)}]{li-etal-2022-secret}
Li, Yucheng, Frank Guerin, and Chenghua Lin. 2022.
\newblock The secret of metaphor on expressing stronger emotion.
\newblock In \emph{Proceedings of the 3rd Workshop on Figurative Language Processing (FLP)}, pages 39--43.

\bibitem[{Li et~al.(2023{\natexlab{a}})Li, Wang, Lin, Guerin, and Barrault}]{li-etal-2023-framebert}
Li, Yucheng, Shun Wang, Chenghua Lin, Frank Guerin, and Loic Barrault. 2023{\natexlab{a}}.
\newblock {F}rame{BERT}: Conceptual metaphor detection with frame embedding learning.
\newblock In \emph{Proceedings of the 17th Conference of the European Chapter of the Association for Computational Linguistics}, pages 1558--1563.

\bibitem[{Li et~al.(2023{\natexlab{b}})Li, Zhu, Lu, and Yin}]{li-etal-2023-synthetic}
Li, Zhuoyan, Hangxiao Zhu, Zhuoran Lu, and Ming Yin. 2023{\natexlab{b}}.
\newblock Synthetic data generation with large language models for text classification: Potential and limitations.
\newblock In \emph{Proceedings of the 2023 Conference on Empirical Methods in Natural Language Processing}, pages 10443--10461.

\bibitem[{Lin(2004)}]{rouge}
Lin, Chin-Yew. 2004.
\newblock {ROUGE}: A package for automatic evaluation of summaries.
\newblock In \emph{Text Summarization Branches Out}, pages 74--81.

\bibitem[{Loakman, Maladry, and Lin(2023)}]{loakman-etal-2023-iron}
Loakman, Tyler, Aaron Maladry, and Chenghua Lin. 2023.
\newblock The iron(ic) melting pot: Reviewing human evaluation in humour, irony and sarcasm generation.
\newblock In \emph{Findings of the Association for Computational Linguistics: EMNLP 2023}, pages 6676--6689.

\bibitem[{Loakman, Tang, and Lin(2023)}]{twistlist}
Loakman, Tyler, Chen Tang, and Chenghua Lin. 2023.
\newblock {T}wist{L}ist: Resources and baselines for tongue twister generation.
\newblock In \emph{Proceedings of the 61st Annual Meeting of the Association for Computational Linguistics (Volume 2: Short Papers)}, pages 579--589.

\bibitem[{Lu et~al.(2022)Lu, Welleck, West, Jiang, Kasai, Khashabi, Le~Bras, Qin, Yu, Zellers, Smith, and Choi}]{lu-etal-2022-neurologic}
Lu, Ximing, Sean Welleck, Peter West, Liwei Jiang, Jungo Kasai, Daniel Khashabi, Ronan Le~Bras, Lianhui Qin, Youngjae Yu, Rowan Zellers, Noah~A. Smith, and Yejin Choi. 2022.
\newblock {N}euro{L}ogic a*esque decoding: Constrained text generation with lookahead heuristics.
\newblock In \emph{Proceedings of the 2022 Conference of the North American Chapter of the Association for Computational Linguistics: Human Language Technologies}, pages 780--799.

\bibitem[{Manjavacas, Kestemont, and Karsdorp(2019)}]{manjavacas-etal-2019-generation}
Manjavacas, Enrique, Mike Kestemont, and Folgert Karsdorp. 2019.
\newblock Generation of hip-hop lyrics with hierarchical modeling and conditional templates.
\newblock In \emph{Proceedings of the 12th International Conference on Natural Language Generation}, pages 301--310.

\bibitem[{McCutchen and Perfetti(1982)}]{MCCUTCHEN-VISUAL}
McCutchen, Deborah and Charles~A. Perfetti. 1982.
\newblock The visual tongue-twister effect: Phonological activation in silent reading.
\newblock \emph{Journal of Verbal Learning and Verbal Behavior}, 21(6):672--687.

\bibitem[{Mortensen et~al.(2016)Mortensen, Littell, Bharadwaj, Goyal, Dyer, and Levin}]{mortensen-etal-2016-panphon}
Mortensen, David~R., Patrick Littell, Akash Bharadwaj, Kartik Goyal, Chris Dyer, and Lori Levin. 2016.
\newblock {P}an{P}hon: A resource for mapping {IPA} segments to articulatory feature vectors.
\newblock In \emph{Proceedings of {COLING} 2016, the 26th International Conference on Computational Linguistics: Technical Papers}, pages 3475--3484.

\bibitem[{O'Halloran(2020)}]{oHalloran-apnoea}
O'Halloran, Ken~D. 2020.
\newblock A tongue-twister to translation? increased complexity of genioglossus movement during wakefulness in persons with obstructive sleep apnoea.
\newblock \emph{The Journal of Physiology}, 598(3):435--436.

\bibitem[{OpenAI et~al.(2023)OpenAI, Achiam, Adler, Agarwal, Ahmad, Akkaya et~al.}]{openai2023gpt4}
OpenAI, Josh Achiam, Steven Adler, Sandhini Agarwal, Lama Ahmad, Ilge Akkaya, et~al. 2023.
\newblock Gpt-4 technical report.

\bibitem[{Ouyang et~al.(2022)Ouyang, Wu, Jiang, Almeida, Wainwright, Mishkin, Zhang, Agarwal, Slama, Ray, Schulman, Hilton, Kelton, Miller, Simens, Askell, Welinder, Christiano, Leike, and Lowe}]{RLHF-openai}
Ouyang, Long, Jeff Wu, Xu~Jiang, Diogo Almeida, Carroll~L. Wainwright, Pamela Mishkin, Chong Zhang, Sandhini Agarwal, Katarina Slama, Alex Ray, John Schulman, Jacob Hilton, Fraser Kelton, Luke Miller, Maddie Simens, Amanda Askell, Peter Welinder, Paul Christiano, Jan Leike, and Ryan Lowe. 2022.
\newblock Training language models to follow instructions with human feedback.

\bibitem[{Papineni et~al.(2002)Papineni, Roukos, Ward, and Zhu}]{bleu}
Papineni, Kishore, Salim Roukos, Todd Ward, and Wei-Jing Zhu. 2002.
\newblock {B}leu: a method for automatic evaluation of machine translation.
\newblock In \emph{Proceedings of the 40th Annual Meeting of the Association for Computational Linguistics}, pages 311--318.

\bibitem[{Ploujnikov and Ravanelli(2022)}]{soundchoice}
Ploujnikov, Artem and Mirco Ravanelli. 2022.
\newblock {SoundChoice: Grapheme-to-Phoneme Models with Semantic Disambiguation}.
\newblock In \emph{Proc. Interspeech 2022}, pages 486--490.

\bibitem[{Popescu-Belis et~al.(2023)Popescu-Belis, Atrio, Bernath, Boisson, Ferrari, Theimer-Lienhard, and Vernikos}]{popescu-belis-etal-2023-gpoet}
Popescu-Belis, Andrei, {\`A}lex~R. Atrio, Bastien Bernath, Etienne Boisson, Teo Ferrari, Xavier Theimer-Lienhard, and Giorgos Vernikos. 2023.
\newblock {GP}oe{T}: a language model trained for rhyme generation on synthetic data.
\newblock In \emph{Proceedings of the 7th Joint SIGHUM Workshop on Computational Linguistics for Cultural Heritage, Social Sciences, Humanities and Literature}, pages 10--20.

\bibitem[{Potash, Romanov, and Rumshisky(2018)}]{potash-etal-2018-evaluating}
Potash, Peter, Alexey Romanov, and Anna Rumshisky. 2018.
\newblock Evaluating creative language generation: The case of rap lyric ghostwriting.
\newblock In \emph{Proceedings of the Second Workshop on Stylistic Variation}, pages 29--38.

\bibitem[{Radford et~al.(2019)Radford, Wu, Child, Luan, Amodei, Sutskever et~al.}]{GPT-2}
Radford, Alec, Jeffrey Wu, Rewon Child, David Luan, Dario Amodei, Ilya Sutskever, et~al. 2019.
\newblock Language models are unsupervised multitask learners.
\newblock \emph{OpenAI blog}, 1(8):9.

\bibitem[{Raffel et~al.(2020)Raffel, Shazeer, Roberts, Lee, Narang, Matena, Zhou, Li, and Liu}]{T5}
Raffel, Colin, Noam Shazeer, Adam Roberts, Katherine Lee, Sharan Narang, Michael Matena, Yanqi Zhou, Wei Li, and Peter~J. Liu. 2020.
\newblock Exploring the limits of transfer learning with a unified text-to-text transformer.
\newblock \emph{Journal of Machine Learning Research}, 21(140):1--67.

\bibitem[{Rose et~al.(2010)Rose, Engel, Cramer, and Cowley}]{rake}
Rose, Stuart, Dave Engel, Nick Cramer, and Wendy Cowley. 2010.
\newblock Automatic keyword extraction from individual documents.
\newblock In \emph{Text Mining: Applications and Theory}, pages 1 -- 20.

\bibitem[{Roush et~al.(2022)Roush, Basu, Moorthy, and Dubovoy}]{roush-etal-2022-language}
Roush, Allen, Sanjay Basu, Akshay Moorthy, and Dmitry Dubovoy. 2022.
\newblock Most language models can be poets too: An {AI} writing assistant and constrained text generation studio.
\newblock In \emph{Proceedings of the Second Workshop on When Creative AI Meets Conversational AI}, pages 9--15, Association for Computational Linguistics, Gyeongju, Republic of Korea.

\bibitem[{Smith and Senter(1967)}]{ARI_index}
Smith, E~A and R~J Senter. 1967.
\newblock Automated readability index.
\newblock \emph{AMRL-TR. Aerospace Medical Research Laboratories (6570th)}, pages 1--14.

\bibitem[{Somoff(2014)}]{SOMOFF-UNMASTERY}
Somoff, Victoria. 2014.
\newblock Four is not fourteen: Tongue twister patterns and the unmastery of language.
\newblock \emph{Western Folklore}, 73(2/3):195--215.

\bibitem[{Sugiharto, Santoso, and Shofyana(2022)}]{Sugiharto-Japanese}
Sugiharto, Prasetyawan, Yan Santoso, and Maila Shofyana. 2022.
\newblock Teaching english pronunciation using tongue twister.
\newblock \emph{Acitya: Journal of Teaching and Education}, 4(1):189--197.

\bibitem[{Sun et~al.(2022)Sun, Narayan-Chen, Oraby, Gao, Chung, Huang, Liu, and Peng}]{sun-etal-2022-context}
Sun, Jiao, Anjali Narayan-Chen, Shereen Oraby, Shuyang Gao, Tagyoung Chung, Jing Huang, Yang Liu, and Nanyun Peng. 2022.
\newblock Context-situated pun generation.
\newblock In \emph{Proceedings of the 2022 Conference on Empirical Methods in Natural Language Processing}, pages 4635--4648.

\bibitem[{Tang et~al.(2022)Tang, Lin, Huang, Guerin, and Zhang}]{tang-etal-2022-etrica}
Tang, Chen, Chenghua Lin, Henglin Huang, Frank Guerin, and Zhihao Zhang. 2022.
\newblock {E}tri{CA}: Event-triggered context-aware story generation augmented by cross attention.
\newblock In \emph{Findings of the Association for Computational Linguistics: EMNLP 2022}.

\bibitem[{Tang et~al.(2023)Tang, Zhang, Loakman, Lin, and Guerin}]{tang-etal-2023-enhancing}
Tang, Chen, Hongbo Zhang, Tyler Loakman, Chenghua Lin, and Frank Guerin. 2023.
\newblock Enhancing dialogue generation via dynamic graph knowledge aggregation.
\newblock In \emph{Proceedings of the 61st Annual Meeting of the Association for Computational Linguistics (Volume 1: Long Papers)}, pages 4604--4616.

\bibitem[{Tian et~al.(2023)Tian, Narayan-Chen, Oraby, Cervone, Sigurdsson, Tao, Zhao, Chung, Huang, and Peng}]{tian-etal-2023-unsupervised}
Tian, Yufei, Anjali Narayan-Chen, Shereen Oraby, Alessandra Cervone, Gunnar Sigurdsson, Chenyang Tao, Wenbo Zhao, Tagyoung Chung, Jing Huang, and Nanyun Peng. 2023.
\newblock Unsupervised melody-to-lyrics generation.
\newblock In \emph{Proceedings of the 61st Annual Meeting of the Association for Computational Linguistics (Volume 1: Long Papers)}, pages 9235--9254.

\bibitem[{Tian, Sheth, and Peng(2022)}]{tian-etal-2022-unified}
Tian, Yufei, Divyanshu Sheth, and Nanyun Peng. 2022.
\newblock A unified framework for pun generation with humor principles.
\newblock In \emph{Findings of the Association for Computational Linguistics: EMNLP 2022}, pages 3253--3261.

\bibitem[{Touvron et~al.(2023)Touvron, Martin, Stone, Albert, Almahairi et~al.}]{llama2}
Touvron, Hugo, Louis Martin, Kevin Stone, Peter Albert, Amjad Almahairi, et~al. 2023.
\newblock Llama 2: Open foundation and fine-tuned chat models.

\bibitem[{Valitutti et~al.(2013)Valitutti, Toivonen, Doucet, and Toivanen}]{valitutti-etal-2013-everything}
Valitutti, Alessandro, Hannu Toivonen, Antoine Doucet, and Jukka~M. Toivanen. 2013.
\newblock {``}let everything turn well in your wife{''}: Generation of adult humor using lexical constraints.
\newblock In \emph{Proceedings of the 51st Annual Meeting of the Association for Computational Linguistics (Volume 2: Short Papers)}, pages 243--248.

\bibitem[{Wang et~al.(2023)Wang, Li, Lin, Barrault, and Guerin}]{wang-etal-2023-metaphor}
Wang, Shun, Yucheng Li, Chenghua Lin, Loic Barrault, and Frank Guerin. 2023.
\newblock Metaphor detection with effective context denoising.
\newblock In \emph{Proceedings of the 17th Conference of the European Chapter of the Association for Computational Linguistics}, pages 1404--1409.

\bibitem[{Wang et~al.(2024)Wang, Zhang, Wu, Loakman, Huang, and Lin}]{wang2024mmte}
Wang, Shun, Ge~Zhang, Han Wu, Tyler Loakman, Wenhao Huang, and Chenghua Lin. 2024.
\newblock {MMTE}: Corpus and metrics for evaluating machine translation quality of metaphorical language.
\newblock In \emph{Conference on Empirical Methods in Natural Language Processing (EMNLP)}.

\bibitem[{Whitehouse, Choudhury, and Aji(2023)}]{whitehouse-etal-2023-llm}
Whitehouse, Chenxi, Monojit Choudhury, and Alham Aji. 2023.
\newblock {LLM}-powered data augmentation for enhanced cross-lingual performance.
\newblock In \emph{Proceedings of the 2023 Conference on Empirical Methods in Natural Language Processing}, pages 671--686.

\bibitem[{Wilshire(1999)}]{WILSHIRE-1999}
Wilshire, Carolyn~E. 1999.
\newblock The “tongue twister” paradigm as a technique for studying phonological encoding.
\newblock \emph{Language and Speech}, 42(1):57--82.

\bibitem[{W{\"o}ckener et~al.(2021)W{\"o}ckener, Haider, Miller, Nguyen, Nguyen, Pham, Belouadi, and Eger}]{wockener-etal-2021-end}
W{\"o}ckener, J{\"o}rg, Thomas Haider, Tristan Miller, The-Khang Nguyen, Thanh Tung~Linh Nguyen, Minh~Vu Pham, Jonas Belouadi, and Steffen Eger. 2021.
\newblock End-to-end style-conditioned poetry generation: What does it take to learn from examples alone?
\newblock In \emph{Proceedings of the 5th Joint SIGHUM Workshop on Computational Linguistics for Cultural Heritage, Social Sciences, Humanities and Literature}, pages 57--66.

\bibitem[{Wong et~al.(2019)Wong, Chan, Ng, and Zhu}]{wong-broca}
Wong, Min~Ney, Yanky Chan, Manwa~L. Ng, and Frank~F. Zhu. 2019.
\newblock Effects of transcranial direct current stimulation over the broca’s area on tongue twister production.
\newblock \emph{International Journal of Speech-Language Pathology}, 21(2):182--188.
\newblock PMID: 29642741.

\bibitem[{Wright(2016)}]{Wright_2016}
Wright, Ernest~Vincent. 2016.
\newblock \emph{Gadsby: A story of over 50,000 words without using the letter “E”}.
\newblock Digital Ninjas Media, Inc.

\bibitem[{Xue et~al.(2021)Xue, Song, Wu, Tan, Zhang, Qin, Zhang, and Liu}]{xue-etal-2021-deeprapper}
Xue, Lanqing, Kaitao Song, Duocai Wu, Xu~Tan, Nevin~L. Zhang, Tao Qin, Wei-Qiang Zhang, and Tie-Yan Liu. 2021.
\newblock {D}eep{R}apper: Neural rap generation with rhyme and rhythm modeling.
\newblock In \emph{Proceedings of the 59th Annual Meeting of the Association for Computational Linguistics and the 11th International Joint Conference on Natural Language Processing (Volume 1: Long Papers)}, pages 69--81.

\bibitem[{Xue et~al.(2022)Xue, Barua, Constant, Al-Rfou, Narang, Kale, Roberts, and Raffel}]{xue-etal-2022-byt5}
Xue, Linting, Aditya Barua, Noah Constant, Rami Al-Rfou, Sharan Narang, Mihir Kale, Adam Roberts, and Colin Raffel. 2022.
\newblock {B}y{T}5: Towards a token-free future with pre-trained byte-to-byte models.
\newblock \emph{Transactions of the Association for Computational Linguistics}, 10:291--306.

\bibitem[{Yang et~al.(2023)Yang, Xiao, Wang, Zhang, Bian, Yin, Lv, Pan, Wang, Yan, Yang, Deng, Wang, Liu, Ai, Dong, Zhao, Xu, Sun, Zhang, Liu, Ji, Xie, Dai, Fang, Su, Song, Liu, Ru, Ma, Wang, Liu, Lin, Nie, Guo, Sun, Zhang, Li, Li, Cheng, Chen, Zeng, Wang, Chen, Men, Yu, Pan, Shen, Wang, Li, Jiang, Gao, Zhang, Zhou, and Wu}]{baichuan}
Yang, Aiyuan, Bin Xiao, Bingning Wang, Borong Zhang, Ce~Bian, Chao Yin, Chenxu Lv, Da~Pan, Dian Wang, Dong Yan, Fan Yang, Fei Deng, Feng Wang, Feng Liu, Guangwei Ai, Guosheng Dong, Haizhou Zhao, Hang Xu, Haoze Sun, Hongda Zhang, Hui Liu, Jiaming Ji, Jian Xie, JunTao Dai, Kun Fang, Lei Su, Liang Song, Lifeng Liu, Liyun Ru, Luyao Ma, Mang Wang, Mickel Liu, MingAn Lin, Nuolan Nie, Peidong Guo, Ruiyang Sun, Tao Zhang, Tianpeng Li, Tianyu Li, Wei Cheng, Weipeng Chen, Xiangrong Zeng, Xiaochuan Wang, Xiaoxi Chen, Xin Men, Xin Yu, Xuehai Pan, Yanjun Shen, Yiding Wang, Yiyu Li, Youxin Jiang, Yuchen Gao, Yupeng Zhang, Zenan Zhou, and Zhiying Wu. 2023.
\newblock Baichuan 2: Open large-scale language models.

\bibitem[{Yang, Tang, and Lin(2024)}]{yang2023improving}
Yang, Bohao, Chen Tang, and Chenghua Lin. 2024.
\newblock Improving medical dialogue generation with abstract meaning representations.
\newblock In \emph{ICASSP 2024 - 2024 IEEE International Conference on Acoustics, Speech and Signal Processing (ICASSP)}, pages 11826--11830.

\bibitem[{Yang et~al.(2024)Yang, Tang, Zhao, Xiao, and Lin}]{Yang2023EffectiveDO}
Yang, Bohao, Chen Tang, Kun Zhao, Chenghao Xiao, and Chenghua Lin. 2024.
\newblock Effective distillation of table-based reasoning ability from {LLM}s.
\newblock In \emph{Proceedings of the 2024 Joint International Conference on Computational Linguistics, Language Resources and Evaluation (LREC-COLING 2024)}, pages 5538--5550.

\bibitem[{Yao et~al.(2023)Yao, Chen, Hanjie, Yang, and Narasimhan}]{yao2023collie}
Yao, Shunyu, Howard Chen, Austin~W. Hanjie, Runzhe Yang, and Karthik Narasimhan. 2023.
\newblock Collie: Systematic construction of constrained text generation tasks.

\bibitem[{Yu et~al.(2023)Yu, Song, Lu, He, Tan, Ye, Zhang, and Bian}]{yu-etal-2023-musicagent}
Yu, Dingyao, Kaitao Song, Peiling Lu, Tianyu He, Xu~Tan, Wei Ye, Shikun Zhang, and Jiang Bian. 2023.
\newblock {M}usic{A}gent: An {AI} agent for music understanding and generation with large language models.
\newblock In \emph{Proceedings of the 2023 Conference on Empirical Methods in Natural Language Processing: System Demonstrations}, pages 246--255.

\bibitem[{Yuan et~al.(2024)Yuan, Lin, Wang, Tian, Wu, Shen, Zhang, Wu, Liu, Zhou, Xue, Ma, Liu, Zheng, Li, Ma, Liang, Chi, Liu, Wang, Lin, Liu, Jiang, Huang, Chen, Fu, Benetos, Xia, Dannenberg, Xue, Kang, and Guo}]{yuan-etal-2024-chatmusician}
Yuan, Ruibin, Hanfeng Lin, Yi~Wang, Zeyue Tian, Shangda Wu, Tianhao Shen, Ge~Zhang, Yuhang Wu, Cong Liu, Ziya Zhou, Liumeng Xue, Ziyang Ma, Qin Liu, Tianyu Zheng, Yizhi Li, Yinghao Ma, Yiming Liang, Xiaowei Chi, Ruibo Liu, Zili Wang, Chenghua Lin, Qifeng Liu, Tao Jiang, Wenhao Huang, Wenhu Chen, Jie Fu, Emmanouil Benetos, Gus Xia, Roger Dannenberg, Wei Xue, Shiyin Kang, and Yike Guo. 2024.
\newblock {C}hat{M}usician: Understanding and generating music intrinsically with {LLM}.
\newblock In \emph{Findings of the Association for Computational Linguistics}, pages 6252--6271.

\bibitem[{Zhang et~al.(2022)Zhang, Zhang, Mao, and Chang}]{zhang-etal-2022-qiuniu}
Zhang, Le, Rongsheng Zhang, Xiaoxi Mao, and Yongzhu Chang. 2022.
\newblock {Q}iu{N}iu: A {C}hinese lyrics generation system with passage-level input.
\newblock In \emph{Proceedings of the 60th Annual Meeting of the Association for Computational Linguistics: System Demonstrations}, pages 76--82.

\bibitem[{Zhang et~al.(2020{\natexlab{a}})Zhang, Kishore, Wu, Weinberger, and Artzi}]{bert-score}
Zhang, Tianyi, Varsha Kishore, Felix Wu, Kilian~Q. Weinberger, and Yoav Artzi. 2020{\natexlab{a}}.
\newblock Bertscore: Evaluating text generation with bert.
\newblock In \emph{International Conference on Learning Representations}.

\bibitem[{Zhang et~al.(2021)Zhang, Kamigaito, Aoki, Takamura, and Okumura}]{zhang-etal-2021-generic}
Zhang, Ying, Hidetaka Kamigaito, Tatsuya Aoki, Hiroya Takamura, and Manabu Okumura. 2021.
\newblock Generic mechanism for reducing repetitions in encoder-decoder models.
\newblock In \emph{Proceedings of the International Conference on Recent Advances in Natural Language Processing (RANLP 2021)}, pages 1606--1615.

\bibitem[{Zhang et~al.(2020{\natexlab{b}})Zhang, Sun, Galley, Chen, Brockett, Gao, Gao, Liu, and Dolan}]{dialogpt}
Zhang, Yizhe, Siqi Sun, Michel Galley, Yen-Chun Chen, Chris Brockett, Xiang Gao, Jianfeng Gao, Jingjing Liu, and Bill Dolan. 2020{\natexlab{b}}.
\newblock {DIALOGPT} : Large-scale generative pre-training for conversational response generation.
\newblock In \emph{Proceedings of the 58th Annual Meeting of the Association for Computational Linguistics: System Demonstrations}, pages 270--278.

\bibitem[{Zhuo et~al.(2023)Zhuo, Yuan, Pan, Ma, Li, Zhang, Liu, Dannenberg, Fu, Lin, Benentos, Xue, and Guo}]{zhuo2023lyricwhiz}
Zhuo, Le, Ruibin Yuan, Jiahao Pan, Yinghao Ma, Yizhi Li, Ge~Zhang, Si~Liu, Roger Dannenberg, Jie Fu, Chenghua Lin, Emmanouil Benentos, Wang Xue, and Yike Guo. 2023.
\newblock Lyricwhiz: Robust multilingual lyrics transcription by whispering to chatgpt.
\newblock International Society for Music Information Retrieval Conference (ISMIR).

\end{thebibliography}

\end{document}